\def\ThisFile{CTDQ_arXiv.tex}
\def\ThisFileDate{2026/02/22}
\documentclass[11pt]{article}
\usepackage{ifthen,pifont,latexsym}
\usepackage[T1]{fontenc}
\usepackage{color}
\usepackage{times,avant}
\usepackage[round,authoryear]{natbib}
\usepackage{amsmath}
\usepackage{amsfonts}
\usepackage{amssymb}
\usepackage{setspace}
\usepackage[pdftex]{graphicx}
{\end{description}}
{\end{itemize}}
{\end{enumerate}}
\def\Includefigs{true} 
\gdef\TD{$\mathcal{T}$} 
\def\VD{$\mathcal{V}$}  
\DeclareMathOperator*{\argmax}{arg\,max}

\def\NGS{Next Generation Sequencers (NGS)
  \gdef\NGS{NGS}
  \gdef\NGSs{NGSs}}
\def\NGSs{Next Generation Sequencers (NGSs)
  \gdef\NGS{NGS}
  \gdef\NGSs{NGSs}}
\def\SNP{single nucleotide polymorphism (SNP)
  \gdef\SNP{SNP}
  \gdef\SNPs{SNPs}}
\def\MDS{multidimensional scaling (MDS)
  \gdef\MDS{MDS}}
\def\NCBI{National Center for Biotechnology Information (NCBI)
    \gdef\NCBI{NCBI}}
\def\MLE{maximum likelihood estimator (MLE)
  \gdef\MLE{MLE}
  \gdef\MLEs{MLEs}}
\def\ECDFs{empirical cumulative distribution functions (ECDFs)
  \gdef\ECDF{ECDF}
  \gdef\ECDFs{ECDFs}}
\def\ROC{receiver operating characteristic (ROC)
    \gdef\ROC{ROC}}
\def\TSE{total survey error (TSE)
    \gdef\TSE{TSE}}

\title{Effects of Training Data Quality on\\Classifier Performance}

\author{
  Alan F. Karr\thanks{Department of Statistics, Operations and Data Science, Temple University, Philadelphia PA 19122 and Fraunhofer USA Center Mid-Atlantic, Riverdale MD 20737; alan.karr@temple.edu}
  \and
  Regina Ruane\thanks{Computational Social Science Laboratory, University of Pennsylvania, Philadelphia, PA 19104; jeanneruane@gmail.com}
}

\date{\today}

\begin{document}

\maketitle

\begin{abstract}
We describe extensive numerical experiments assessing and quantifying how classifier performance depends on the quality of the training data, a frequently neglected component of the analysis of classifiers. 

More specifically, in the scientific context of metagenomic assembly of short DNA reads into ``contigs,'' we examine the effects of degrading the quality of the training data by multiple mechanisms, and for four classifiers---Bayes classifiers, neural nets, partition models and random forests. We investigate both individual behavior and congruence among the classifiers. We find breakdown-like behavior that holds for all four classifiers, as degradation increases and they move from being mostly correct to only coincidentally correct, because they are wrong in the same way. In the process, a picture of spatial heterogeneity emerges: as the training data move farther from analysis data, classifier decisions degenerate, the boundary becomes less dense, and congruence increases.
\end{abstract}

\section{Introduction}\label{sec.intro}
As articulated in \cite{amost2024}, (supervised) classifiers should be construed as software systems with at least three components, the classifier itself, the \emph{training data} on which it draws, and the \emph{analysis data} to which it is applied. When diagnosing problems, all three of these components matter. In this paper, we focus on the training data, which we believe are (too) often neglected, since quality of the training data affect performance on all analysis datasets. While there may be general acknowledgement that classifiers become less trustworthy the  more the analysis data depart from the training data, there is little quantification of either ``less trustworthy'' or ``depart from.''

To address this neglect, we conducted and report here an extensive set of numerical experiments with
\begin{itemize}
\item 
A single analysis dataset consisting of short DNA sequences;
\item
Four classifiers: a Bayes classifier, a neural net, a partition model, and a random forest model;
\item
A single parent training dataset whose quality is degraded in multiple ways, especially, but not exclusively, using the SNP degradation in \cite{dqdegradation-2021}.
\end{itemize}
The experimental results include correctness, Boundary Status \citep{bayesboundary2026}, congruence among the four classifiers, and Neighbor Similarity \citep{bayesboundary2026}.


It is important to understand the restriction to decreased data quality, the detailed rationale for which can be found in \cite{dqdegradation-2021}. In reality, improving data quality is expensive and often statistically problematic. (Exceptions are imputation and editing in official statistics, for which there is strong theoretical and methodological basis.) By contrast, data quality can readily be decreased, provided that the mechanism is contextually credible, as it is in \cite{dqdegradation-2021}. Here, we use both it and others that are equally credible.

Finally, we note the important adversarial aspect of data quality degradation. While data quality problems are typically conceived as the result of uncontrollable but benign external forces, they may also result from actions deliberately intended to cause harm, such as attempts to ``poison'' genomic databases 
\citep{farbiash-puzis-2020, amost2024}.

\section{Background and Problem Formulation}\label{sec.background}
This section contains background on data quality, metagenomic assembly, classifiers, and classifier boundaries.

\subsection{Data Quality}\label{subseg.dq}
Attention to data quality has waxed and waned over time \citep{dqreport, dq-statmeth06, keller-dq-frameworks2017}. Given currently generated volumes of data, data quality effort per data element may be at an all-time low. There are exceptions, of course, especially official statistics, in which the \TSE\ paradigm \citep{omb2002, omb-standards-2006, tse-2017, sdl-tse-2017} has perpetuated and to some degree codified longstanding attention to data quality. Other than data volume, perhaps the most pressing issues are inability to quantify data quality at any level, from the data point to the dataset, and properly to incorporate data quality as a source of uncertainty in statistical analyses.

\begin{figure}
\centering
\includegraphics[width=3in]{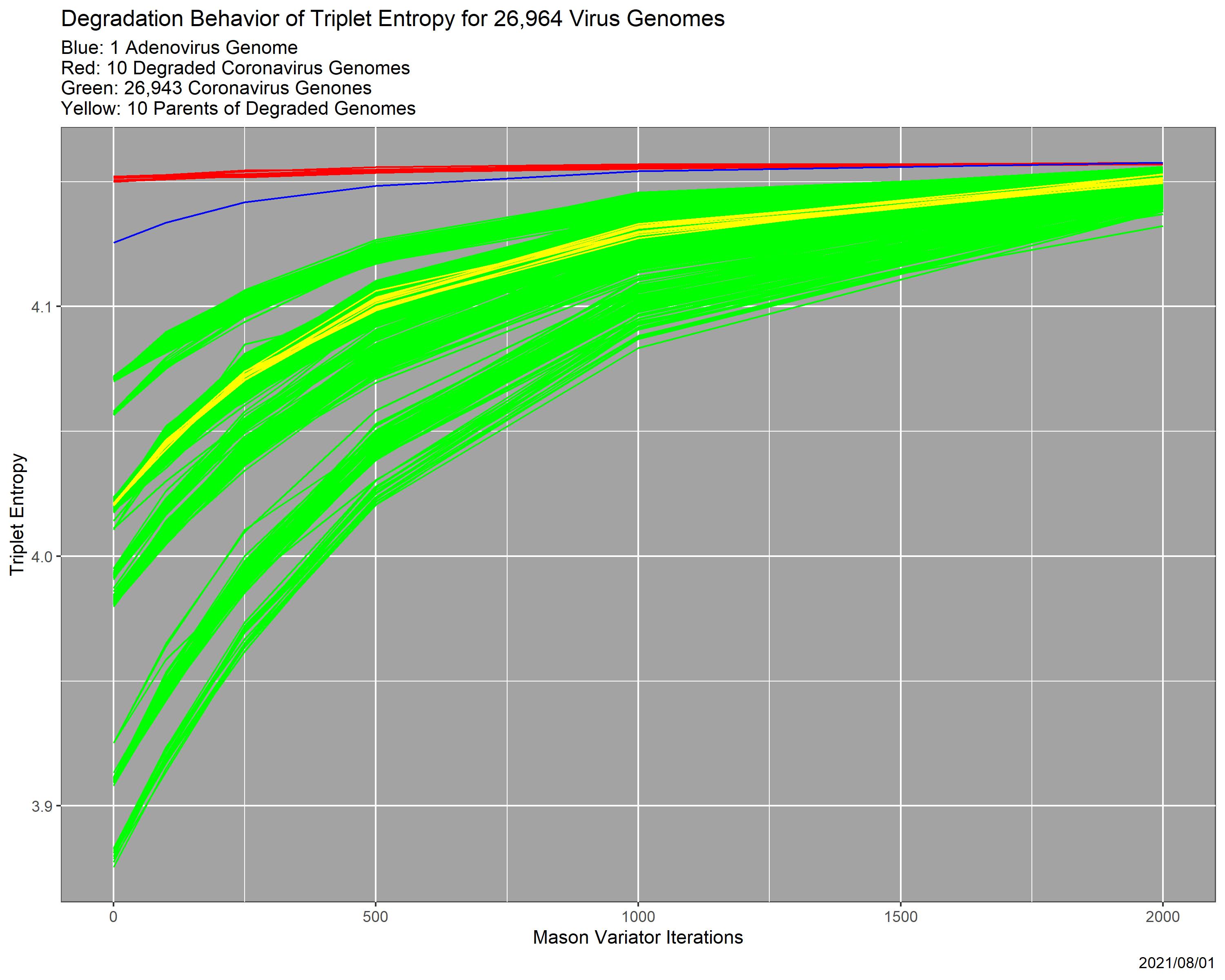}
\caption{Effect of SNP degradation \citep{dqdegradation-2021} on the entropy of 26964 virus genomes.}
\label{fig.entropy-degradation}
\end{figure}

Initial investigations in \cite{dqdegradation-2021} exploit the ease with which we can decrease (synonymously, ``degrade'') data quality in ways that both are plausible scientifically and can be parameterized. This setting appears in Section \ref{subsec.snp}. The effects of deliberately decreasing data quality can also be used to characterize and even quantify the quality of individual data points. Across a range of scientific settings, higher quality data points are more ``fragile,'' and are affected more by the same amount of degradation than low quality data points. This is illustrated graphically in Figure \ref{fig.entropy-degradation}. There, 26,964 virus genomes are degraded by successive applications of the \texttt{Mason\_variator} haplotype generation software \citep{fu_mi_publications962}. In the figure, there are 26,943 coronavirus genomes in green, previously degraded versions of 10 of these in red, and one adenovirus genome in blue. (For reference, the parents of the 10 degraded genomes in red are in yellow.) The $x$-axis is the number of iterations of the \texttt{Mason\_variator}. The $y$-axis is entropy, with the interpretation that higher values mean lower quality.

Evidently, there are ``decreasing returns to scale'' from increased degradation. The coronavirus genomes fall into several classes with respect to initial entropy, which is argued in \cite{markovstructure-2021} to reflect inherent quality. The ten previously degraded genomes begin with the lowest quality (highest entropy) and are barely affected by further degradation. This is the heart of our contention that ``the higher the quality, the greater the vulnerability to degradation.''

\subsection{Scientific Context}\label{subsec.context}
As in \cite{dqdegradation-2021} and \cite{bayesboundary2026}, the scientific problem underlying this paper is to classify short (in our case, length 101) DNA sequences as having arisen from one of three candidate virus genomes. The broader context is reference-guided \emph{metagenomic assembly}---piecing together fragments (reads) of DNA from multiple potential sources into longer sequences called contigs, as performed, for example, by \texttt{MetaCompass} \citep{metacompass2017, metacompass2024}, with the assistance of a (reference) database of sequenced candidate genomes. The role of the reference database is to reduce the computational burden of creating the contigs, which is formulated as finding an Eulerian path in a de Bruijn graph \citep{debruijn2023}. The medical setting involved samples from human digestive and reproductive tracts \citep{amost2024}.

\subsection{Classifiers}\label{subsec.classifiers}
In this paper, a classifier is a function $C$ from a finite input space $\mathcal{I}$ to a finite output space $\mathcal{O}$. Elements of $\mathcal{I}$ are \emph{inputs} or \emph{data}, and $C(x) \in \mathcal{O}$ is the \emph{decision} or \emph{result} for input $x$. Although it is not a logical necessity, almost always $|\mathcal{I}| >> |\mathcal{O}|$, where $|\mathcal{S}|$ is the cardinality of the set $\mathcal{S}$.

Following \cite{bayesboundary2026}, we assume that the input space $\mathcal{I}$ is an \emph{loopless, undirected graph}, whose edgeset $E$ defines neighboring inputs: $x,y \in \mathcal{I}$ are \emph{neighbors} if and only $x \neq y$ and $\{x,y\} \in E$. We denote by $\mathcal{N}(x) = \{y: \{x,y\} \in E\}$ the set of neighbors, or neighborhood, of $x$. For ease of interpretation, we further assume that $\mathcal{I}$ is \emph{connected}: for any $x,y \in \mathcal{I}$, there is a path $(x_1=x, \dots, x_k = y)$ that connects them: $\{x_j, x_{j+1}\} \in E$ for each $j$.

In this paper, $\mathcal{I}$ consists of all \NGS-generated DNA sequences of length 101. With five possible values for each nucleotide (see below for an explanation of the fifth value), $|\mathcal{I}| = 5^{101}$. With three candidate genomes, $|\mathcal{O}| = 3$. Neighboring reads differ in exactly one of 101 bases, which is known as a \SNP.


\subsection{Classifier Boundaries}\label{subsec.boundaries}
The boundary $\mathcal{B}$ associated with $C$ is the set of elements of $\mathcal{I}$ at least one of whose neighbors is classified differently:
\begin{equation}
\mathcal{B}_C = \{x \in \mathcal{I}: C(x') \neq C(x) \mbox{ for some } x' \in \mathcal{N}(x)\}.
\label{eq.boundary}
\end{equation}
For DNA reads, this means that there are points differing at only one base that are classified differently. Crucially, whenever two inputs are classified differently, any path in $\mathcal{I}$ connecting them must contain at least two (neighboring) boundary points. 

Below we focus on \emph{Hamming paths}, which are the shortest. Given reads $R \neq R'$ with Hamming distance $k$ (They differ in $k$ locations.), a Hamming path from $R$ to $R'$ replaces one differing nucleotide in $R$ at a time by the corresponding nucleotide in $R'$. Therefore, each successive pair on the path are neighbors. Such a path $\big(R_0 = R, \dots, R_k = R'\big)$ has length $k+1$; there are $k!$ of them. The ``simplest'' of these paths simply moves left to right in the character strings. Because $C(R) \neq C(R')$, there is at least one value of $j$ for which $C(R_j) \neq C(R_{j+1})$, in which case, the neighbors $R_j$ and $R_{j+1}$ both belong to $\mathcal{B}_C$. 

\subsection{Surrogate Uncertainty Measures}\label{subsec.surrogate-uncertainty-measures}
The boundary engenders two measures of uncertainty that are applicable to any classifier whose input space is a graph \cite{bayesboundary2026}. For $x \in \mathcal{I}$, both are functions of the distribution on $\mathcal{O}$ of $\{C(x'): x' \in \mathcal{N}(x)\}$. We define them in generality.

The \emph{Boundary Status}, of $x \in \mathcal{I}$ is the number of points in $\mathcal{O}$ other than that assigned to $x$ itself appearing among the decisions for its neighbors:
\begin{equation}
\mbox{BS}_C(x) = \sum_{z \in \mathcal{O}, z \neq C(x)} 1\left(z \in \{C(x'): x' \in \mathcal{N}(x)\}\right), 
\label{eq.bs}
\end{equation}
where $1(\cdot)$ is an indicator function, equal to one if its argument is true and zero otherwise. In particular, $\mbox{BS}_C(x) = 0$ means that all neighbors are classified the same as $x$, i.e., $x$ is not on the boundary. At the other extreme, $\mbox{BS}_C(x) = |\mathcal{O}| - 1$ means that all of the other possible decisions appear among the neighbors. Note that $\mbox{BS}_C(x)$ does not depend on whether $C(x) \in  \{C(x'): x' \in \mathcal{N}(x)\}.$ 

The second measure, \emph{Neighbor Similarity}, quantifies the extent to which the neighbors of a point have the same decision it does. The Neighbor Similarity for an input point $x$ is
\begin{equation}
\mbox{NS}_C(x) = 1 - H(q_x, q_{\mathcal{N}(x)}),
\label{eq.ns}
\end{equation}
where $H$ denotes Hellinger distance \citep{nikulin-hd-2010}, $q_x$ is the degenerate probability distribution on $\mathcal{O}$ concentrated on $C(x)$, and $q_{\mathcal{N}(x)}$ is the probability distribution on $\mathcal{O}$ of $\{C(x'): x' \in \mathcal{N}(x)\}$. Therefore,
\begin{equation}
\mathcal{B}_C = \{x \in \mathcal{I}: \mbox{NS}_C(x) < 1\}.
\label{eq.boundary-ns}
\end{equation}
Unlike $\mbox{BS}_C$, $\mbox{NS}_C$ takes account only of \emph{how many} neighbors differ, not their values.

Our principal proposed interpretation of $\mbox{NS}_C(x)$ is as uncertainty regarding the classification of $x$. If $\mbox{NS}_C(x) = 1$, \emph{all} neighbors of $x$ have the same decision as $R$, and, intuitively, we would be more certain of the decision for $x$. At the other extreme, if $\mbox{NS}_C(x)$ were zero, moving to any neighbor of $x$ changes the decision, in which case we would be highly uncertain about $C(x)$. In the case study in \cite{bayesboundary2026}, this does not occur: the maximum number of differently classified neighbors is 393 (of 404).

However---and this is important, as discussed further in Section \ref{sec.results}, we must distinguish certainty from correctness. In some regions of $\mathcal{I}$, Neighbor Similarity may be high because ``everything'' is assigned the same, but incorrect, value. 

\section{Experimental Protocol}\label{sec.protocol}
Here we present the protocol for our numerical experiments,


\subsection{Mathematical Preliminaries}\label{subsec.math}
To establish notation, a DNA sequence $S$ is a character string chosen from the nucleotide (base) alphabet  $\{A, C, G, T\}$, where A is adenine, C is cytosine, G is guanine and T is thiamine. At one extreme, $S$ may be an entire genome--for instance, a virus or a human chromosome. At another, it may be a read generated by a \NGS. Given a sequence $S$, its length is $|S|$; the $i^{\mathrm{th}}$ base in $S$ is $S(i)$; and the bases from location $i$ to location $j > i$ are $S(i:j)$.

We focus on triplets, whose distributions are 64-dimensional summaries of sequences.The \emph{triplet distribution} $P_3(\cdot|S)$ of a sequence $S$ is defined as
\begin{equation}
P_3(b_1b_2b_3|S) = \mathrm{Prob}\big\{S(k:k+2) = b_1b_2b_3\big\}
\label{eq.tripletdistribution}
\end{equation}
for each choice of $b_1b_2b_3$, where $b_1$, $b_2$ and $b_3$ are elements of $\{A, C, G, T\}$, and $k$ is chosen at random from $1, \dots, |S|-2$. An equivalent perspective is that of a second-order Markov chain \citep{markovstructure-2021}.

That triplets suffice in this setting is argued at length in \cite{bayesboundary2026} and \cite{markovstructure-2021}. Briefly, the Hellinger distances between the triplet distributions for the three genomes are so large \citep{bayesboundary2026} that the triplet distributions differentiate the genomes. And, of course, triplets are fundamental biologically because they encode amino acids. Other cases appear in the literature, especially quartets, also referred to as tetranucleotides \citep{pride-tetra-2003, teeling-tetra-classification-2004, teeling-tetra-web-2004}. In our experience, quartets are more burdensome computationally than triplets without being more useful.

\subsection{Training and Validation Datasets}\label{subsec.datasets}
As the ``root'' training dataset \TD\ for all four classifiers (Section \ref{subsec.four-classifiers}), we use the same dataset employed in \cite{dqdegradation-2021} and \cite{bayesboundary2026}. It consists of 5869 simulated \NGS\ reads of length 101:
\begin{description}
\item[Adeno:] 
1966 reads from an adenovirus genome of length 34,125, downloaded with the read simulator \texttt{Art};
\item[COVID:] 
1996 reads from a SARS-CoV-2 genome of length 29,926 contained in a coronavirus dataset downloaded from \NCBI\ in November, 2020;
\item[SARS:] 
1907 reads from a SARS-CoV genome of length 29,751 from the same database as COVID.
\end{description}
Thus, $\mathcal{O} = \{\mbox{Adeno}, \mbox{COVID}, \mbox{SARS}\}$. 

We employed the \texttt{Mason\_simulator} software \citep{fu_mi_publications962} to simulate Illumina \NGS\ reads from each genome, with approximate 6X coverage. (Illumina manufactures \NGSs\ that employ an optical technology; see \texttt{https://www.illumina.com/}.) \texttt{Mason\_simulator} introduces errors in the form of transpositions (\SNPs), insertions, deletions and undetermined bases. The latter are cases in which the sequencer detects a nucleotide, but is unable to determine whether it is A, C, G or T. Following convention, these appear in the simulated reads as ``N,'' and must be accommodated in computations. Parameters of the \texttt{Mason\_simulator} were set at default values.

The validation dataset \VD, which we earlier also termed the analysis dataset, is effectively a clone of the training dataset, consisting of 2000 reads from each of the three genomes generated using the  \texttt{Mason\_simulator}. It is used to evaluate classifier performance.

\subsection{The Four Classifiers}\label{subsec.four-classifiers}
Here we introduce the four classifiers investigated in this paper, as well as compare their performance on the training dataset. Throughout, the response is the read source and the predictors are the triplet distribution for the read.

\subsubsection{Bayes Classifier}\label{subsubsec.bayes}
The first classifier is that in \cite{bayesboundary2026}. Three likelihood functions, denoted by $L(\cdot|\mbox{Adeno})$, $L(\cdot|\mbox{COVID})$, and $L(\cdot|\mbox{SARS})$, are calculated from the triplet distributions. We assume a uniform prior $\pi_R = (1/3, 1/3, 1/3)$ on $\mathcal{O}$ for each read. We then use Bayes' theorem and the three likelihoods to calculate posterior probabilities $p(\cdot|R)$ over $\mathcal{O}$, which yield the classifier decisions $C(R) = \argmax_{x\in \mathcal{O}} p(x|R)$. Table \ref{tab.confusion-bayes} contains the resultant confusion matrix for training dataset, which shows that the Bayes classifier performs well, albeit not spectacularly. 

\begin{table}[htbp]
\caption{Confusion matrix for the Bayes classifier and undegraded training data $\mathcal{T}$. The correct classification rate is 81.58\%.}
\label{tab.confusion-bayes}
\begin{center}
\begin{tabular}{l|rrr|r}
\hline
 & \multicolumn{3}{c|}{Decision} &
\\
Source & Adeno & COVID & SARS & Sum
\\
\hline
  Adeno & 1580 & 114 & 272 & 1966 \\
  COVID & 58 & 1738 & 200 & 1996 \\
  SARS & 253 & 184 & 1470 & 1907 \\
\hline
  Sum & 1891 & 2036 & 1942 & 5869 \\
\hline
\end{tabular}\end{center}
\end{table}

\subsubsection{Neural Net Classifier}\label{subsubsec.nn}
Our neural net is an ``out-of-the box'' version of the \texttt{neuralnet} package in R with default settings. The algorithm is resilient backpropagation with weight backtracking \citep{riedmiller1994}, the  activation function is logistic, and there is one hidden neuron per layer. The confusion matrix for \TD\ is in Table \ref{tab.confusion-neuralnet}.

\begin{table}[ht]
\caption{Confusion matrix for the neural net classifier and undegraded training data $\mathcal{T}$. The correct classification rate is 76.20\%.}
\label{tab.confusion-neuralnet}
\begin{center}
\begin{tabular}{r|rrr|r}
  \hline
 & Adeno & COVID & SARS & Sum \\
  \hline
Adeno & 1626 & 40 & 300 & 1966 \\
  COVID & 51 & 1724 & 221 & 1996 \\
  SARS & 532 & 253 & 1122 & 1907 \\
\hline
  Sum & 2209 & 2017 & 1643 & 5869 \\
   \hline
\end{tabular}
\end{center}
\end{table}

\subsubsection{Partition Model Classifier}\label{subsubsec.pm}
We employ a regression tree \citep{friedmanhastietibshirani-2001}, as implemented in the \texttt{rpart} package in R. Standard heuristics for pruning, which select a complexity parameter on the basis of 10-fold cross-validation, lead to a tree with 549 terminal nodes, which is still clearly over-fitted. Instead, we pruned to a tree with 61 terminal nodes. The confusion matrix appears in Table \ref{tab.confusion-partition}.

\begin{table}[ht]
\caption{Confusion matrix for the partition model classifier and undegraded training data $\mathcal{T}$. The correct classification rate is 75.81\%.}
\label{tab.confusion-partition}
\begin{center}
\begin{tabular}{r|rrr|r}
  \hline
 & 0 & 1 & 2 &  \\
  \hline
 & Adeno & COVID & SARS & Sum \\
  \hline
  Adeno & 1491 & 93 & 382 & 1966 \\
  COVID & 130 & 1640 & 226 & 1996 \\
  SARS & 333 & 256 & 1318 & 1907 \\
\hline
  Sum & 1954 & 1989 & 1826 & 5869 \\
   \hline
\end{tabular}
\end{center}
\end{table}

\subsubsection{Random Forest Classifier}\label{subsubsec.rf}
This classifier is the R package \texttt{randomforest} with no modifications. Table \ref{tab.confusion-randomforest} contains the confusion matrix.

\begin{table}[ht]
\caption{Confusion matrix for the random forest classifier and undegraded training data $\mathcal{T}$. The correct classification rate is 91.78\%.}
\label{tab.confusion-randomforest}
\begin{center}
\begin{tabular}{r|rrr|r}
  \hline
 & 0 & 1 & 2 & Sum \\
  \hline
  Adeno & 1815 & 62 & 89 & 1966 \\
  COVID & 57 & 1896 & 43 & 1996 \\
  SARS & 117 & 114 & 1676 & 1907 \\
  \hline
  Sum & 1989 & 2072 & 1808 & 5869 \\
   \hline
\end{tabular}
\end{center}
\end{table}

\subsection{Structure of an Experiment}\label{subsec.general-description}
All of the experiments reported in Section \ref{sec.results} have the same structure, which we describe here.

\begin{description}
\item[Training data:]
Begin with training dataset \TD\ from Section \ref{subsec.datasets}.
\item[Training data quality degradation:]
Select a degradation method, which in most cases is parameterized in a way that represents increasing degradation, and apply it to the training dataset. For instance, in Section \ref{subsec.snp}, we use the \SNP\ probability-parameterized method employed in \cite{dqdegradation-2021}.
\item[Classifier execution:]
For each value of the degradation parameter, re-train all four classifiers on the degraded training dataset, and apply it to the (never altered) validation dataset \VD.
\item[Classifier performance evaluation:]
For each case (as defined above), summarize classifier behavior for the validation dataset \VD\ in terms of 
\begin{itemize}
\item
Predictions, including the correct classification rate, and
\item
Boundary Status distribution.
\end{itemize}
In Section \ref{subsec.snp}, these appear in Figures \ref{fig.snp-correct}--\ref{fig.snp-bs}.
\item[Classifier congruence:] 
For each value of the degradation parameter, calculate the extent to which the classifiers concur, exemplified by Figure \ref{fig.snp-congruence}. As discussed below, one recurrent finding is that congruence often degenerates slowly as the data quality decreases, drops precipitously, then rises again. 
\end{description}

\section{Results and Findings}\label{sec.results}
Before exploring details, we emphasize the clear message that \emph{training data quality matters}. 

For comparison with results below, Figure \ref{fig.confusion-validation} contains the confusion matrices for \VD\ for the four classifiers trained on  the undegraded training dataset.

\begin{figure}[ht]
\centering
\begin{small}
\begin{tabular}{l|rrr|r}
\hline
 & \multicolumn{3}{c|}{Decision} &
\\
Source & Adeno & COVID & SARS & Sum
\\
\hline
  Adeno & 1566 & 96 & 338 & 2000 \\
  COVID & 58 & 1734 & 208 & 2000 \\
  SARS & 253 & 232 & 1515 & 2000 \\
\hline
  Sum & 1877 & 2062 & 2061 & 6000 \\
\hline
\end{tabular}\hspace{.5in}\begin{tabular}{l|rrr|r}
\hline
 & \multicolumn{3}{c|}{Decision} &
\\
Source & Adeno & COVID & SARS & Sum
\\
\hline
  Adeno & 1649 & 42 & 309 & 2000 \\
  COVID & 98 & 1741 & 200 & 2000 \\
  SARS & 533 & 300 & 1167 & 2000 \\
\hline
  Sum & 2241 & 2083 & 1676 & 6000 \\
\hline
\end{tabular}

\vspace{.5in}\begin{tabular}{l|rrr|r}
\hline
 & \multicolumn{3}{c|}{Decision} &
\\
Source & Adeno & COVID & SARS & Sum
\\
\hline
  Adeno & 1581 & 110 & 309 & 2000 \\
  COVID & 117 & 1567 & 316 & 2000 \\
  SARS & 399 & 281 & 1320 & 2000 \\
\hline
  Sum & 2097 & 1958 & 1945 & 6000 \\
\hline
\end{tabular}\hspace{.5in}\begin{tabular}{l|rrr|r}
\hline
 & \multicolumn{3}{c|}{Decision} &
\\
Source & Adeno & COVID & SARS & Sum
\\
\hline
  Adeno & 1744 & 65 & 191 & 2000 \\
  COVID & 74 & 1831 & 95 & 2000 \\
  SARS & 212 & 207 & 1581 & 2000 \\
\hline
  Sum & 2020 & 2103 & 1867 & 6000 \\
\hline
\end{tabular}
\end{small}
\caption{Confusion matrices for the validation dataset \VD\ for the four classifiers trained on the undegraded training data \TD. \emph{Top left:} Bayes classifier. \emph{Top right}: neural net. \emph{Bottom left:} partition model. \emph{Bottom right:} random forest}
\label{fig.confusion-validation}
\end{figure}

\subsection{SNP Degradation}\label{subsec.snp}
As mentioned previously, the degradation mechanism here is that of \cite{dqdegradation-2021}. There is a parameter called SNP\_Probability belonging to $(0,1)$, and each base in a DNA sequence is changed to another value with that probability, independently of the others. The software used here and in \cite{dqdegradation-2021} is the \texttt{Mason\_variator} \citep{fu_mi_publications962}. The mechanism mimics nature because the changes are \SNPs.

In most cases, the vaiues of SNP\_Probability are 
$$
\{0, 0.01, 0.05, 0.10, 0.25, 0.50, 0.60, 0.65, 0.70, 0.75, 0.80, 0.85, 0.90, 0.95\},
$$
and the randomization is done independently at each step, as opposed to successively degrading additional reads.

\begin{figure}[ht]
\begin{center}
\includegraphics[width=3in]{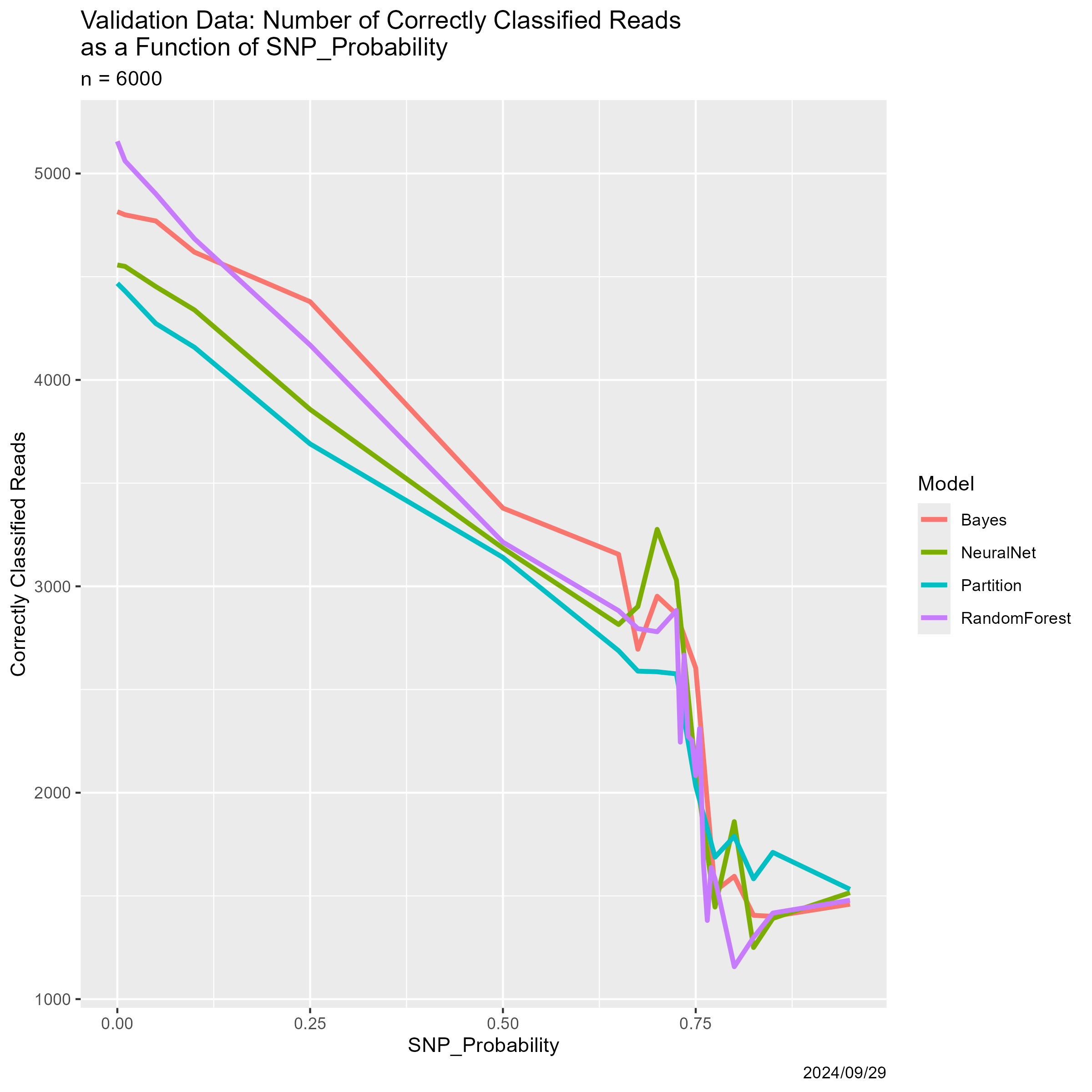}
\end{center}
\caption{SNP degradation: number of correctly classified elements of \VD\ as a function of SNP\_Probability.}
\label{fig.snp-correct}
\end{figure}
\begin{figure}[ht]
\begin{center}
\includegraphics[width=4in]{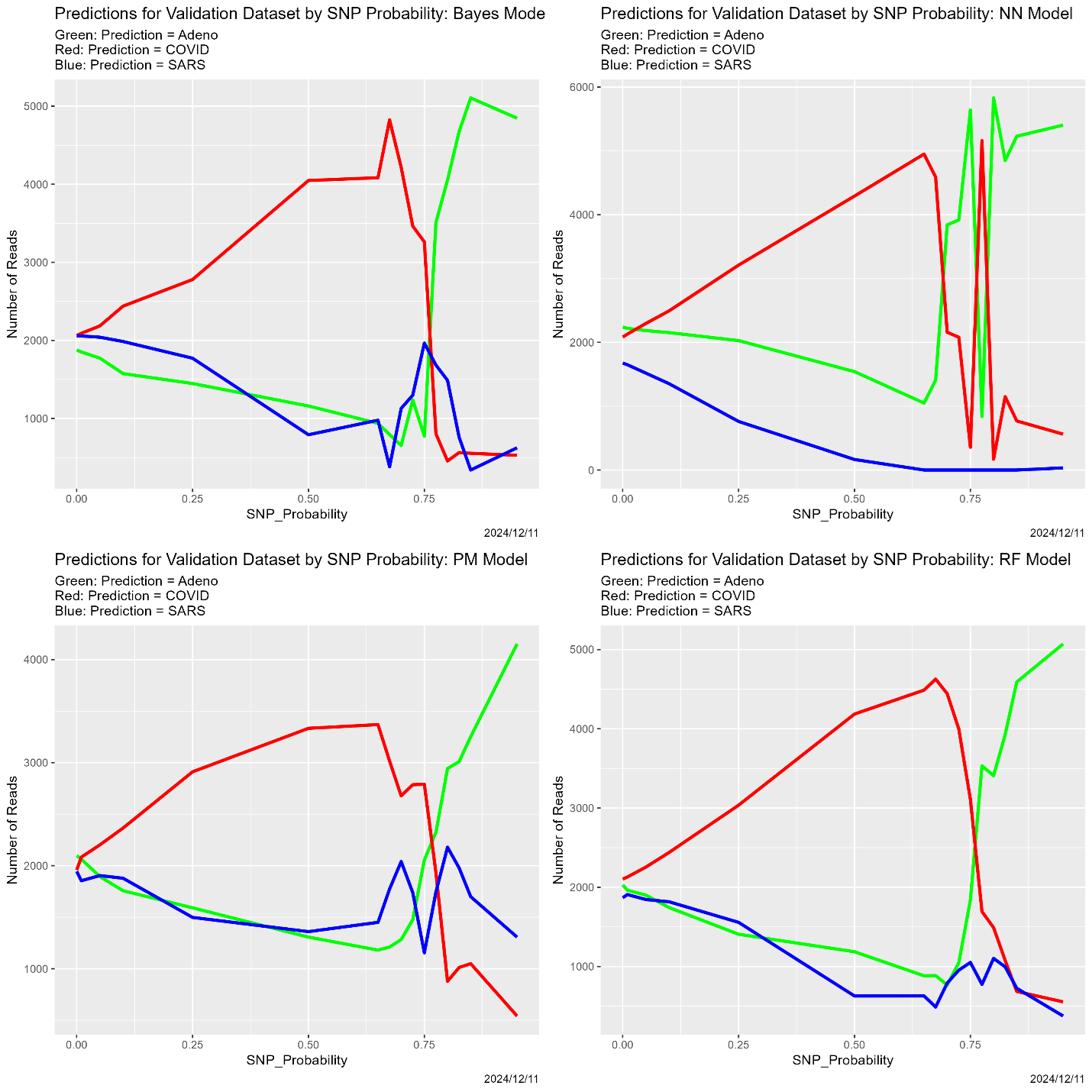}
\end{center}
\caption{SNP degradation: classifier predictions as a function of SNP\_Probability. \emph{Upper left:} Bayes classifier. \emph{Upper right:} neural net. \emph{Lower left:} partition model. \emph{Lower right:} random forest. }
\label{fig.snp-pred}
\end{figure}
\begin{figure}[ht]
\begin{center}
\includegraphics[width=2in]{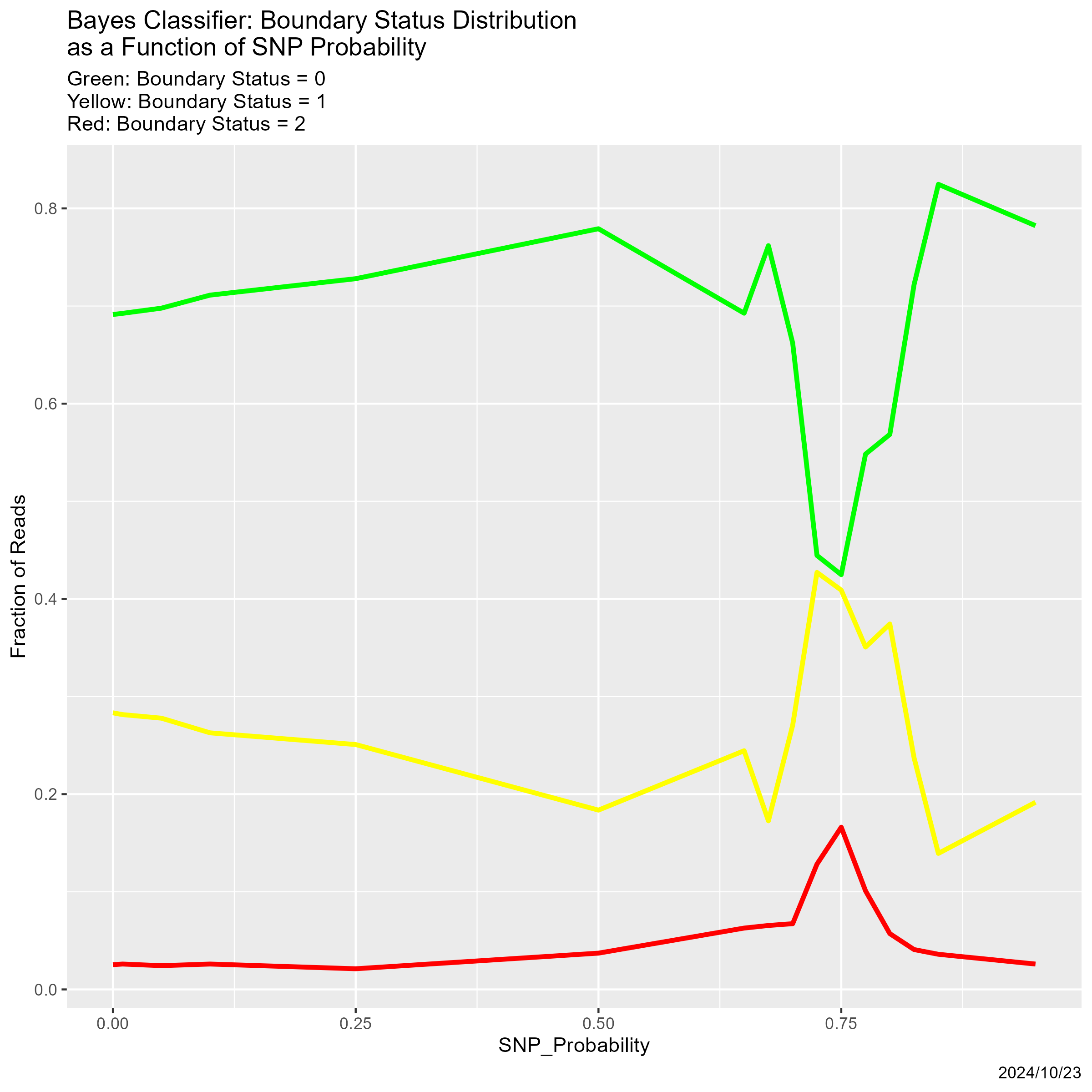}\hspace{.5in}\includegraphics[width=2in]{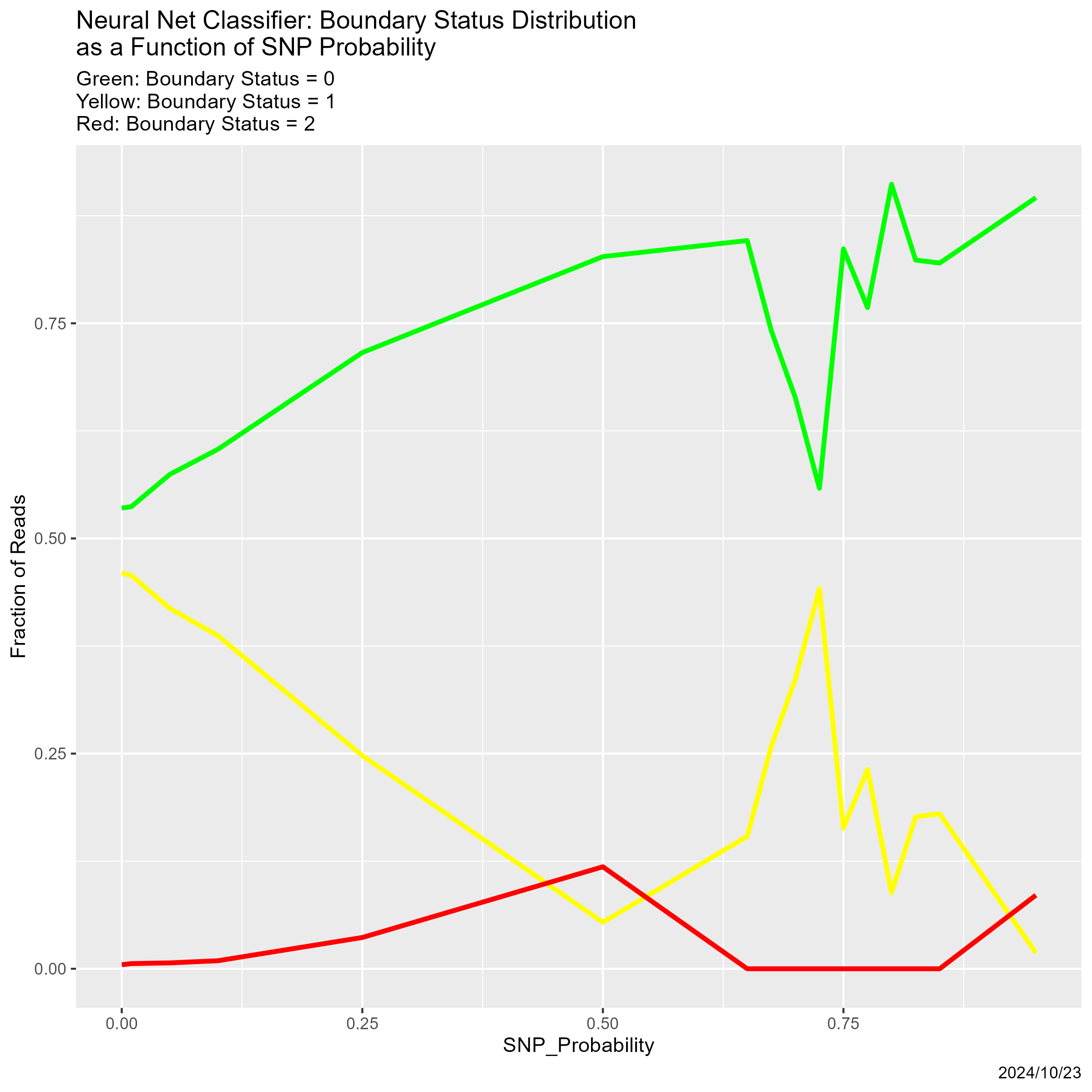}

\vspace{.5in}
\includegraphics[width=2in]{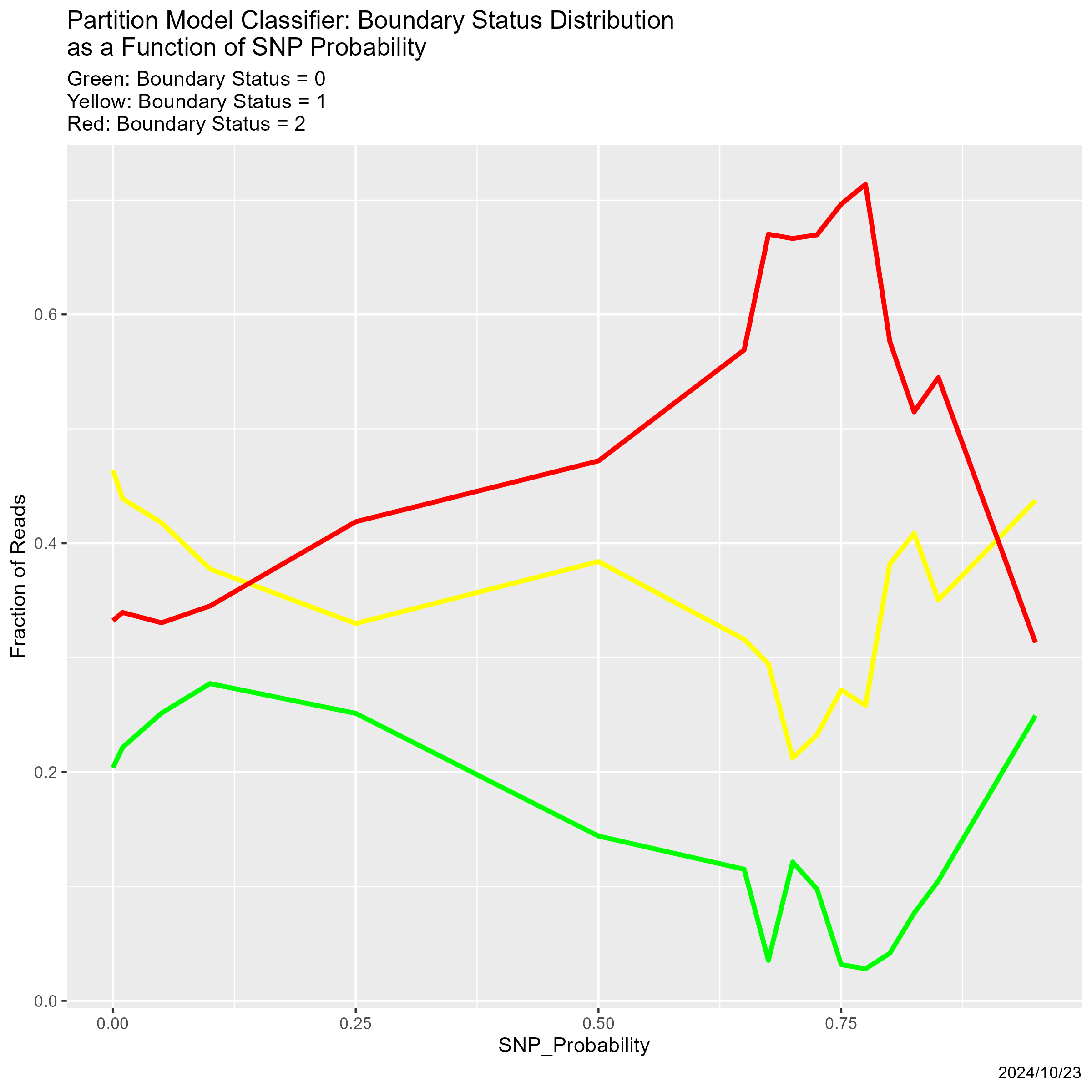}\hspace{.5in}\includegraphics[width=2in]{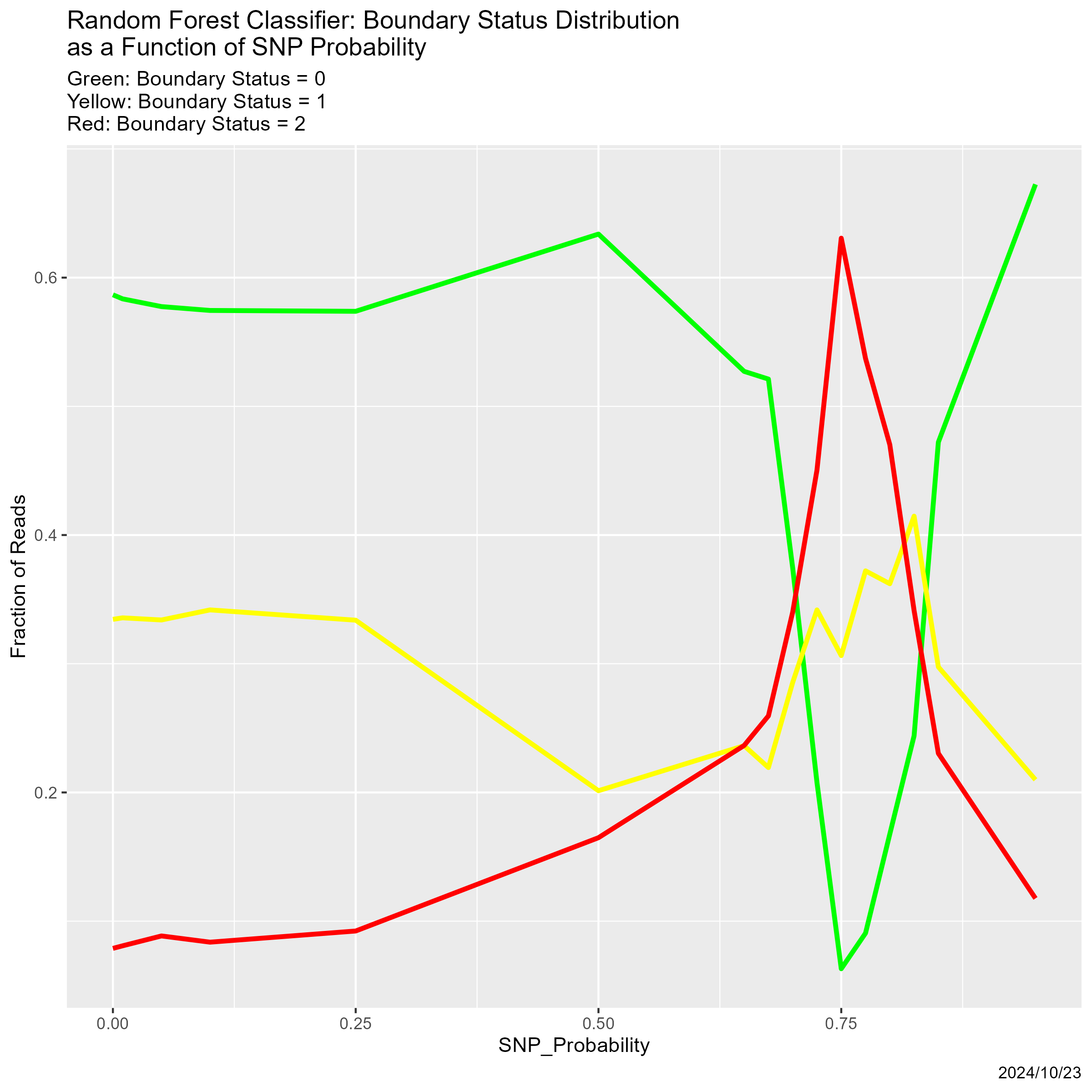}

\end{center}
\caption{SNP degradation: Boundary Status distribution as function of SNP\_Probability. \emph{Upper left:} Bayes classifier. \emph{Upper right:} neural net. \emph{Lower left:} Partition model. \emph{Lower right:} random forest. In each of these, $\mbox{BS} = 0$ is the green line, $\mbox{BS} = 1$ is the yellow line, and $\mbox{BS} = 2$ is the red line.}
\label{fig.snp-bs}
\end{figure}
\begin{figure}[ht]
\begin{center}
\includegraphics[width=3in]{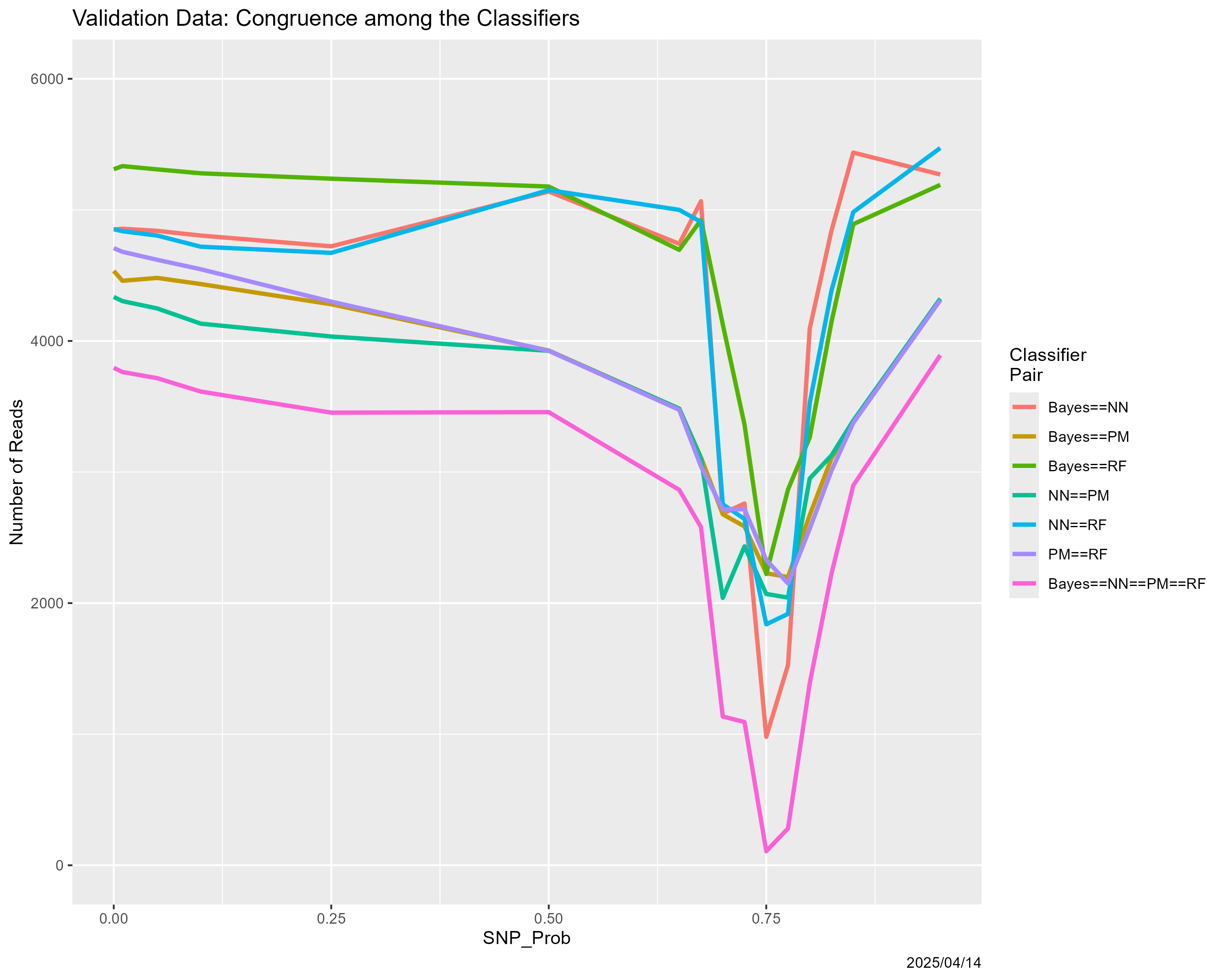}
\end{center}
\caption{SNP degradation: classifier congruence as a function of SNP\_Probability.}
\label{fig.snp-congruence}
\end{figure}

The results appear in Figures \ref{fig.snp-correct}--\ref{fig.snp-congruence}. They are at once unsurprising and unexpected. Starting with Figure \ref{fig.snp-correct}, that correctness decreases as SNP\_Probability increases is completely expected. Until SNP\_Probability is approximately 0.75, the decrease is linear, and the rate of decrease is essentially the same for all four classifiers. What happens in the vicinity of SNP\_Probability = 0.75 is, to least to us, unforeseen. Performance of all four classifiers drops precipitously, and never recovers. This breakdown behavior appears in other cases below, again uniformly across the four classifiers, so it seems not to be a function of the classifier or the mode of degradation. 

Figure \ref{fig.snp-pred} begins to illuminate the situation. Starting at the breakdown, almost all predictions become Adeno, again for all four classifiers. That is, once the training data are degraded sufficiently, the classifiers lose predictive power. This also explains why the number of correctly classified reads does not become zero in Figure \ref{fig.snp-correct}, because those from the Adeno genome are classified correctly. It is not clear why Adeno, rather than COVID or SARS, becomes the default prediction for classifiers trained on what is effectively noise.

In Figure \ref{fig.snp-bs}, things become more complex, and the four classifiers no longer behave identically. The behaviors for the Bayes classifier, the neural net and the random forest are qualitatively similar. The Boundary Status distribution is relatively stable when SNP\_Probability < 0.75, with a slight increase in the proportion of validation dataset reads with $\mbox{BS} = 0$, then there is a sharp drop in the number of validation dataset reads with $\mbox{BS} = 0$ (with most of the compensating increase being $\mbox{BS} = 1$). Past SNP\_Probability = 0.75, the increase in the proportion with $\mbox{BS} = 0$ resumes. Among these three, the random forest stands out because of the sharp increase in $\mbox{BS} = 2$, and corresponding decrease in $\mbox{BS} = 0$. The behavior in these three cases is consistent with the preceding paragraph. For SNP\_Probability < 0.75, the boundary composition is relatively constant. At SNP\_Probability = 0.75, it changes dramatically, with $\mbox{BS}$ values of 1 and 2 becoming much more prevalent. Once SNP\_Probability > 0.75, as everything is predicted to be Adeno, the boundary starts to disappear. Not that the boundary is ever small. The fraction of validation dataset reads for which $\mbox{BS} = 0$ is essentially never above .8 until the classifiers degenerate into predicting everything to be Adeno.

The partition model differs from the other three classifiers in that its boundary is much larger. The proportion of validation dataset reads with $\mbox{BS} = 0$ is the smallest of the three, and never exceeds 0.3. This means that of 70\% validation dataset reads lie on the boundary. But still something happens in the vicinity of SNP\_Probability = 0.75, followed by a return to the behavior for SNP\_Probability < 0.75. The partition model is also an outlier in other cases below.

Figure \ref{fig.snp-congruence} completes the story. It shows that six pairwise congruences between the classifiers---the numbers of reads for which they agree, as well as the four-way congruence. The differences among the pairwise congruences are meaningful, and we return to them momentarily. The overall message reinforces the preceding discussion. For SNP\_Probability < 0.75, congruences are high, and because all four of the classifiers are generally correct. Around SNP\_Probability = 0.75, the congruences go haywire. For SNP\_Probability > 0.75, they become high again, but because the four classifiers are incorrect in the same way: each classifier is predicting ``all Adeno.'' There results an interesting implication regarding ensemble methods. For SNP\_Probability < 0.75 (higher quality training data), ensembling may not help because the four classifiers may generally be correct, whereas for SNP\_Probability > 0.75 (lower quality training data), it may not help because all have degenerated in the same way.

The pairwise congruences in Figure \ref{fig.snp-congruence}, while qualitatively similar, are not identical quantitatively. Moreover, they are ordered numerically in a way that is largely independent of SNP\_Probability. That for Bayes--random forest is highest, followed by Bayes--neural net, then neural net--random forest, then partition model--random forest, Bayes--partition model, and neural net--partition model. Although we do not pursue the issue here, the partition model seems to be the most deviant. 

For this case only, results for Neighbor Similarity appear in Figures \ref{fig.snp-ns-bayes}--\ref{fig.snp-ns-rf}, which we have placed in Appendix A because of their size. There is one figure for each classifier, and within each, one panel per value of SNP\_Probability. The graphs are of densities of the associated values of Neighbor Similarity. Blank panels correspond to cases where Neighbor Similarity is identically one, so that there is no density function. Figure \ref{fig.snp-ns-bayes}, for the Bayes classifier, is consistent with other interpretations. For low and high values of SNP\_Probability, Neighbor Similarity is high, for different reasons. In the vicinity, of SNP\_Probability. it deteriorates. Behavior for the neural net (Figure \ref{fig.snp-ns-nn}) is similar, with less pronounced deterioration. By contrast, for the random forest (Figure \ref{fig.snp-ns-rf}), the deterioration becomes true breakdown. The partition model (Figure \ref{fig.snp-ns-pm,}) is truly different. We simply cannot explain, or even hypothesize a reason for, the persistent trimodality in the distribution of Neighbor Similarity. The deterioration as SNP\_Probability increases, which appears as a leftward shift in the modes, remains.

Finally, an interesting complementarity for the Bayes classifier appears in \cite{bayesboundary2026}, but there with respect to the quality of the validation dataset with the training dataset held constant and undegraded. If \VD\ is replaced by random reads, or reads from bacterial or human genomes, the more distant those are from the training data, the more the decisions for it degenerate into ``all Adeno.''

The big question, of course, is ``Why SNP\_Probability = 0.75?'' We defer this to Section \ref{sec.discussion}.

\subsection{Variants of SNP Degradation}\label{subsec.snp-variants}
\begin{figure}[ht]
\begin{center}
\includegraphics[width=3in]{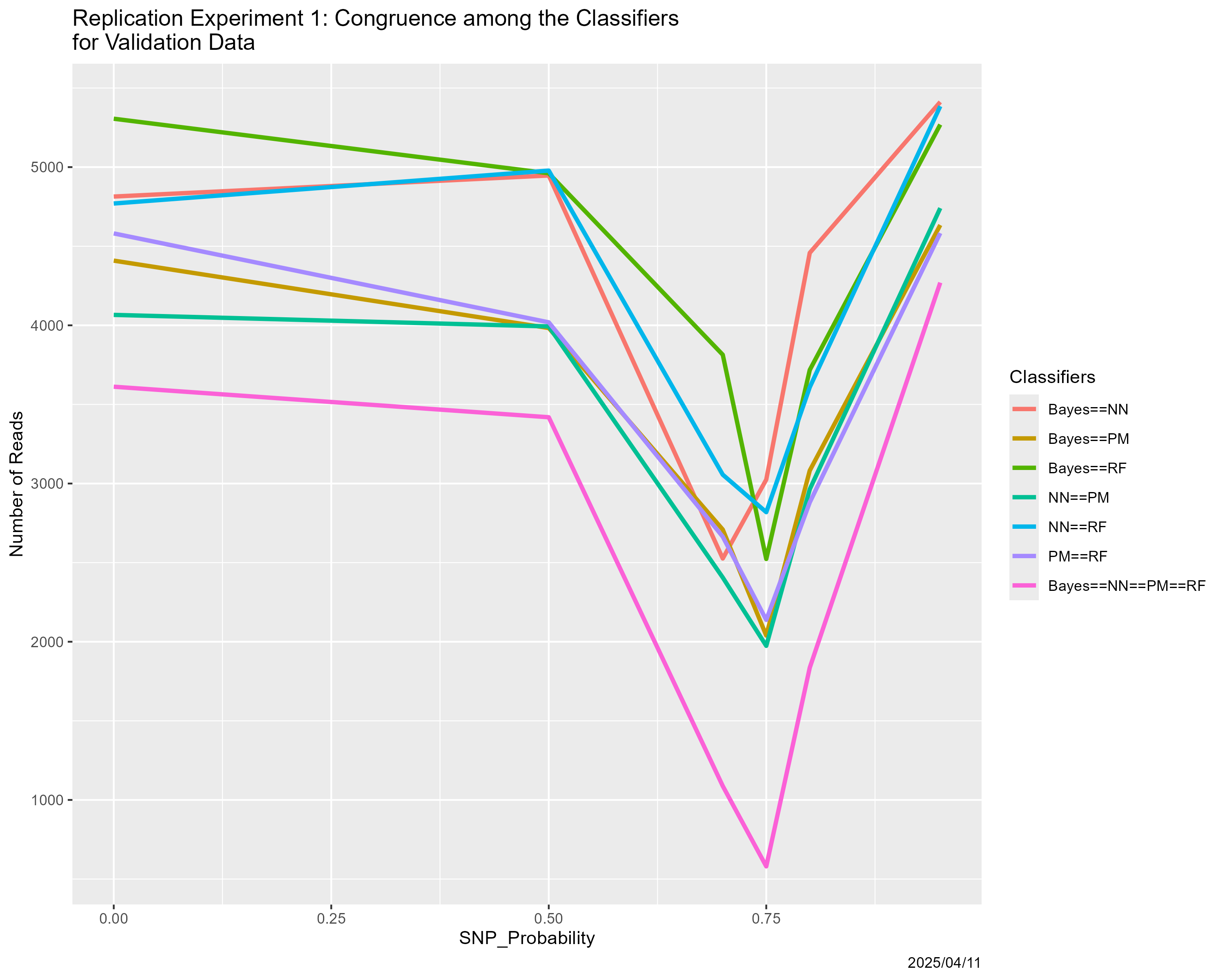}
\end{center}
\caption{SNP degradation: classifier congruence as a function of SNP\_Probability for the \emph{replicated version} of the experiment from Section \ref{subsec.snp}.}
\label{fig.snp-congruence-repl}
\end{figure}

Here we discuss several variants of the degradation in Section \ref{subsec.snp}. Before doing so, we note that the experiment in that section was replicated in its entirety, with results that were virtually identical. For brevity, we include the resultant analog of Figure \ref{fig.snp-congruence}, namely Figure \ref{fig.snp-congruence-repl}. The two are virtually identical.

\begin{figure}[ht]
\centering
\includegraphics[width = 1.75in]{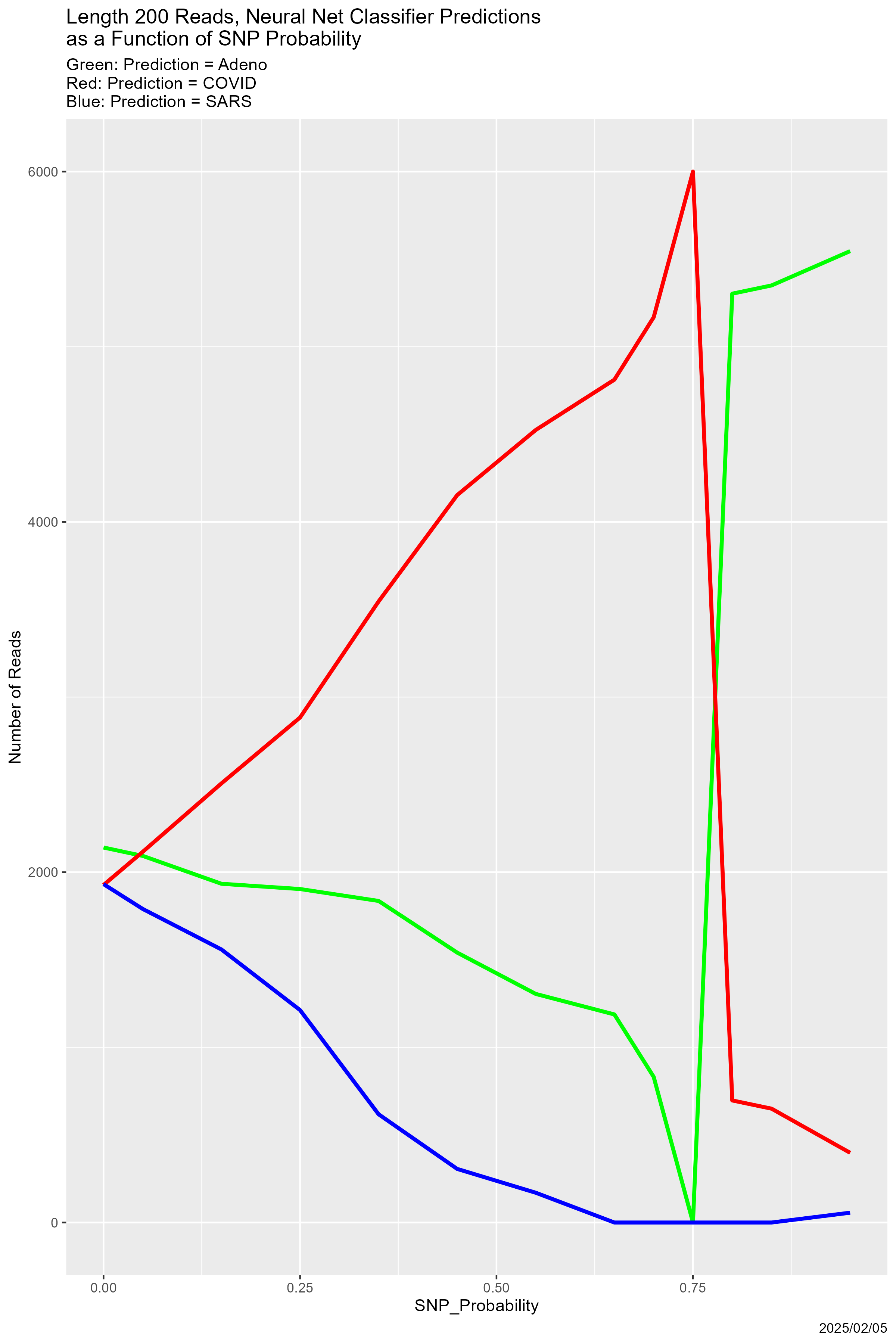}\hspace{.25in}
\includegraphics[width = 1.75in]{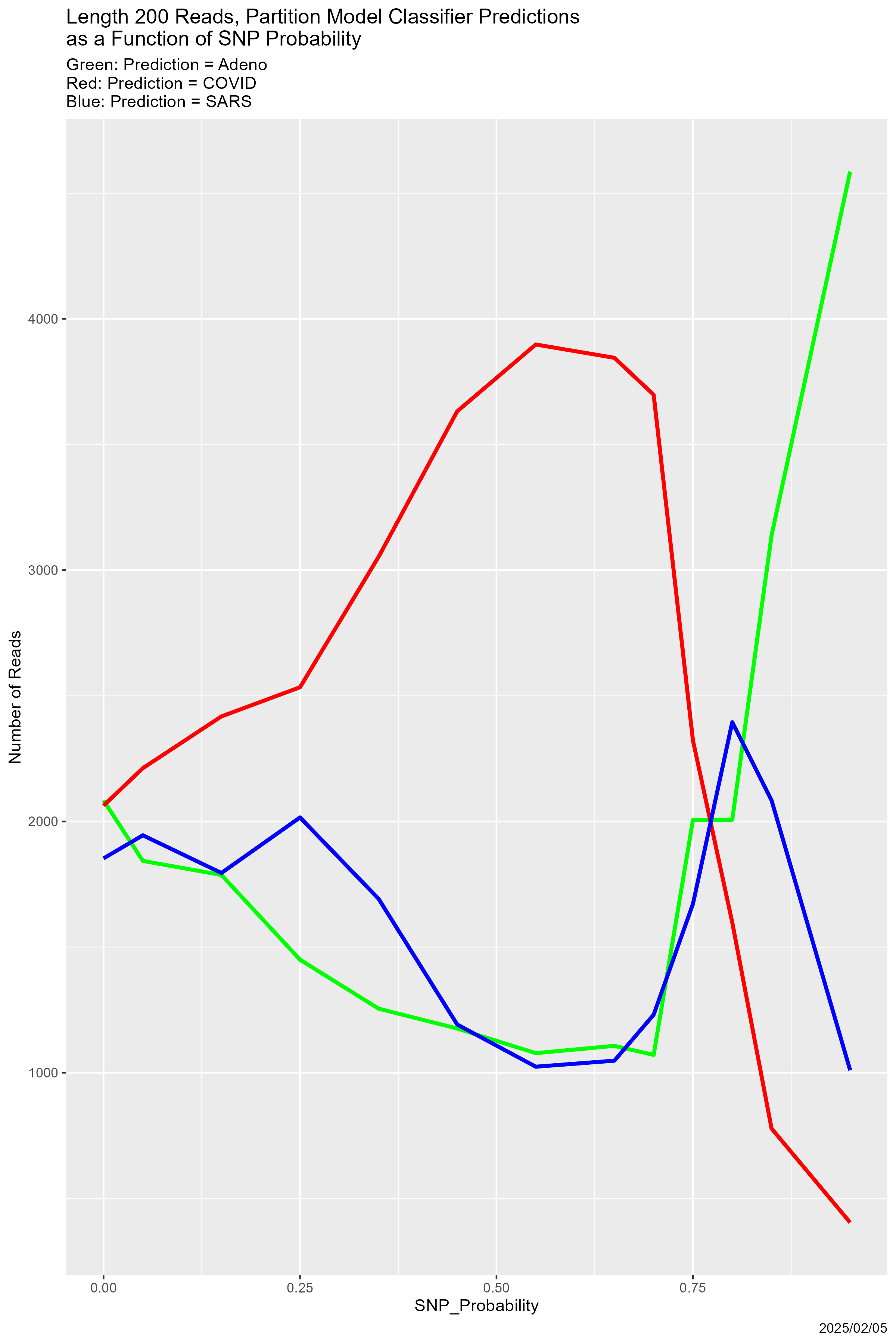}\hspace{.25in}
\includegraphics[width = 1.75in]{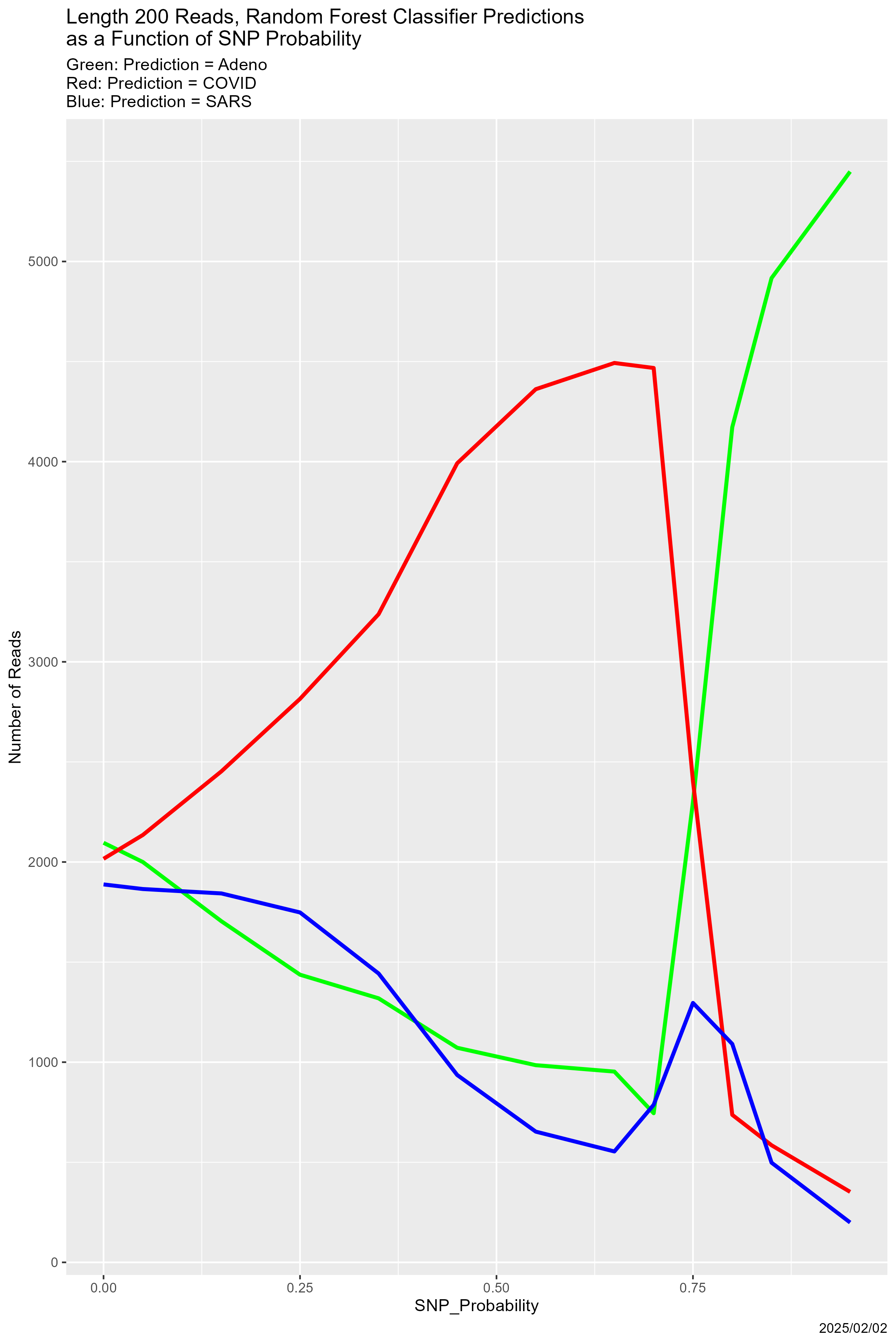}
\caption{Length 200 reads: classifier predictions as a function of SNP\_Probability. \emph{Green:} prediction = Adeno. \emph{Red:} prediction = COVID. \emph{Blue:} prediction = SARS.}
\label{fig.length200-predictions}
\end{figure}

\begin{figure}[ht]
\centering
\includegraphics[width=3in]{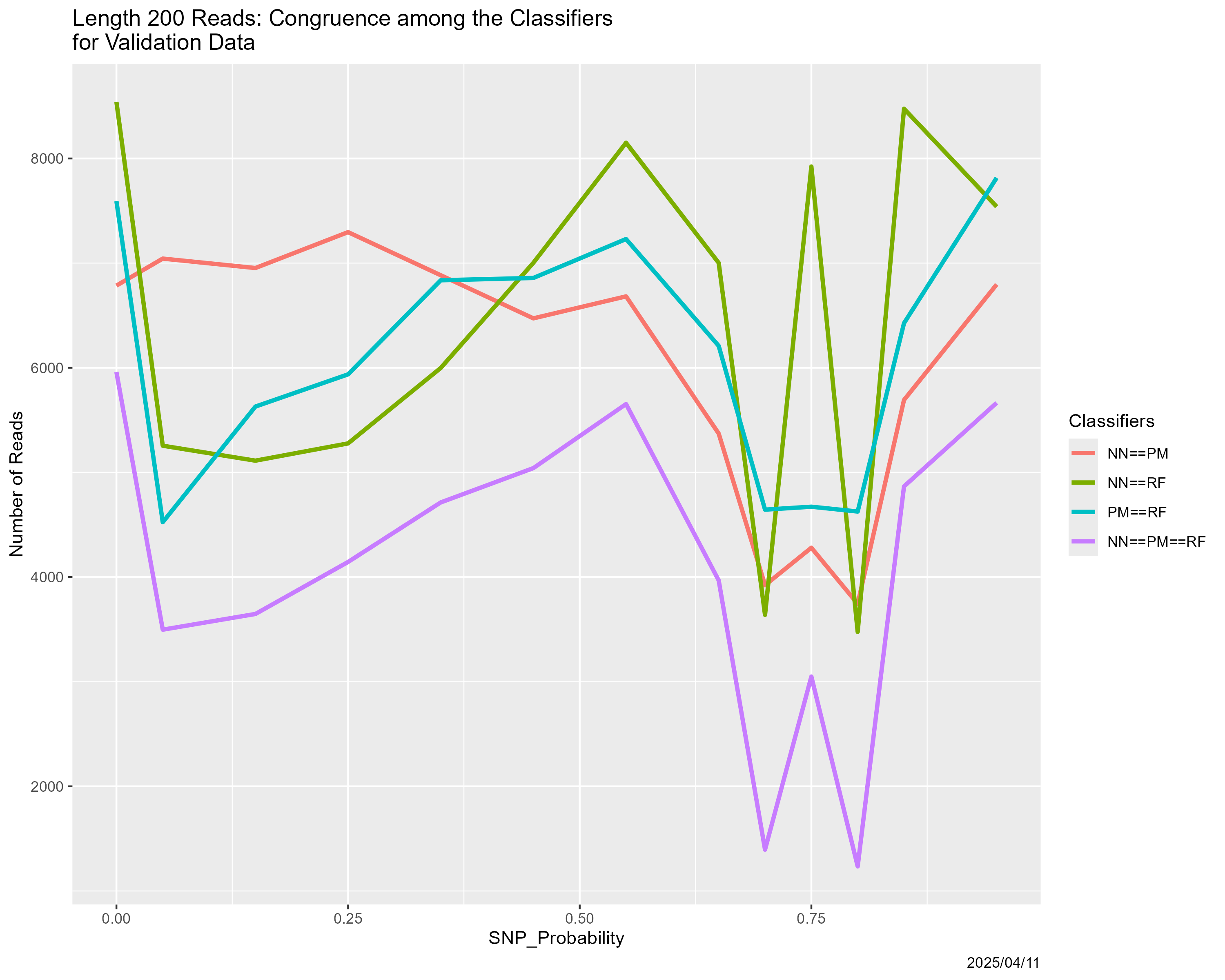}
\caption{Length 200 reads: classifier congruence as a function of SNP\_Probability.}
\label{fig.length200-congruence}
\end{figure}

Nor is anything about the preceding results associated with the read length. Changing the read length to 200 makes no difference. The three panels in Figure \ref{fig.length200-predictions} are virtually identical to the corresponding panels in Figure \ref{fig.snp-pred}, even the to shapes of the plotted lines. Breakdown occurs at SNP\_Probability = 0.75. Figure \ref{fig.length200-congruence}, however, presents complications. While the breakdown is evident, and reproduces Figure \ref{fig.snp-congruence}, the early decline in all congruences involving the random field classifier is novel. There is recovery as SNP\_Probability approaches 0.5, but we lack an explanation. It simply be random variation, because none of our experiments other than the first one has yet been replicated, and even for that one, there are only two replicates.

To control the length of the paper, for the other variants in this section, we include only the figures for predictions and classifier congruence. Also, because of the computational burden associated with the Bayes classifier and since it is treated in detail in \cite{bayesboundary2026}, we have omitted it.

\subsubsection{Initially Degraded Training Data}\label{subsubsec.idtd}
Especially in light of Figure \ref{fig.entropy-degradation}, it is natural to wonder what happens if the initial training data are degraded. Especially, does the breakdown move to the left, meaning that it depends on the absolute degradation? Or, does it remain at 0.75, meaning that degradation is relative? 

\begin{figure}[ht]
\centering
\includegraphics[width = 1.75in]{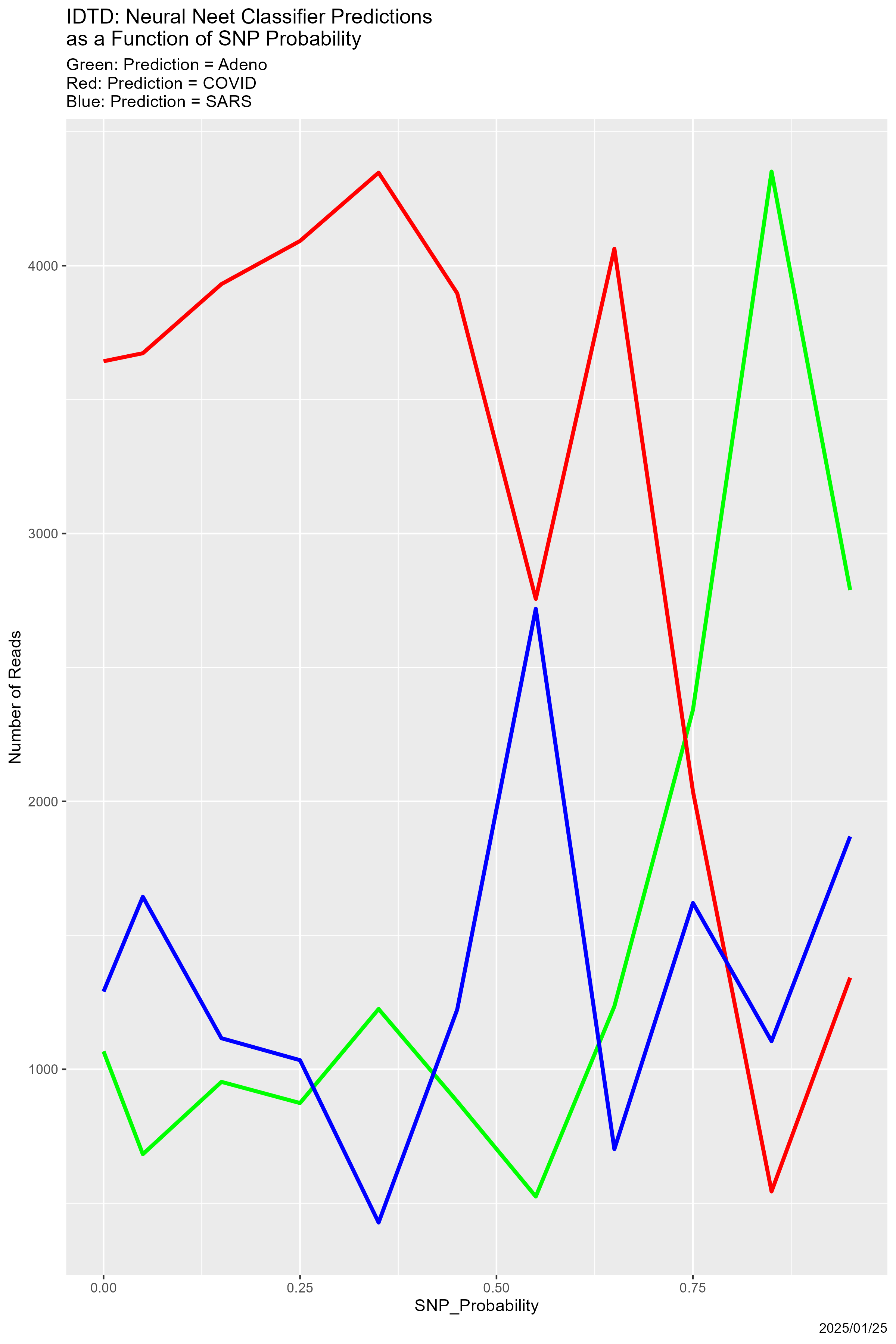}\hspace{.25in}\includegraphics[width = 1.75in]{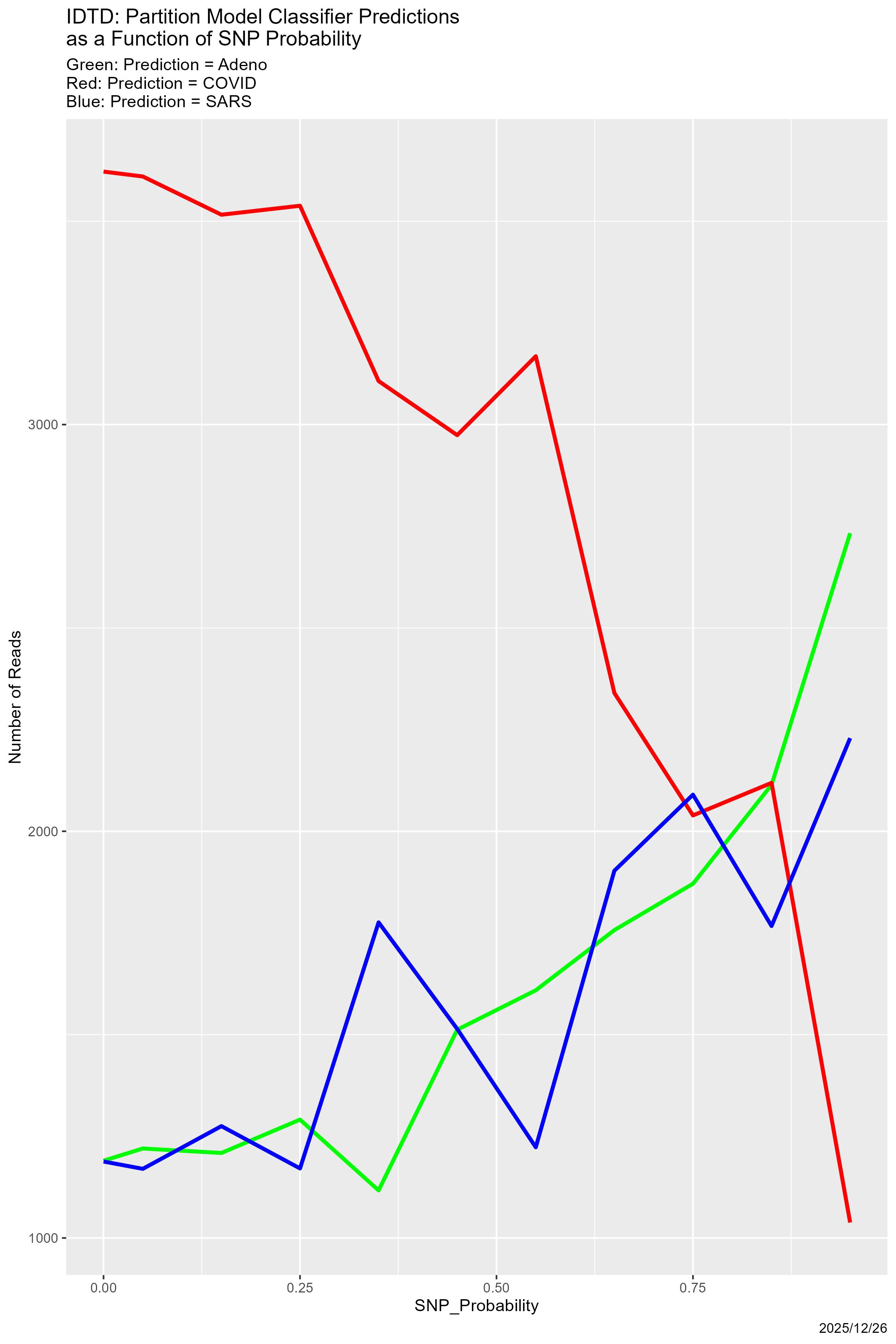}\hspace{.25in}
\includegraphics[width = 1.75in]{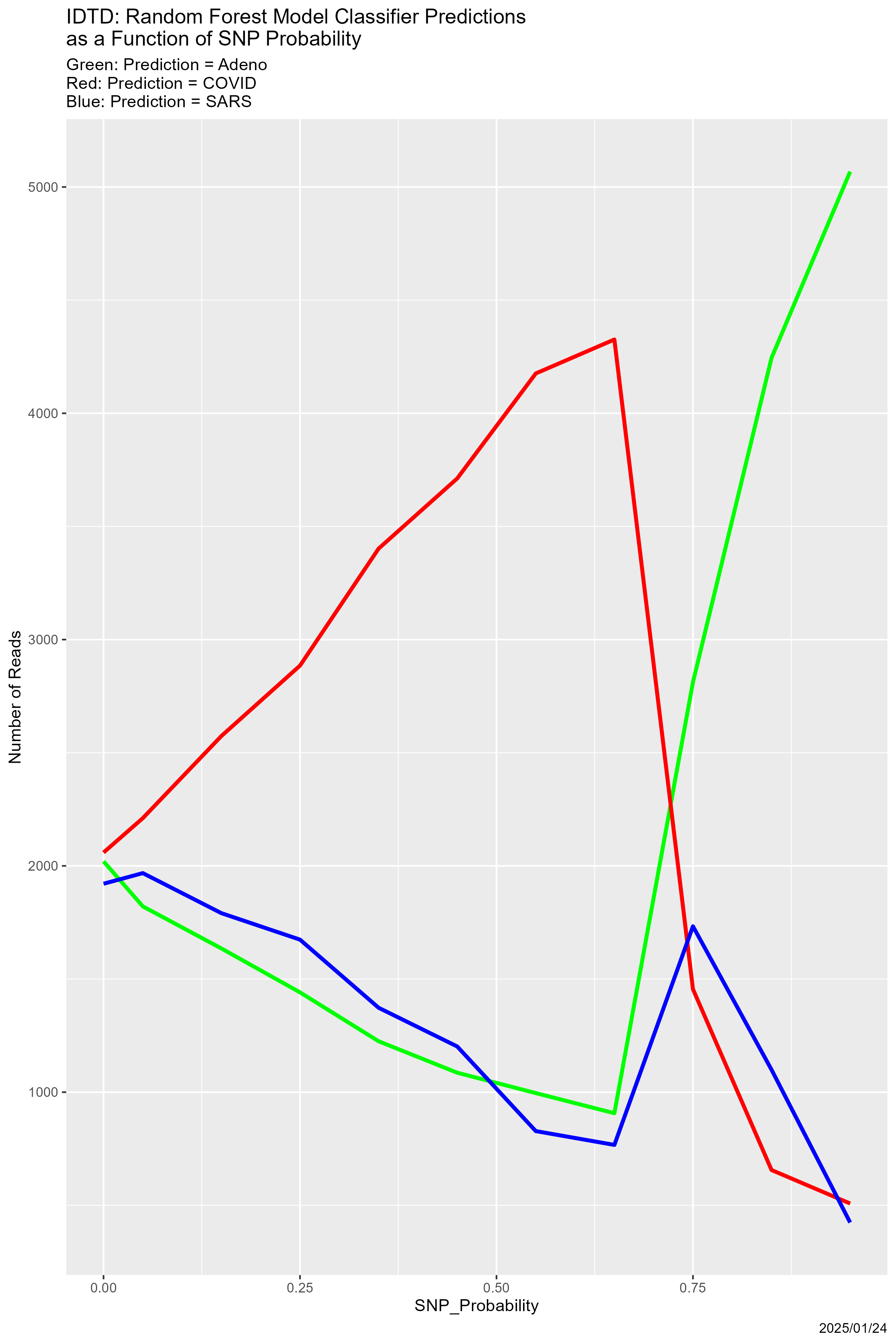}

\vspace{.5in}
\includegraphics[width = 1.75in]{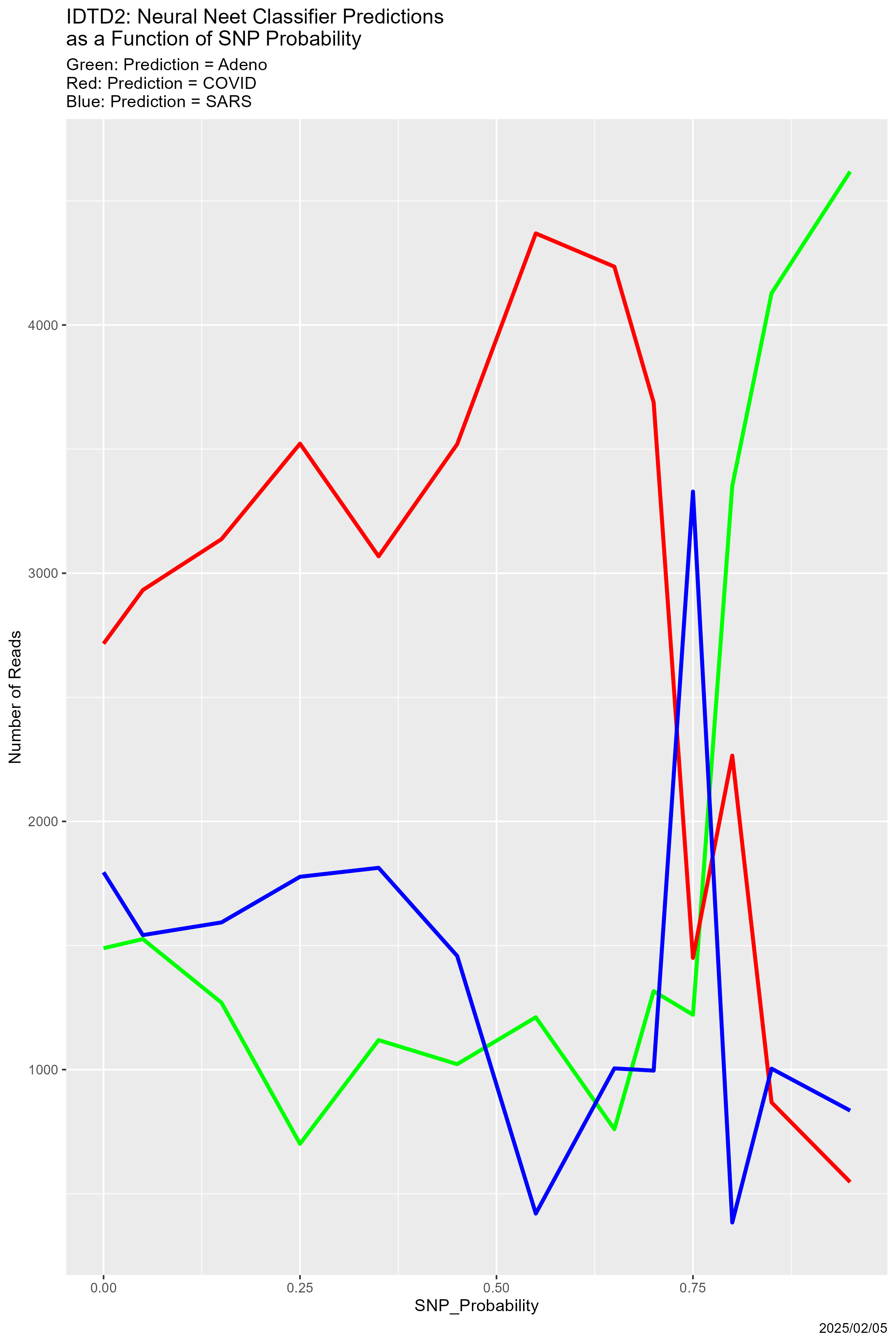}\hspace{.25in}\includegraphics[width = 1.75in]{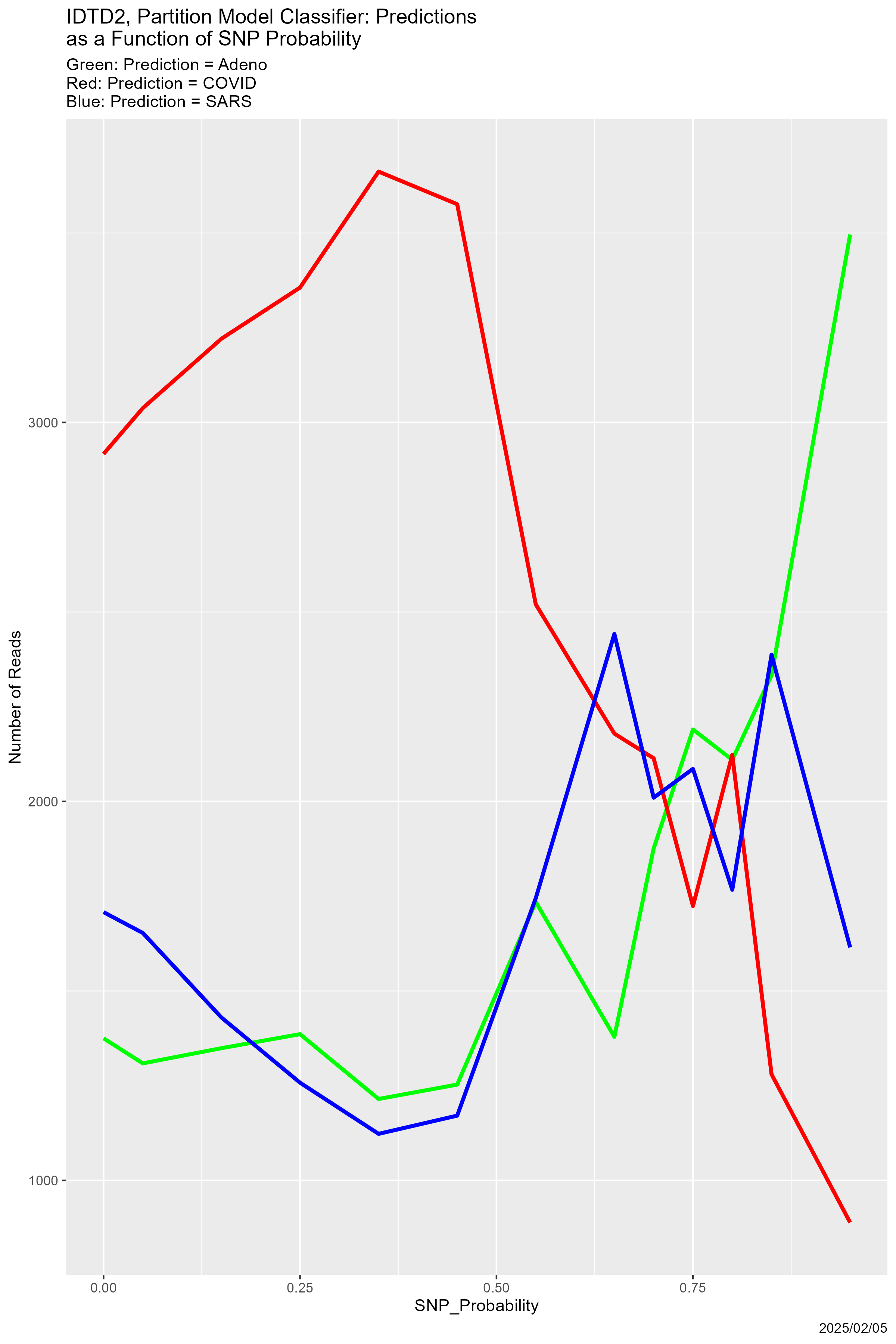}\hspace{.25in}
\includegraphics[width = 1.75in]{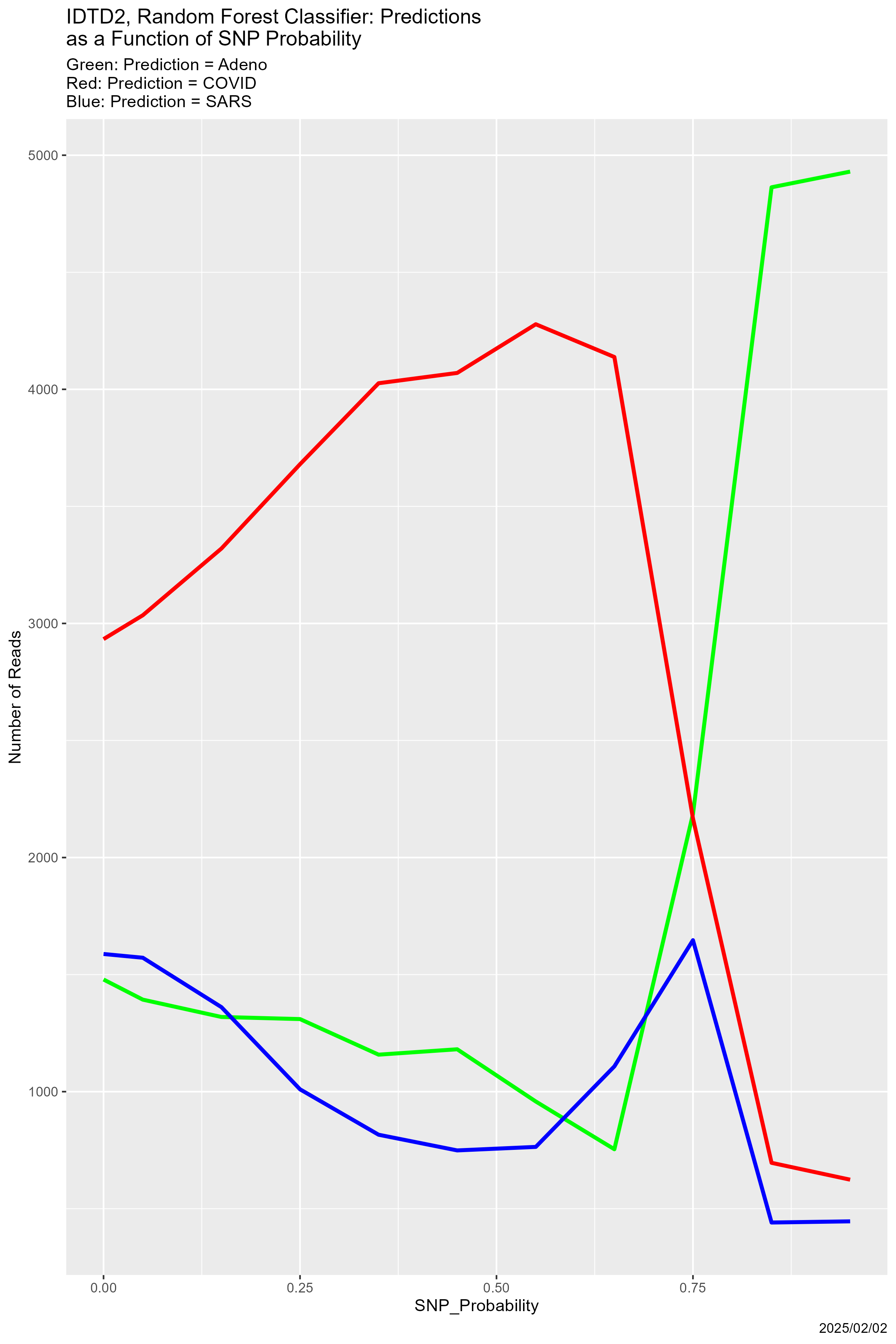}
\caption{Initially degraded training data: classifier predictions as a function of SNP\_Probability. \emph{Top:} SNP\_Probability for initial degradation is 0.9. \emph{Bottom:} SNP\_Probability for initial degradation is 0.5. \emph{Green:} prediction = Adeno. \emph{Red:} prediction = COVID. \emph{Blue:} prediction = SARS.}
\label{fig.idtd-predictions}
\end{figure}

\begin{figure}[ht]
\begin{center}
\includegraphics[width=2.5in]{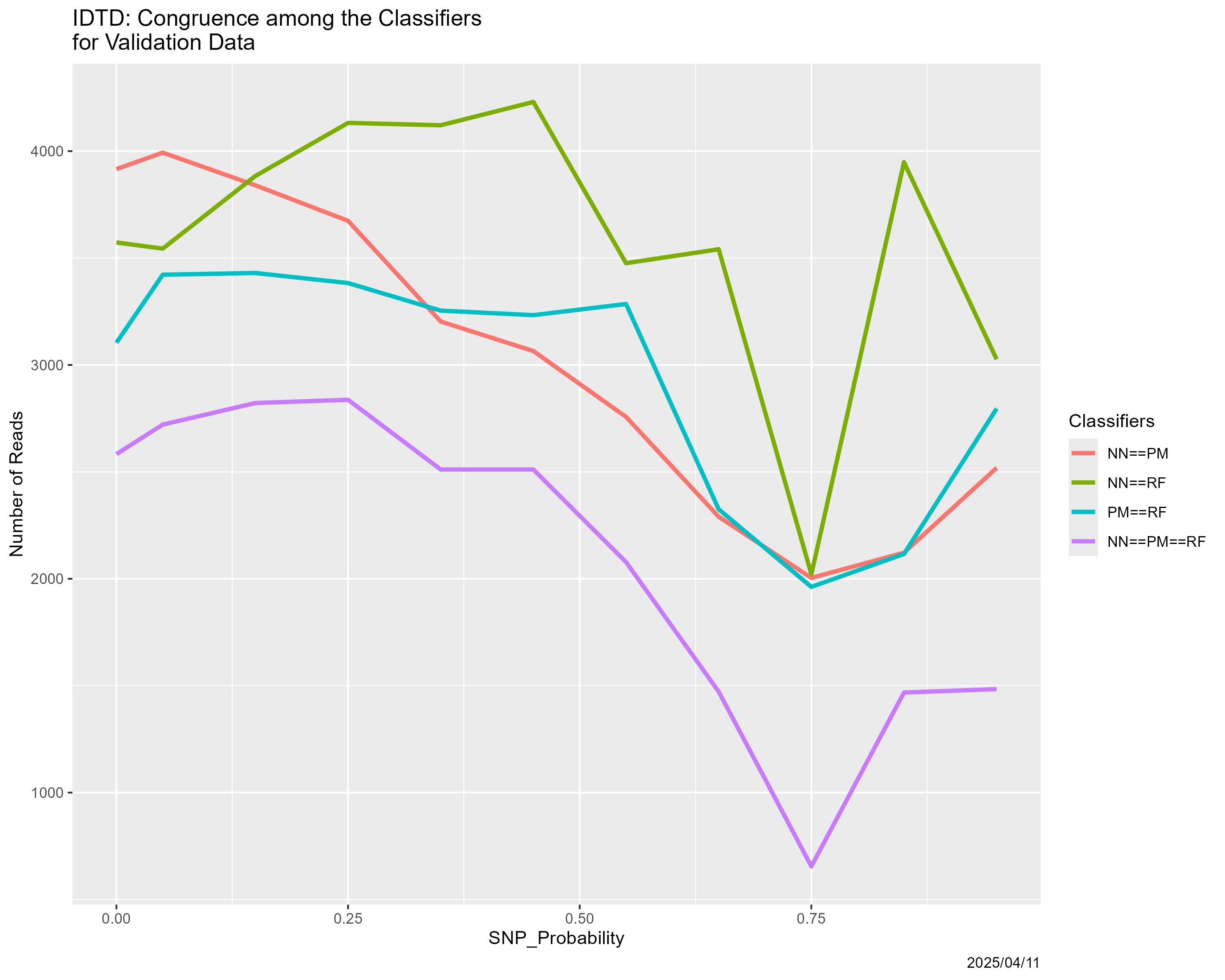}\hspace{.5in}\includegraphics[width=2.5in]{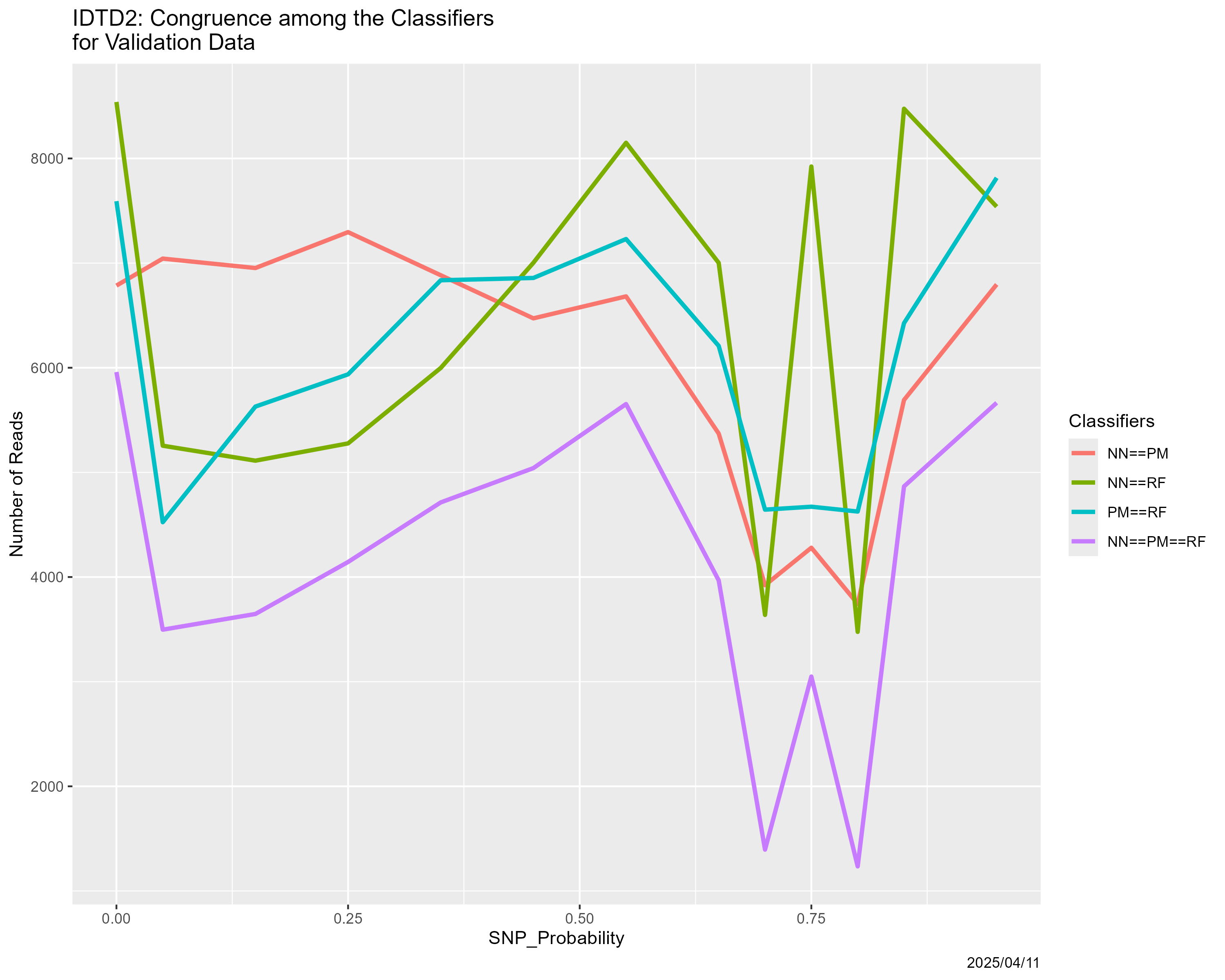}
\end{center}
\caption{Initially degraded training data: classifier congruence as a function of SNP\_Probability. \emph{Left:} SNP\_Probability for initial degradation is 0.9. \emph{Right:} SNP\_Probability for initial degradation is 0.5.}
\label{fig.idtd-congruence}
\end{figure}

We ran two experiments. In the first, the initial training data are the original training dataset \TD\ degraded with SNP\_Probability = 0.9. In the second, the initial training data are the original training data \TD\ degraded with SNP\_Probability = 0.5. Predictions from them are in Figure \ref{fig.idtd-predictions} and congruence for the neural net, partition model and random forest in Figure \ref{fig.idtd-congruence}. 

Figure \ref{fig.idtd-predictions} should be compared to Figure \ref{fig.snp-pred}, ignoring the Bayes classifier results in the upper left-hand panel of the latter. Qualitatively, they are very similar. Unambiguously, the breakdown is still present, and remains at SNP\_Probability = 0.75. The same is true in Figure \ref{fig.idtd-congruence}, whose parallel is Figure \ref{fig.snp-congruence}. That is, the answer to the question in the first paragraph of this section is definitively that breakdown is relative. This is, we believe, as it should be. Whatever the training data are, the classifiers treat it as truth. If truth is wrong, they are wrong. Classifiers cannot be aware of any unobserved ``prior truth.''

\subsubsection{Selective SNP Degradation}\label{subsubsec.snp-selective}
This experiment can be viewed as addressing the interpretations put forth in Section \ref{subsec.snp}. If a random set of reads in \TD\ are selected for degradation, say with probability Sel\_Probability, and each is degraded with probability SNP\_Probability, what happens? This scenario differs from that in Section \ref{subsec.snp} in sense that there, once SNP\_Probability is even moderately large, all reads are altered, whereas here, the degradation is confined to a subset of reads.

\begin{figure}[ht]
\centering
\includegraphics[width=3in]{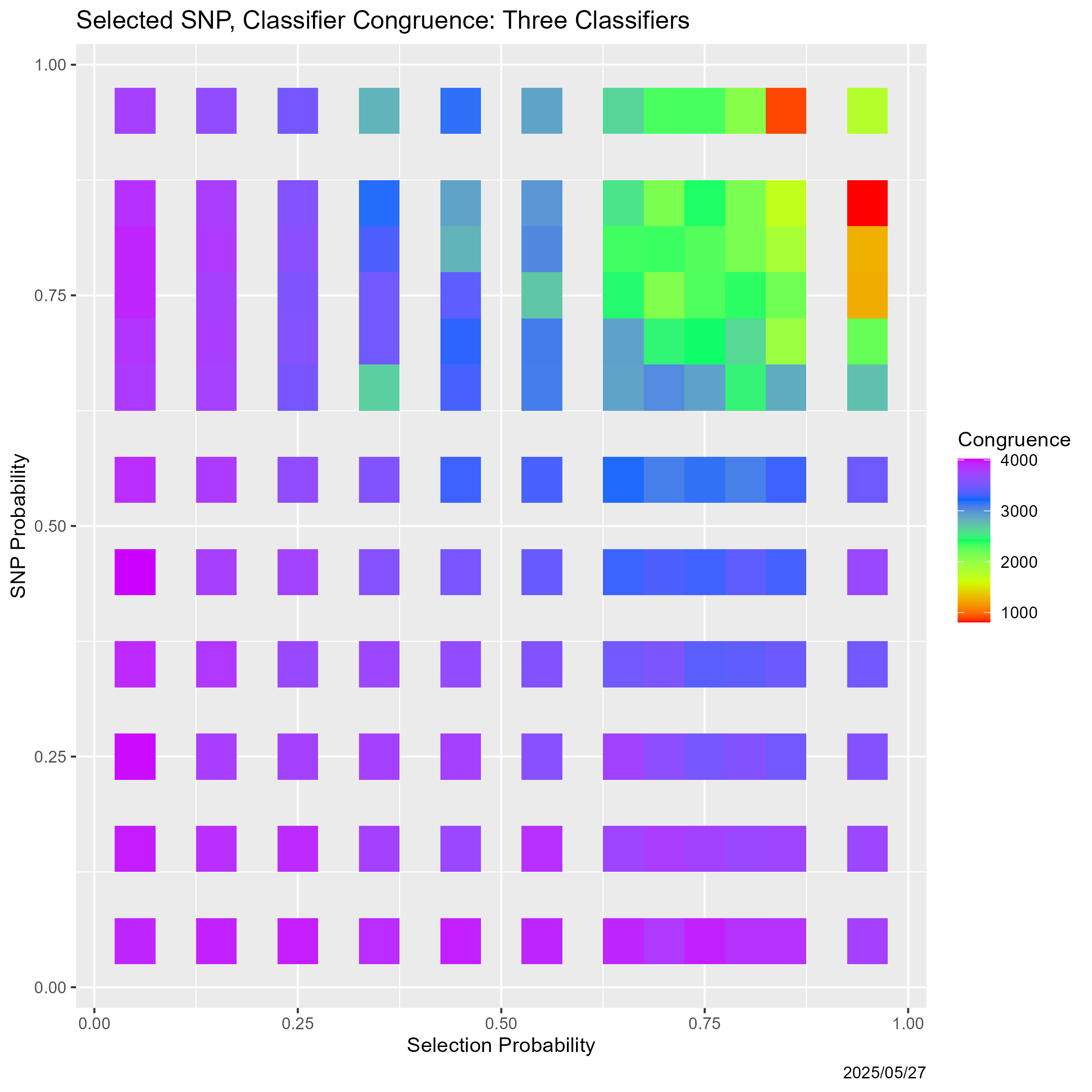}\hspace{.5in}
\includegraphics[width=3in]{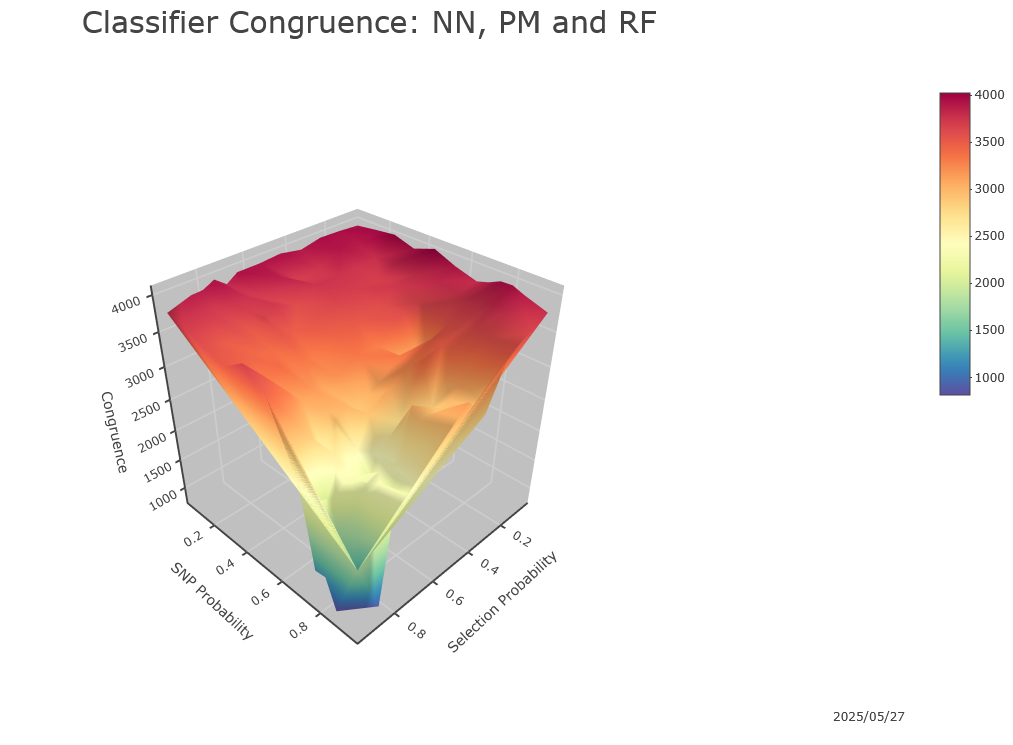}

\caption{Selective SNP degradation: 3-way classifier congruence as a function of Sel\_Probability and SNP\_Probability. \emph{Left:} heatmap. \emph{Right:} 3D surface.}
\label{fig.selective-congruence}
\end{figure}

So there are now two parameters, Sel\_Probability and SNP\_Probability, both ranging over 
$$
\{0.05, 0.15, 0.25, 0.35, 0.45, 0.55, 0.65, 0.70, 0.75, 0.80, 0.85, 0.95\}
$$ 
in a full factorial design. For brevity, we consider only the three-way congruence among the neural net, partition model and random forest. Figure \ref{fig.selective-congruence} contains both a heatmap and a 3-dimensional surface showing it. The most striking feature is that breakdown persists, but only when both probabilities reach 0.75.  There is degeneration when one is small and the other large (the upper left-hand and lower right-hand corners in the heatmap), but it is relatively minor. There is also asymmetry between Sel\_Probability and SNP\_Probability, but it is subtle and requires further investigation before we assert its existence with confidence. We find only limited evidence of a return to congruence following breakdown, but we suspect that this is due to our not including probabilities greater than 0.95.

Can congruence be modeled as a function of Sel\_Probability and SNP\_Probability? The minor asymmetry notwithstanding, by a symmetric function, from Figure \ref{fig.selective-congruence}, a plausible candidate is the minimum of the two, and indeed the adjusted $R^2$ for the model is 0.8256. However, which is crucial, the model dramatically overpredicts congruence once the minimum exceeds 0.75. Put differently, it cannot predict the post-breakdown decline in congruence.

\subsubsection{Protecting Some Read Sources}\label{subsubsection.two-genome}
Here we explore the effect of exempting reads from particular sources from degradation. One motivation is to accommodate laboratory effects, in the sense that the quality of gene sequence data may depend on the laboratory that generated them, a phenomenon that is demonstrably present in \cite{markovstructure-2021}.

Such protection can be done in six ways, of course: protect reads whose source is one genome, or protect all reads except those whose source is one genome. The central question is whether the effects, especially prediction, are localized in the same way. Nor is ``full protection'' necessary: the experiment in Section \ref{subsec.snp} could instead have three SNP\_Probabilities, one for each genome. This is a direction we begin to explore in Section \ref{subsubsec.mixed}.  

\begin{figure}[ht]
\centering
\includegraphics[width = 1.75in]{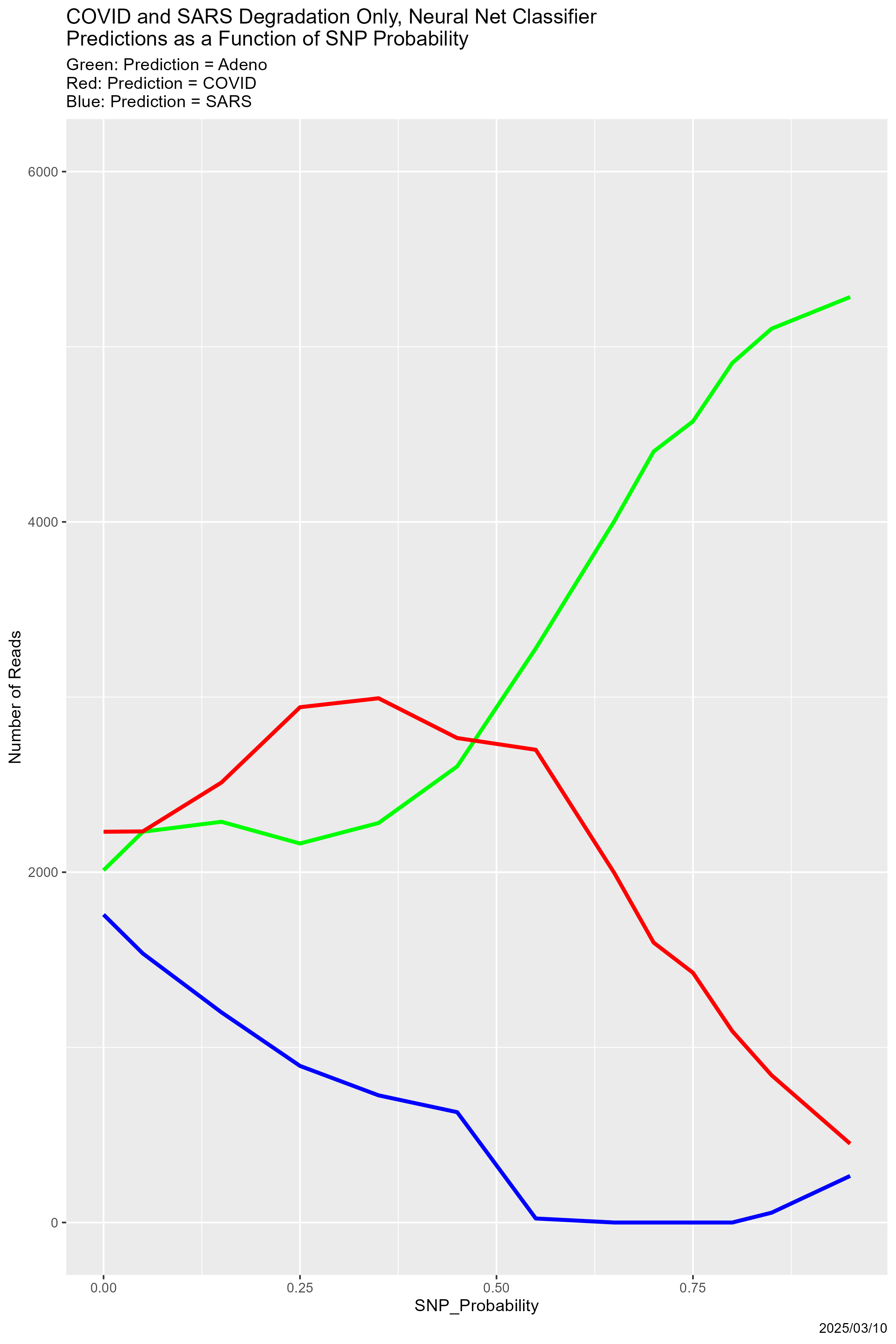}\hspace{.25in}
\includegraphics[width = 1.75in]{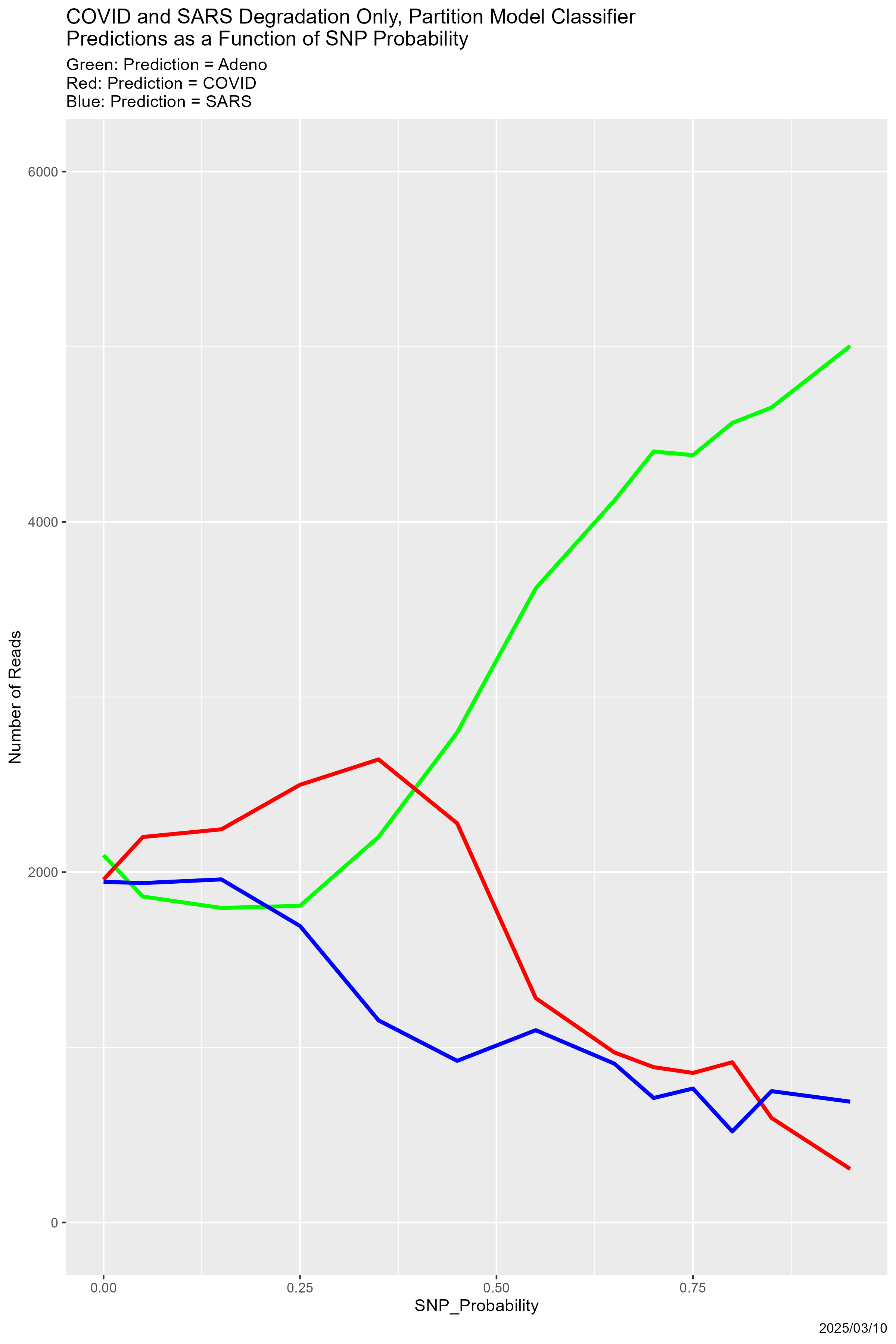}\hspace{.25in}
\includegraphics[width = 1.75in]{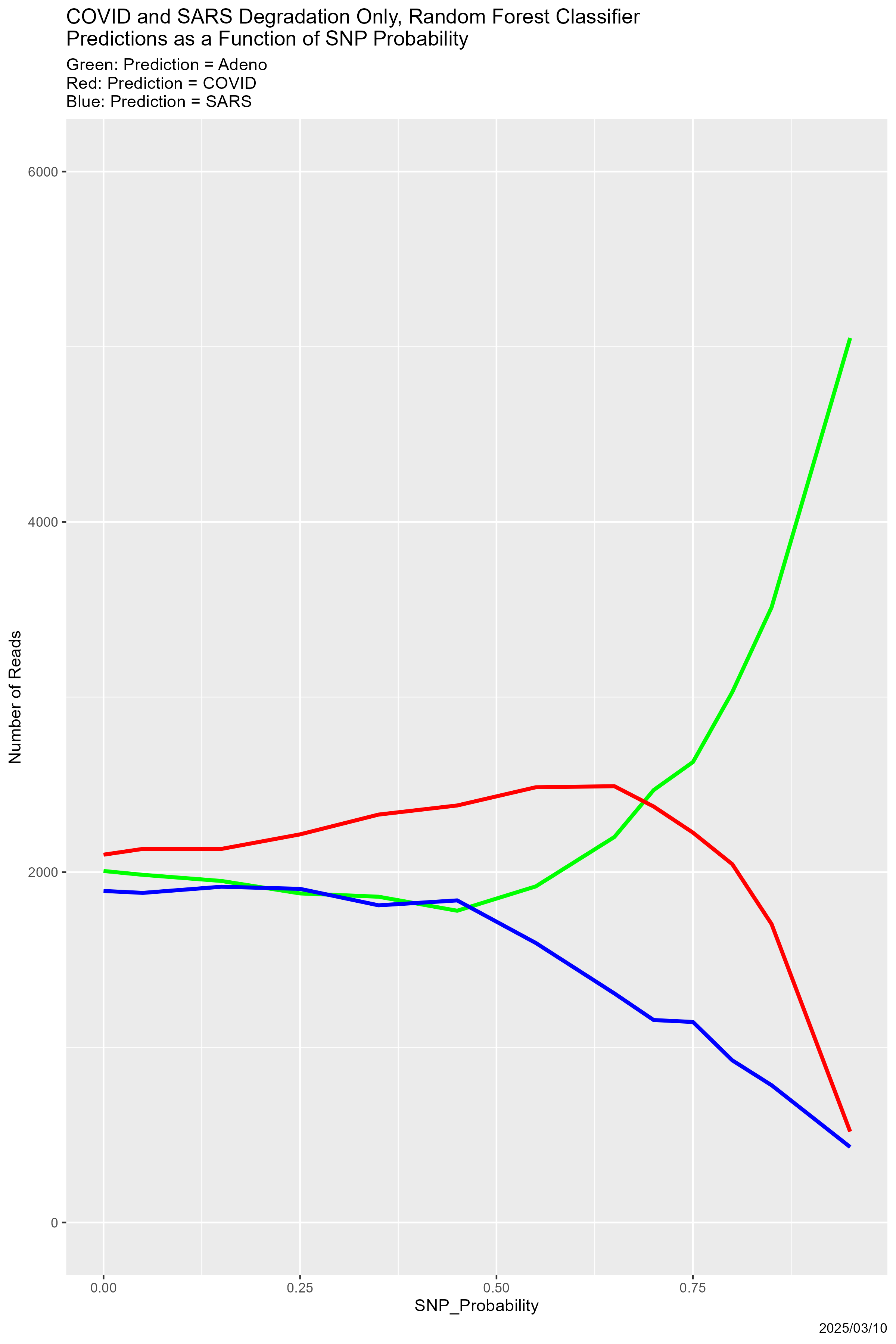}
\caption{SNP degradation not applied to Adeno reads: classifier predictions as a function of SNP\_Probability. \emph{Left:} neural net. \emph{Center:} partition model. \emph{Right:} random forest. \emph{Green:} prediction = Adeno. \emph{Red:} prediction = COVID. \emph{Blue:} prediction = SARS.}
\label{fig.noAdeno-predictions}
\end{figure}

\begin{figure}[ht]
\centering
\includegraphics[width = 1.75in]{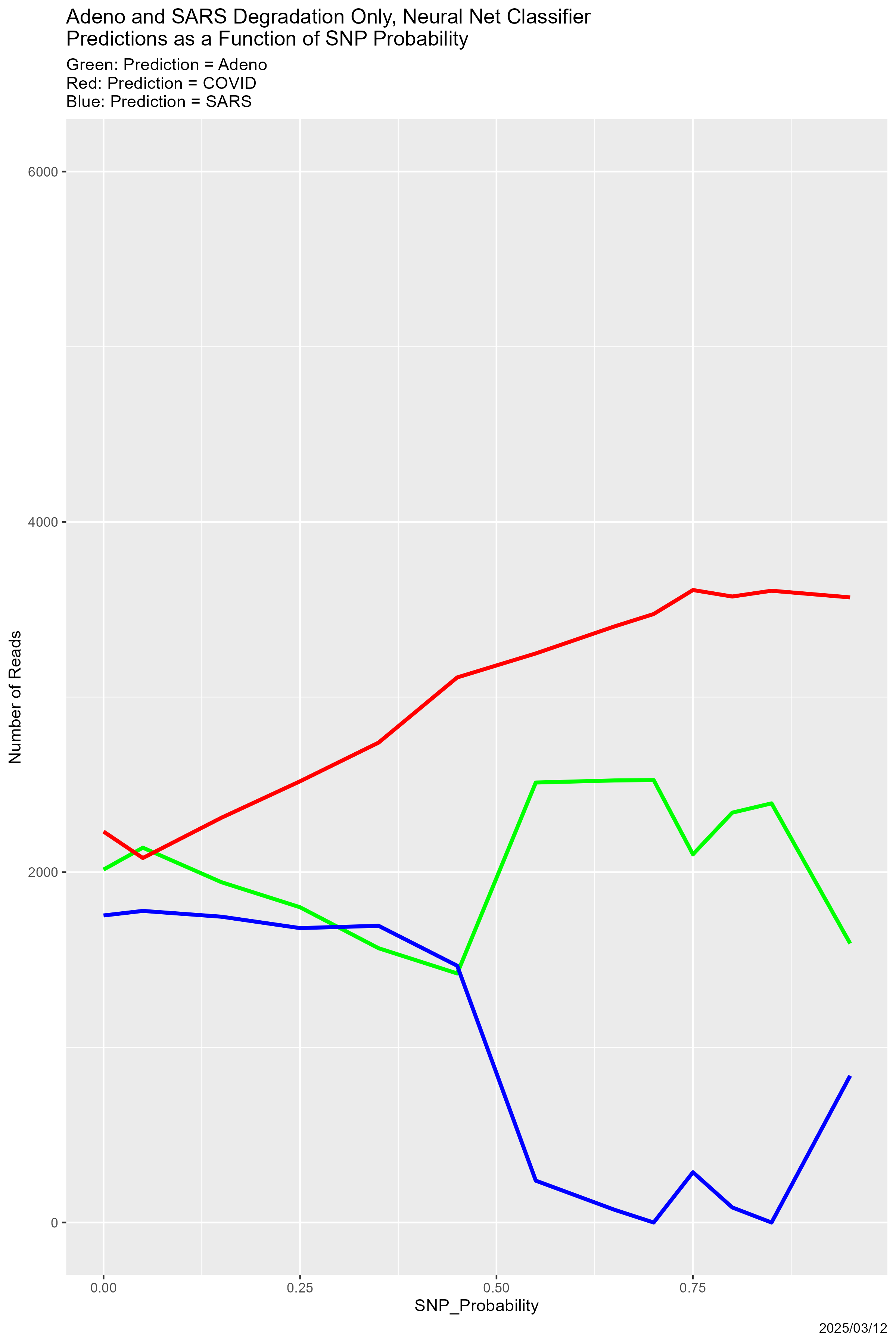}\hspace{.25in}
\includegraphics[width = 1.75in]{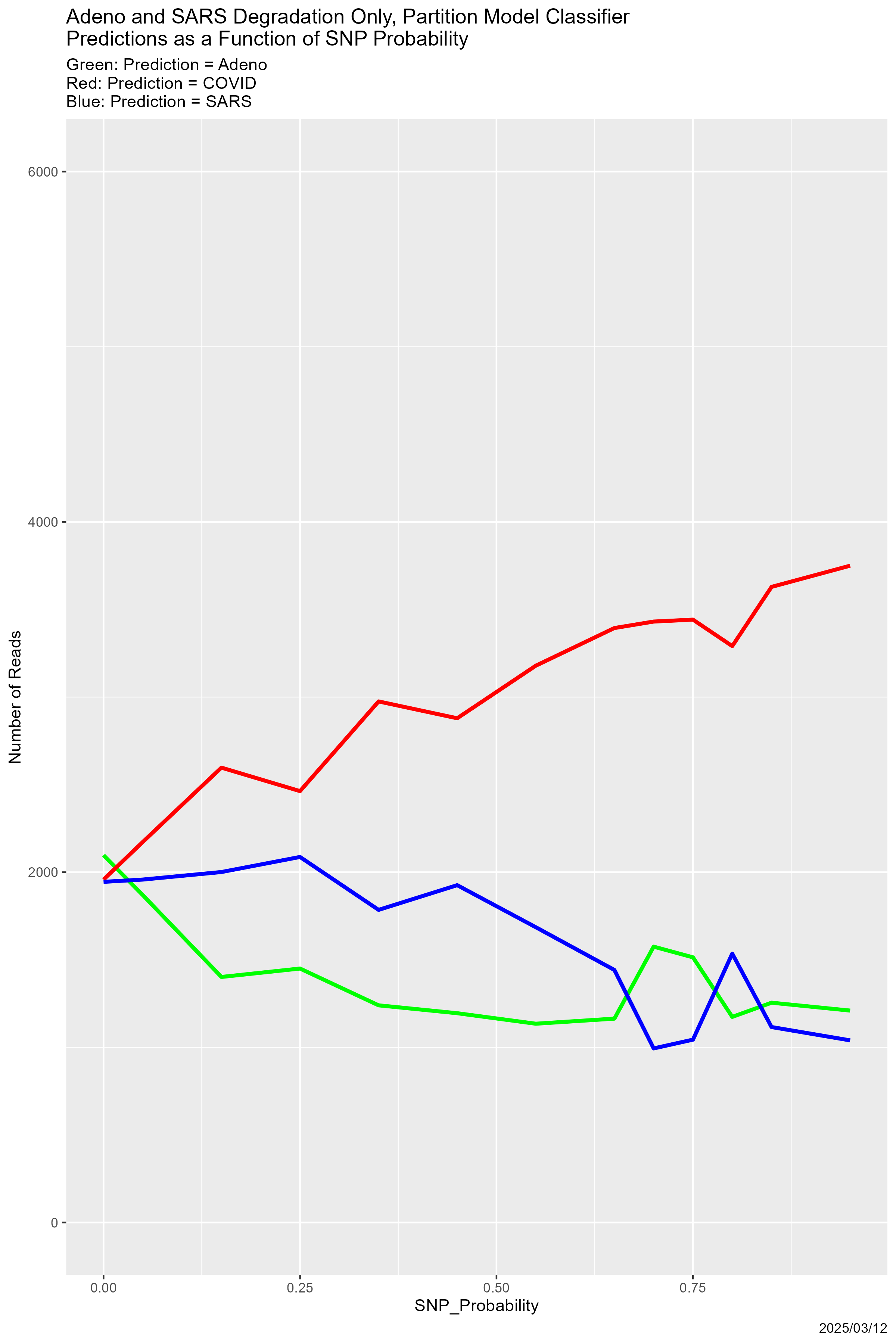}\hspace{.25in}
\includegraphics[width = 1.75in]{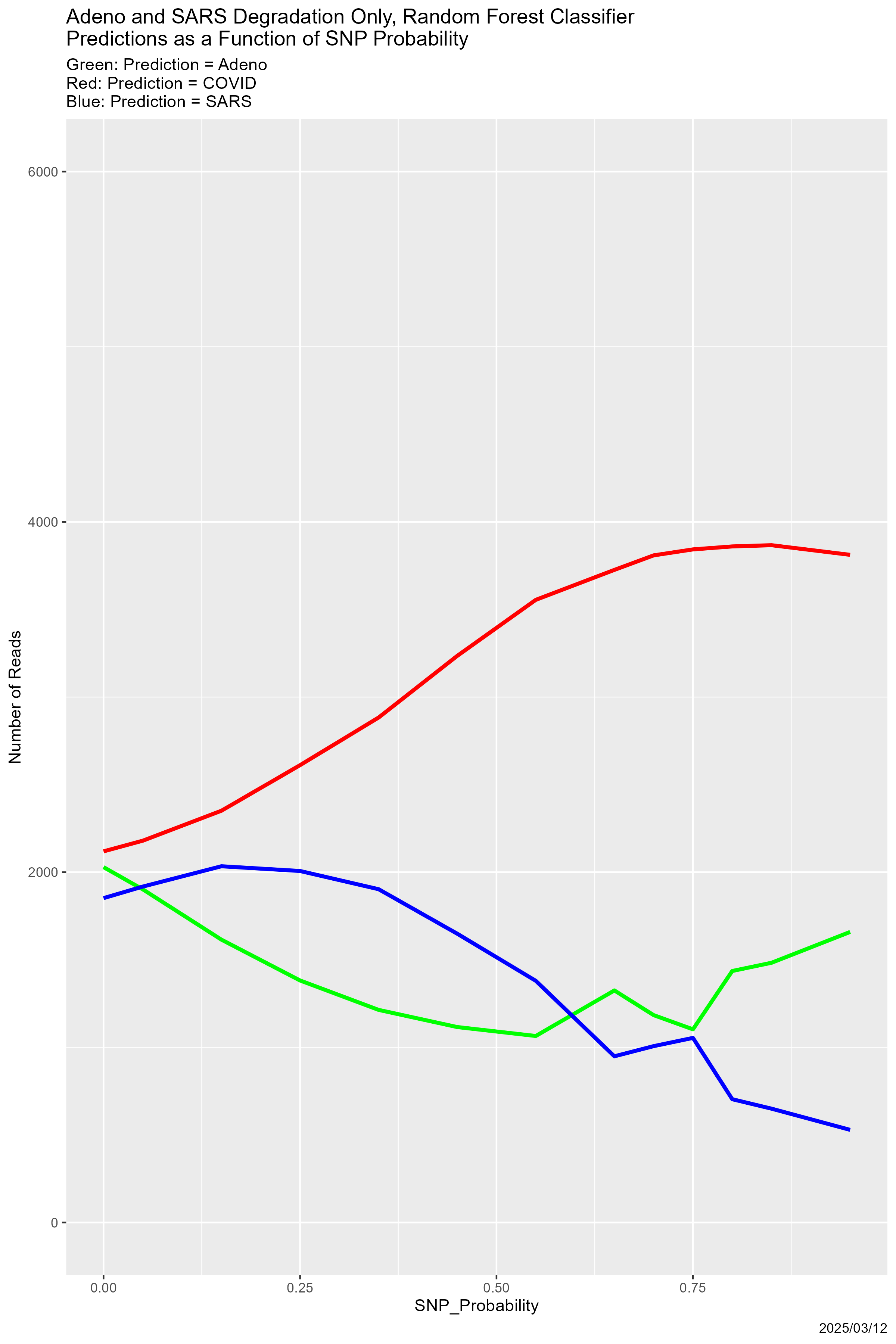}
\caption{SNP degradation not applied to COVID reads: classifier predictions as a function of SNP\_Probability. \emph{Left:} neural net. \emph{Center:} partition model. \emph{Right:} random forest. \emph{Green:} prediction = Adeno. \emph{Red:} prediction = COVID. \emph{Blue:} prediction = SARS.}
\label{fig.noCOVID-predictions}
\end{figure}

\begin{figure}[ht]
\centering
\includegraphics[width = 1.75in]{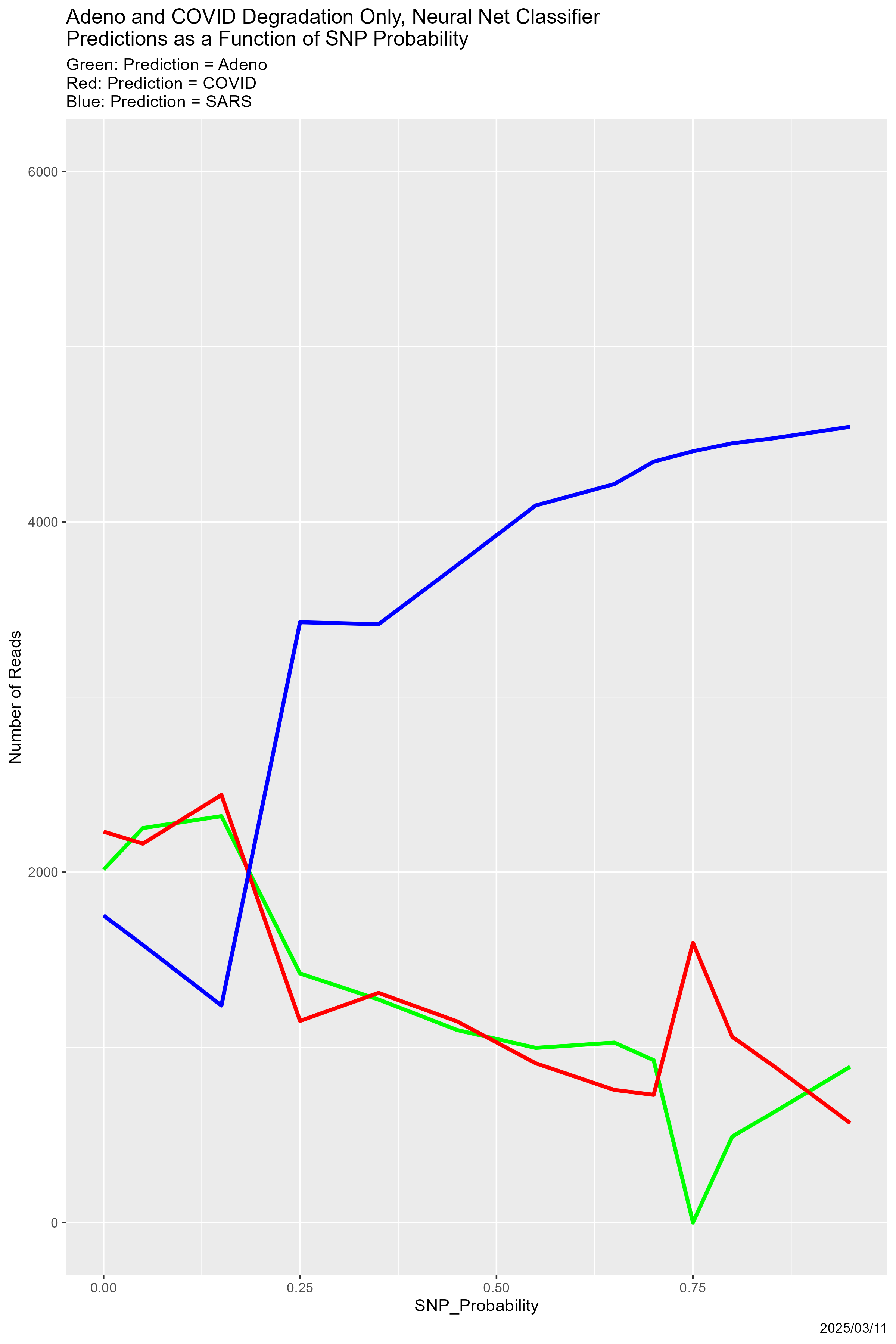}\hspace{.25in}
\includegraphics[width = 1.75in]{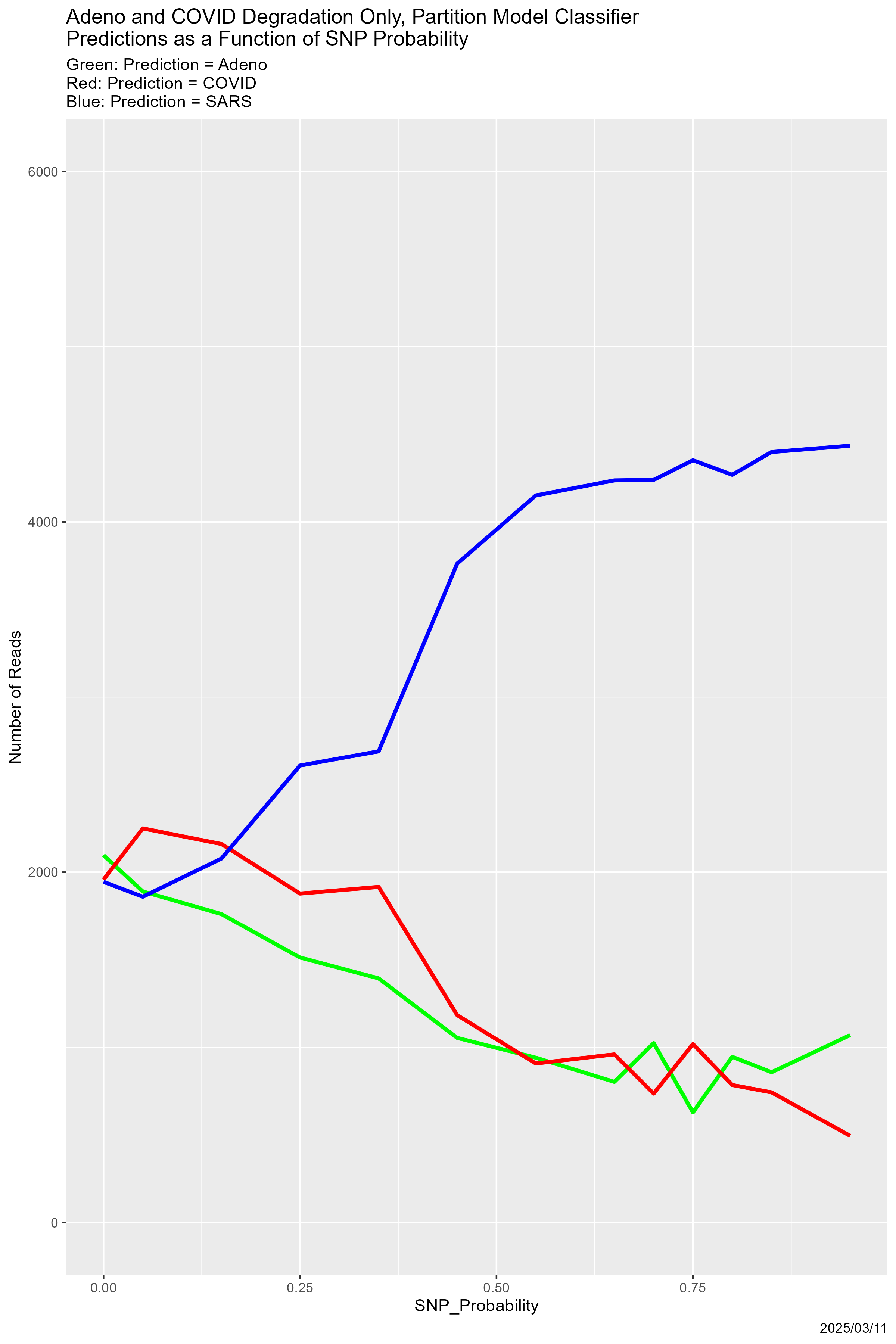}\hspace{.25in}
\includegraphics[width = 1.75in]{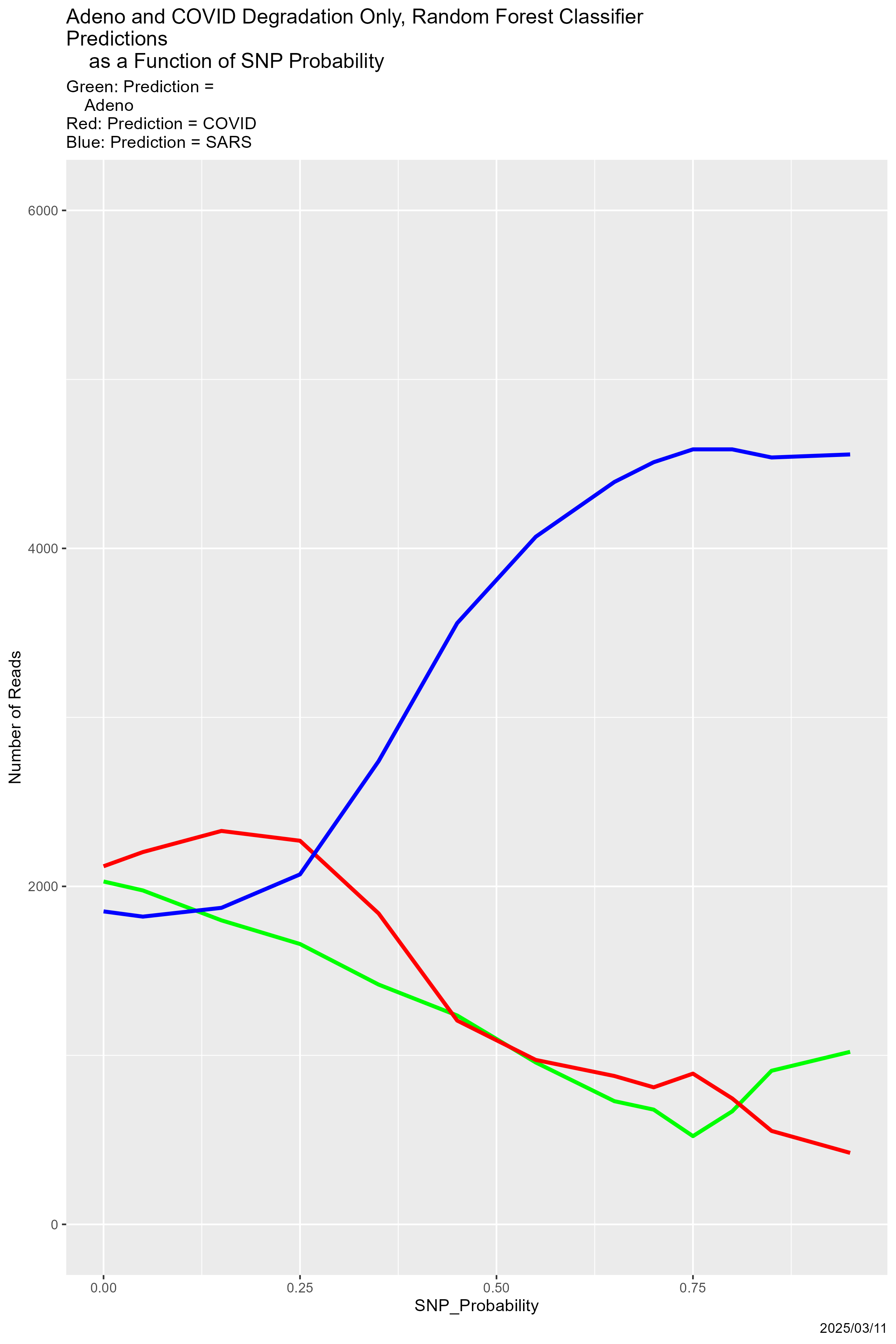}
\caption{SNP degradation not applied to SARS reads: classifier predictions as a function of SNP\_Probability. \emph{Left:} neural net. \emph{Center:} partition model. \emph{Right:} random forest. \emph{Green:} prediction = Adeno. \emph{Red:} prediction = COVID. \emph{Blue:} prediction = SARS.}
\label{fig.noSARS-predictions}
\end{figure}

\begin{figure}[ht]
\centering
\includegraphics[width = 1.75in]{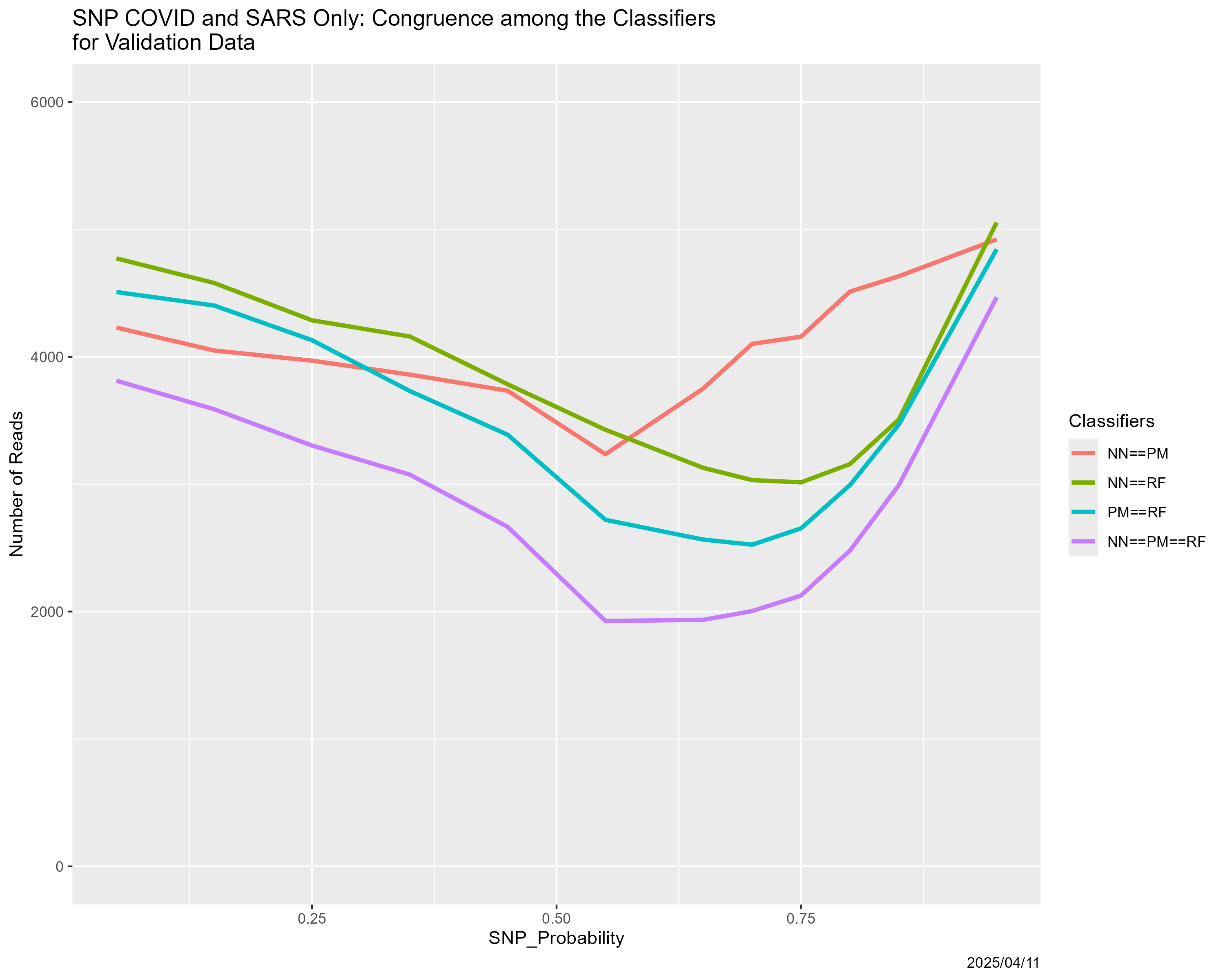}\hspace{.25in}
\includegraphics[width = 1.75in]{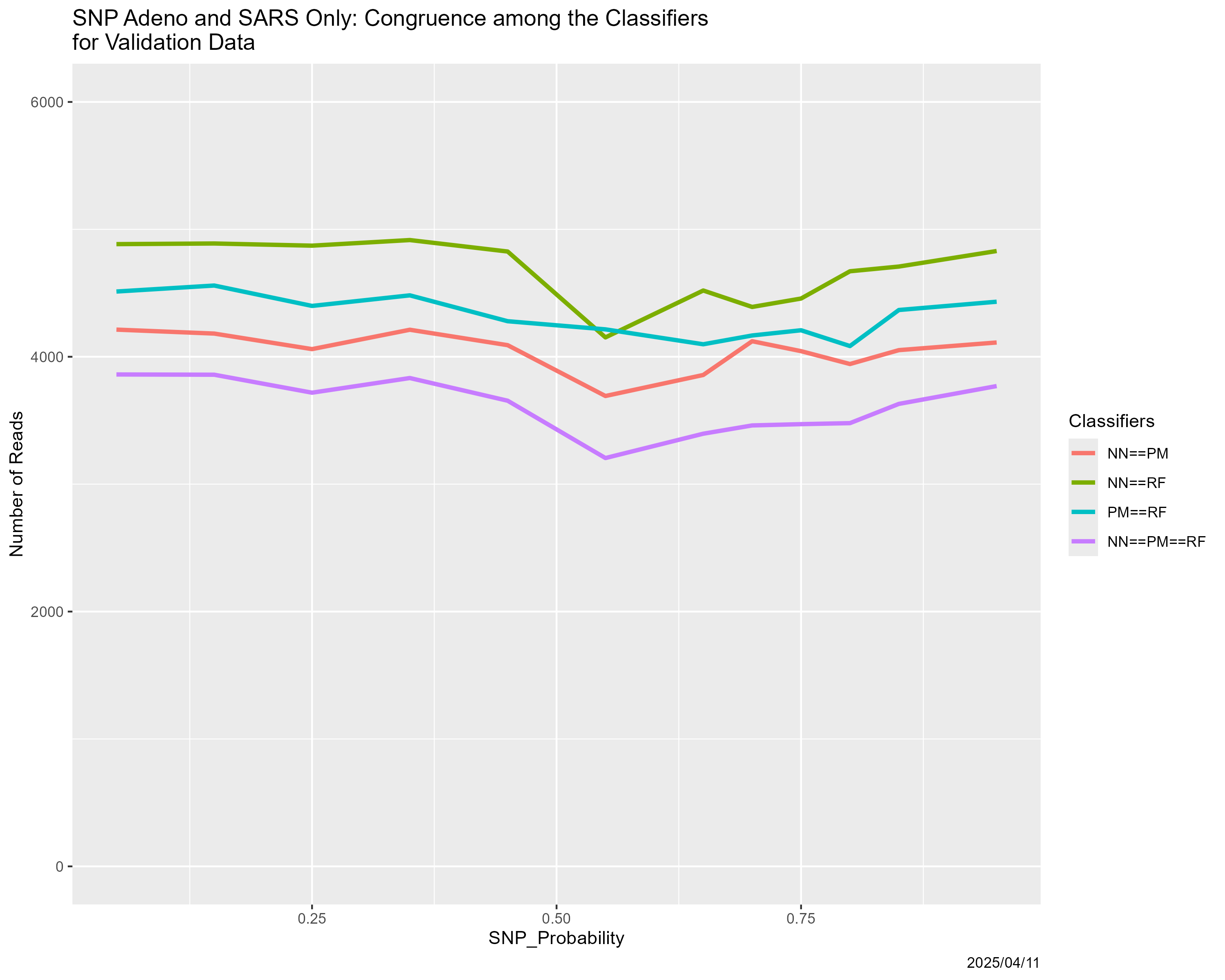}\hspace{.25in}
\includegraphics[width = 1.75in]{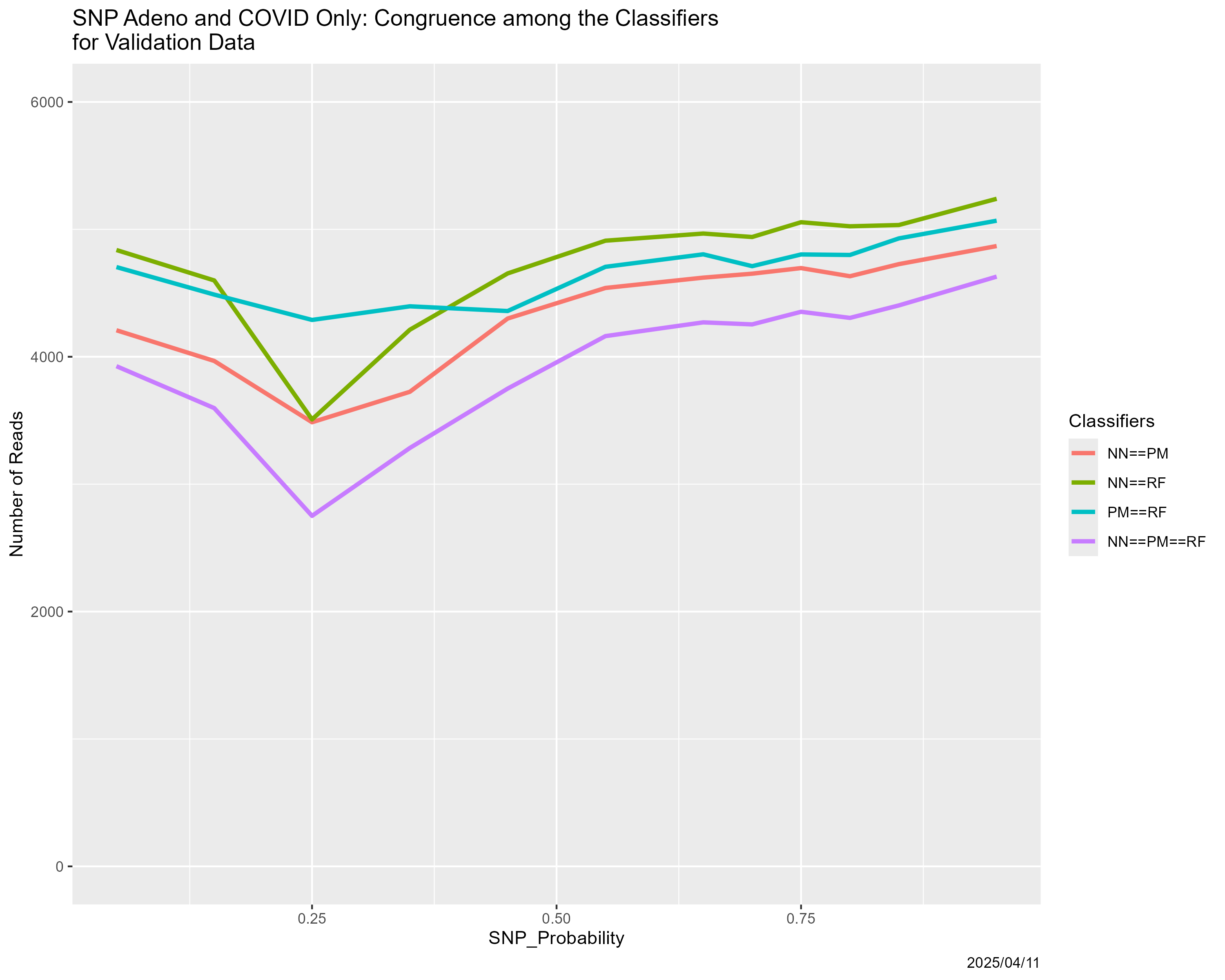}
\caption{SNP degradation not applied to one set of reads: classifier congruence as a function of SNP\_Probability. \emph{Left:} not applied to Adeno reads in training data. \emph{Center:} not applied to COVID reads in training data. \emph{Right:} not applied to SARS reads in training data.}
\label{fig.one-missing-congruence}
\end{figure}

Figures \ref{fig.noAdeno-predictions}--\ref{fig.noSARS-predictions} show predictions from the ``protect one genome'' scenario, with congruence in Figure \ref{fig.one-missing-congruence}. Here the story differs from previous (and subsequent) experiments. In none of Figures \ref{fig.noAdeno-predictions}--\ref{fig.noSARS-predictions} is there more than mild evidence of breakdown, although what evidence there is, for example in Figure \ref{fig.noSARS-predictions}, continues to point to SNP\_Probability = 0.75 as the location. The stronger message is that as reads from two of the genomes are degraded, predictions all become the undegraded genome. This is true for all three genomes and all three classifiers. This is sobering: when there is \emph{differential data quality} in \TD, predictions may be incorrectly ``forced'' onto higher quality results because the real validation data look more like high quality training data than like noise.

\begin{figure}[ht]
\centering
\includegraphics[width = 1.75in]{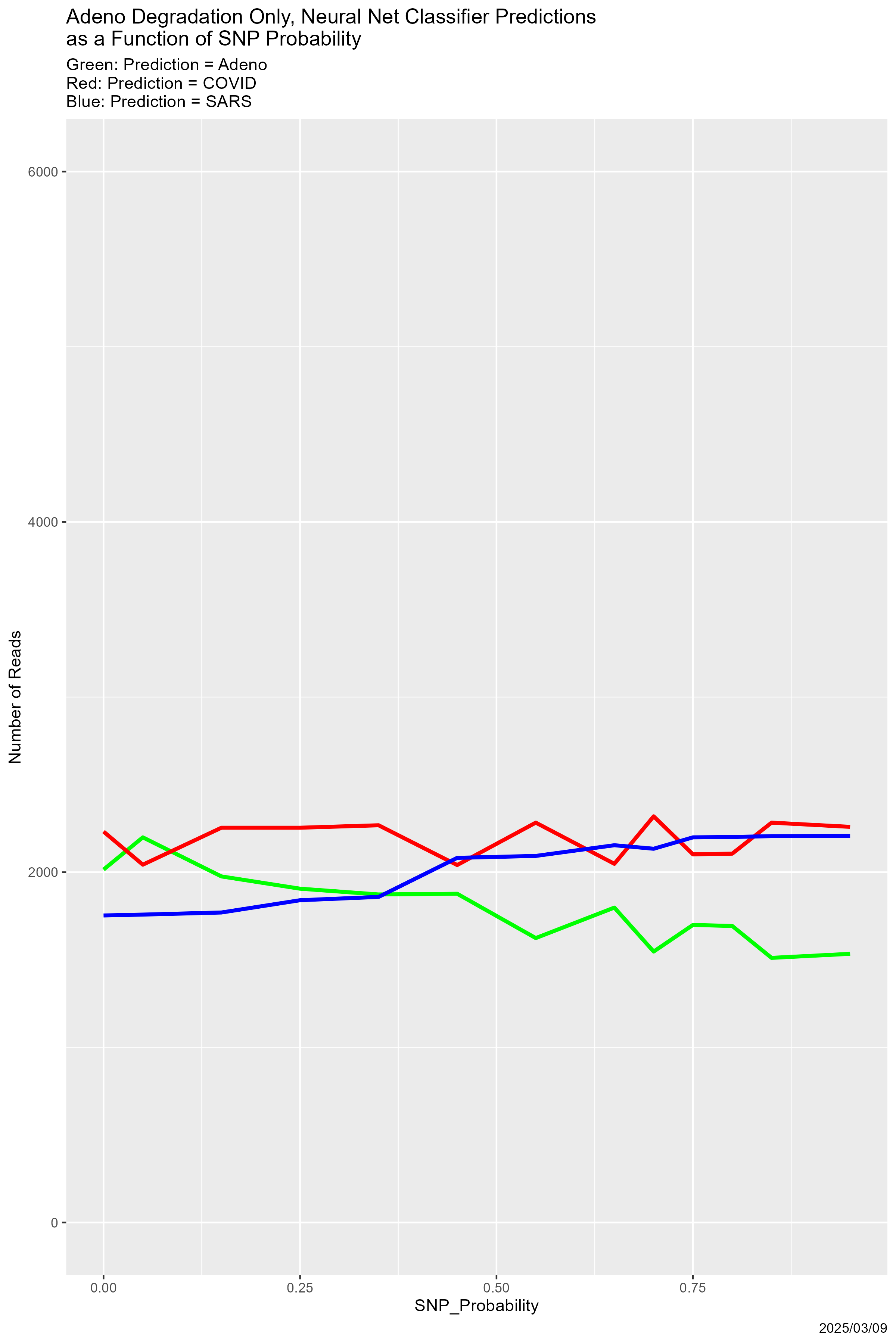}\hspace{.25in}
\includegraphics[width = 1.75in]{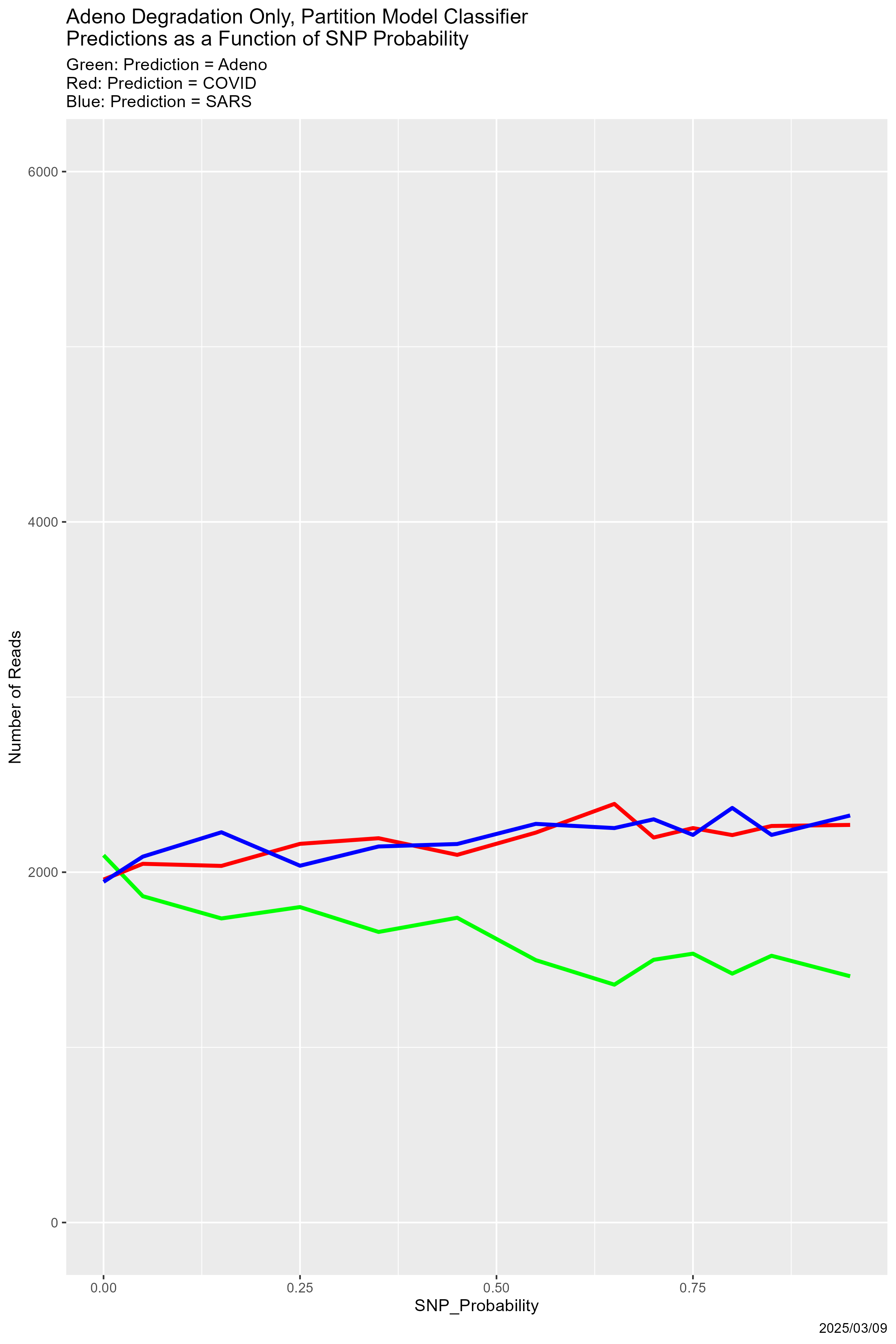}\hspace{.25in}
\includegraphics[width = 1.75in]{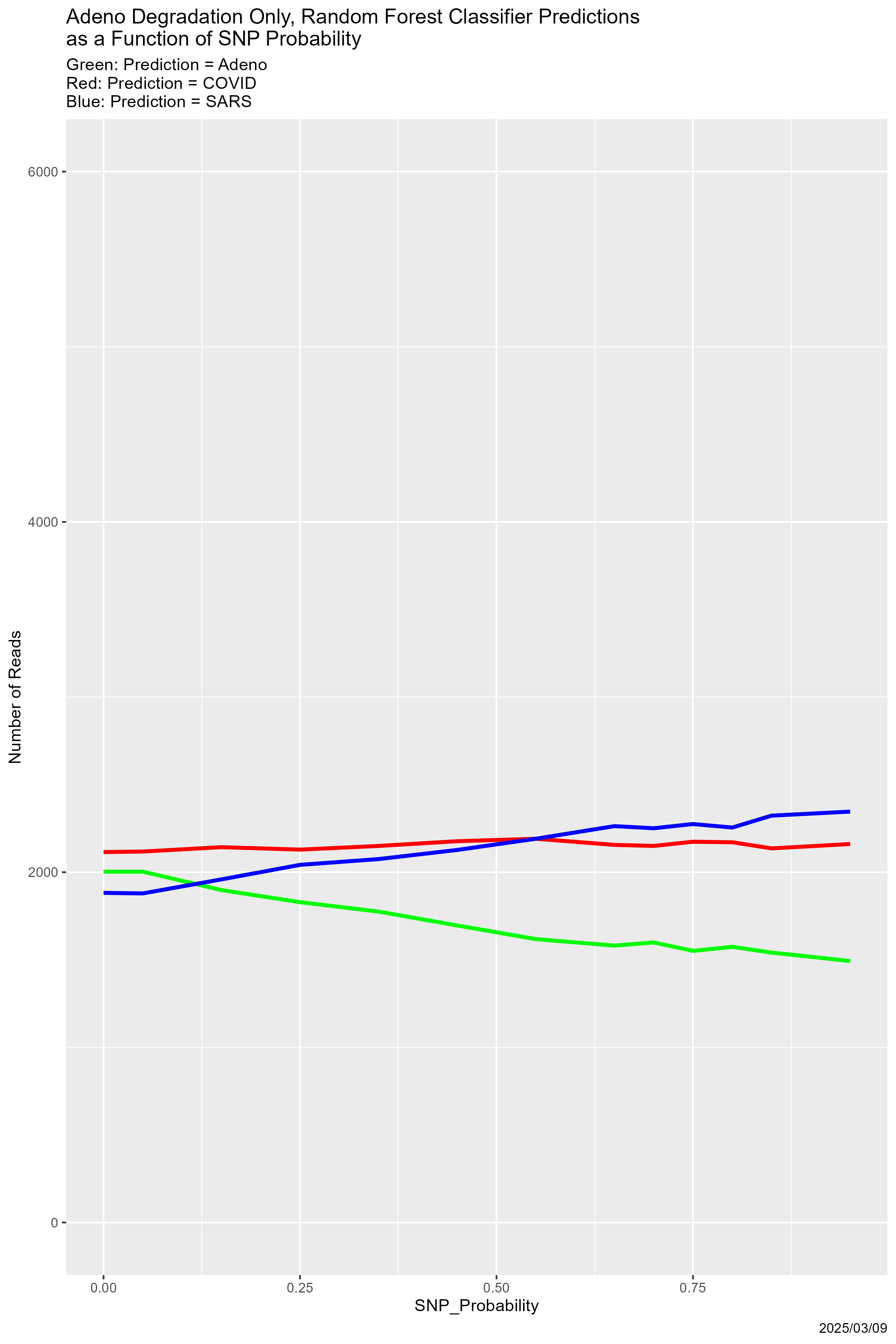}
\caption{SNP degradation applied only to Adeno reads: classifier predictions as a function of SNP\_Probability. \emph{Left:} neural net. \emph{Center:} partition model. \emph{Right:} random forest. \emph{Green:} prediction = Adeno. \emph{Red:} prediction = COVID. \emph{Blue:} prediction = SARS.}
\label{fig.AdenoOnly-predictions}
\end{figure}

\begin{figure}[ht]
\centering
\includegraphics[width = 1.75in]{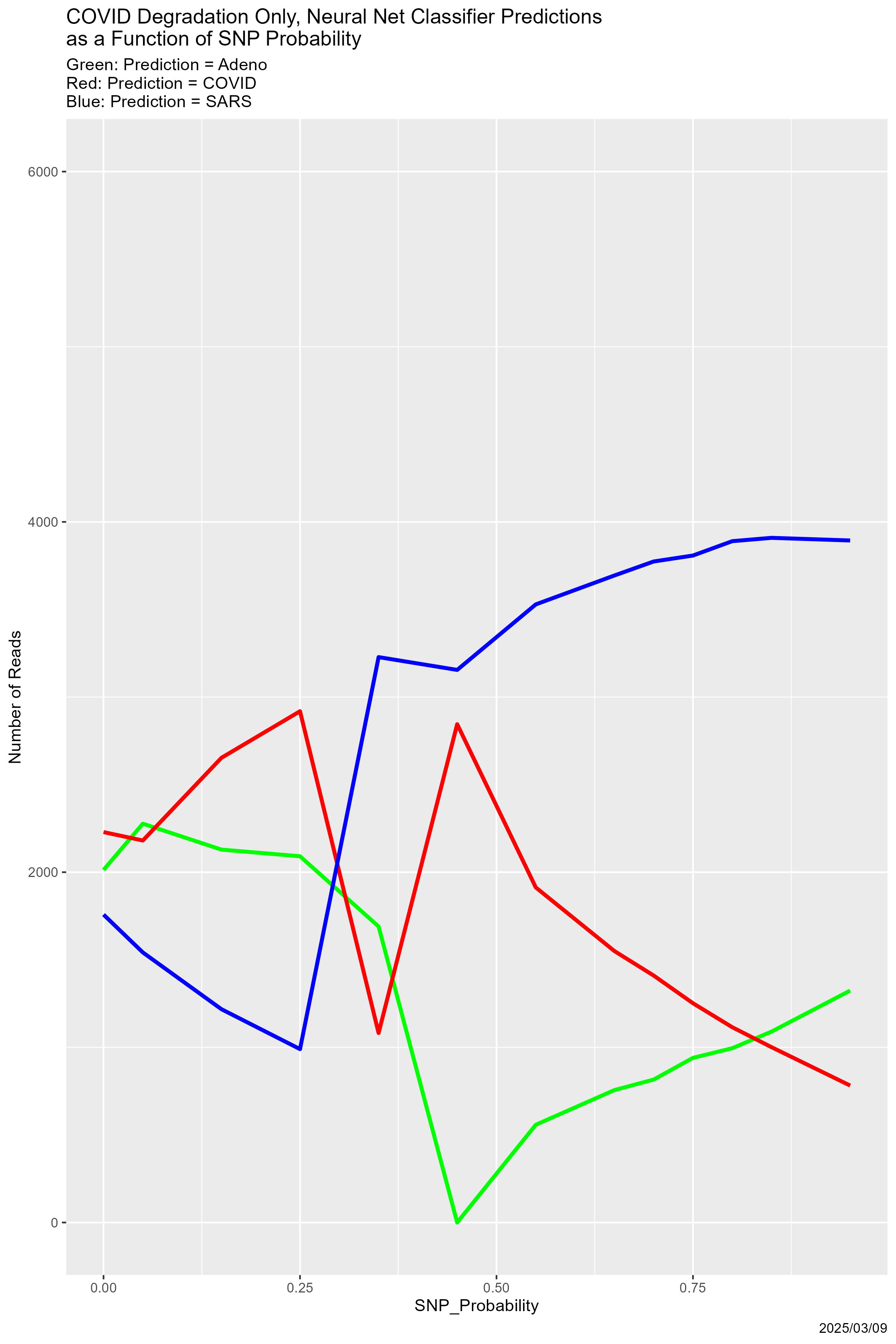}\hspace{.25in}
\includegraphics[width = 1.75in]{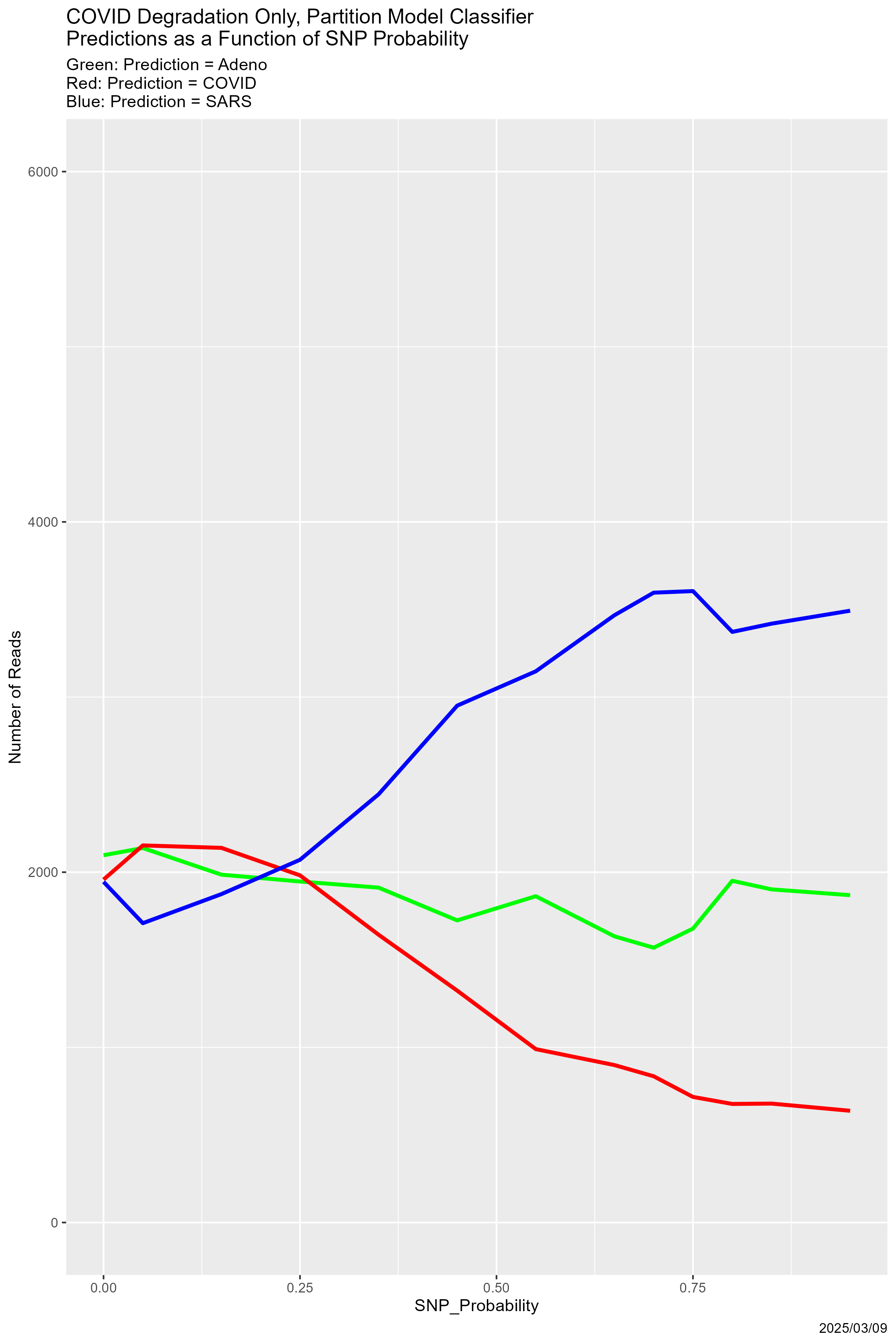}\hspace{.25in}
\includegraphics[width = 1.75in]{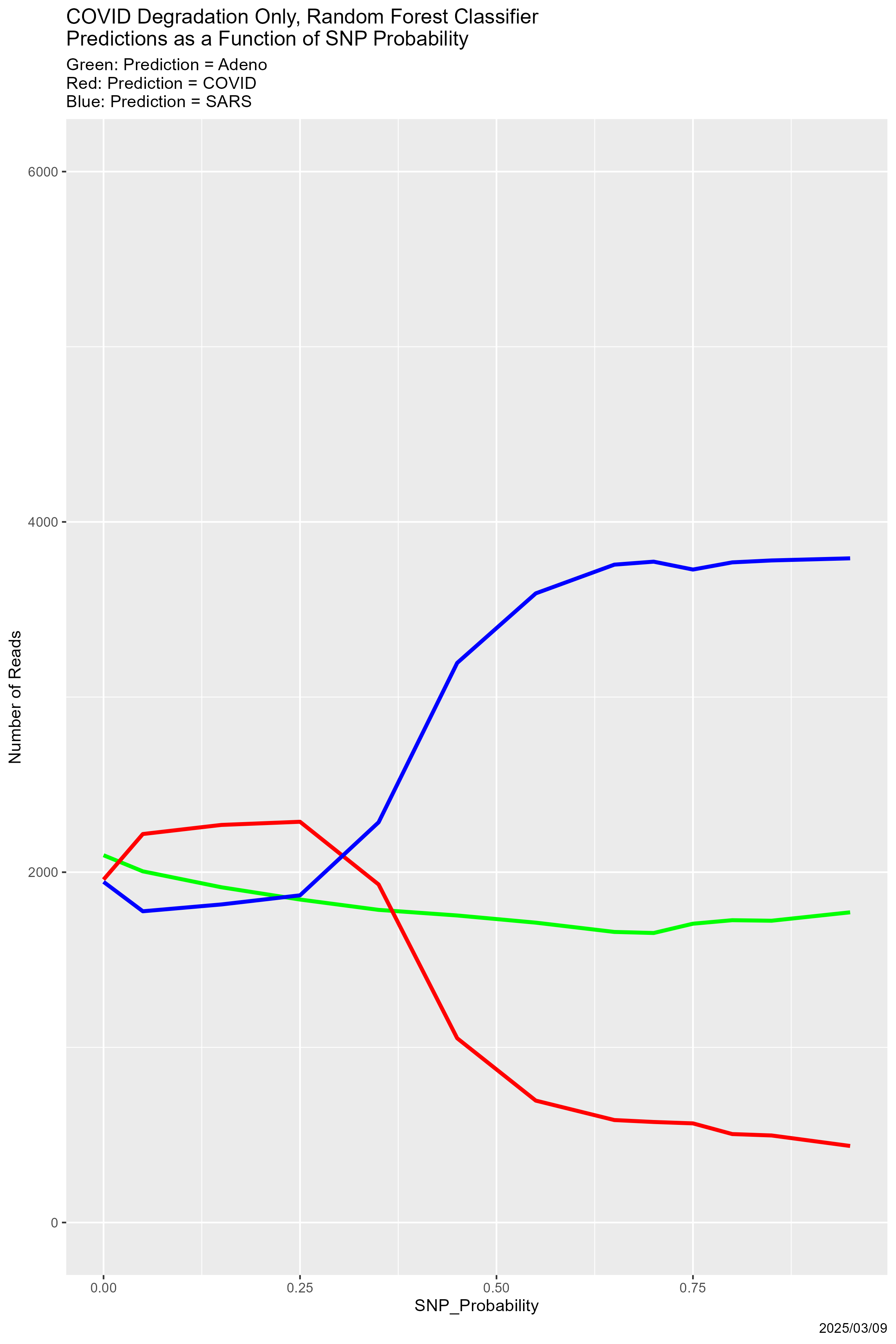}
\caption{SNP degradation applied only to COVID reads: classifier predictions as a function of SNP\_Probability. \emph{Left:} neural net. \emph{Center:} partition model. \emph{Right:} random forest. \emph{Green:} prediction = Adeno. \emph{Red:} prediction = COVID. \emph{Blue:} prediction = SARS.}
\label{fig.COVIDOnly-predictions}
\end{figure}

\begin{figure}[ht]
\centering
\includegraphics[width = 1.75in]{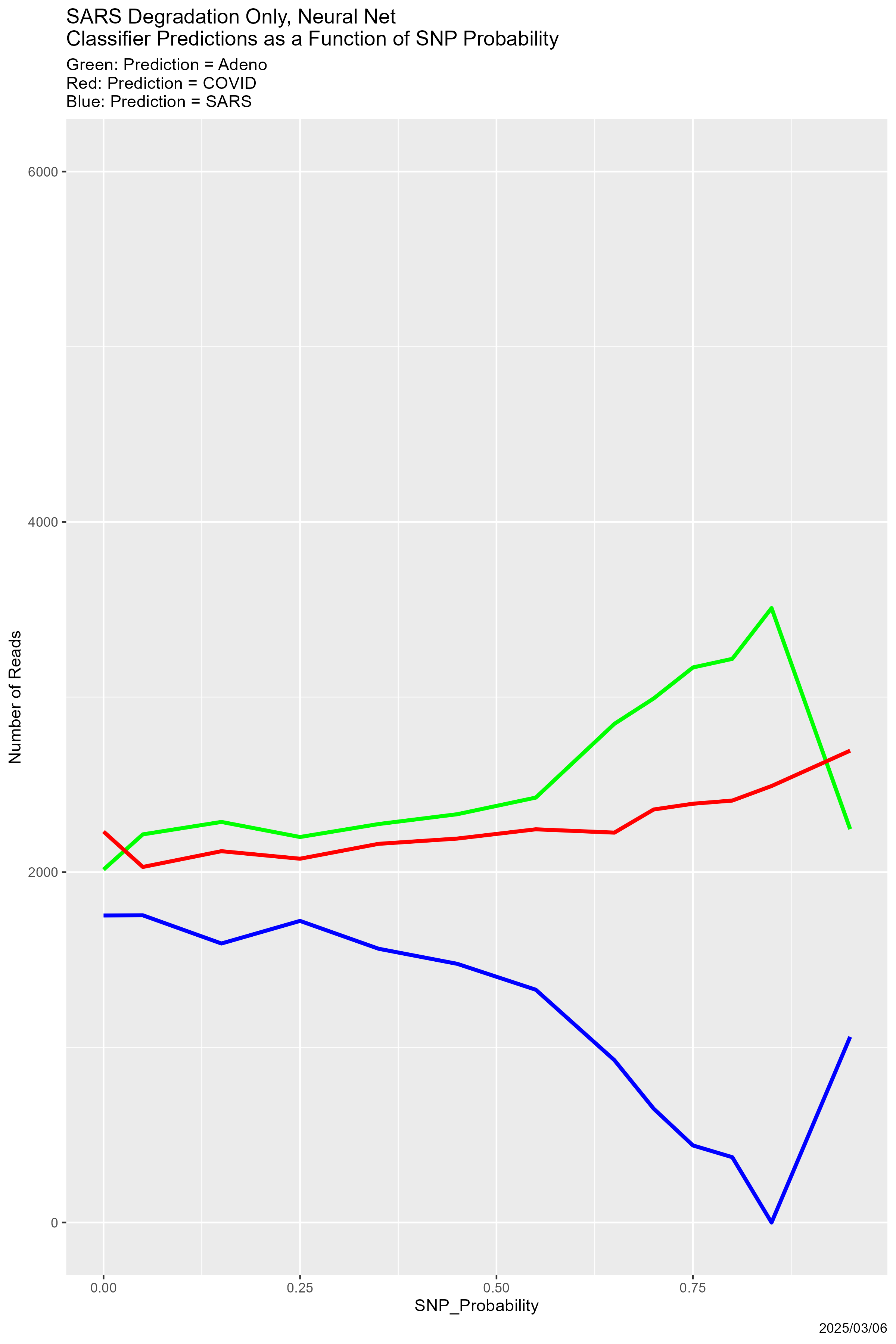}\hspace{.25in}
\includegraphics[width = 1.75in]{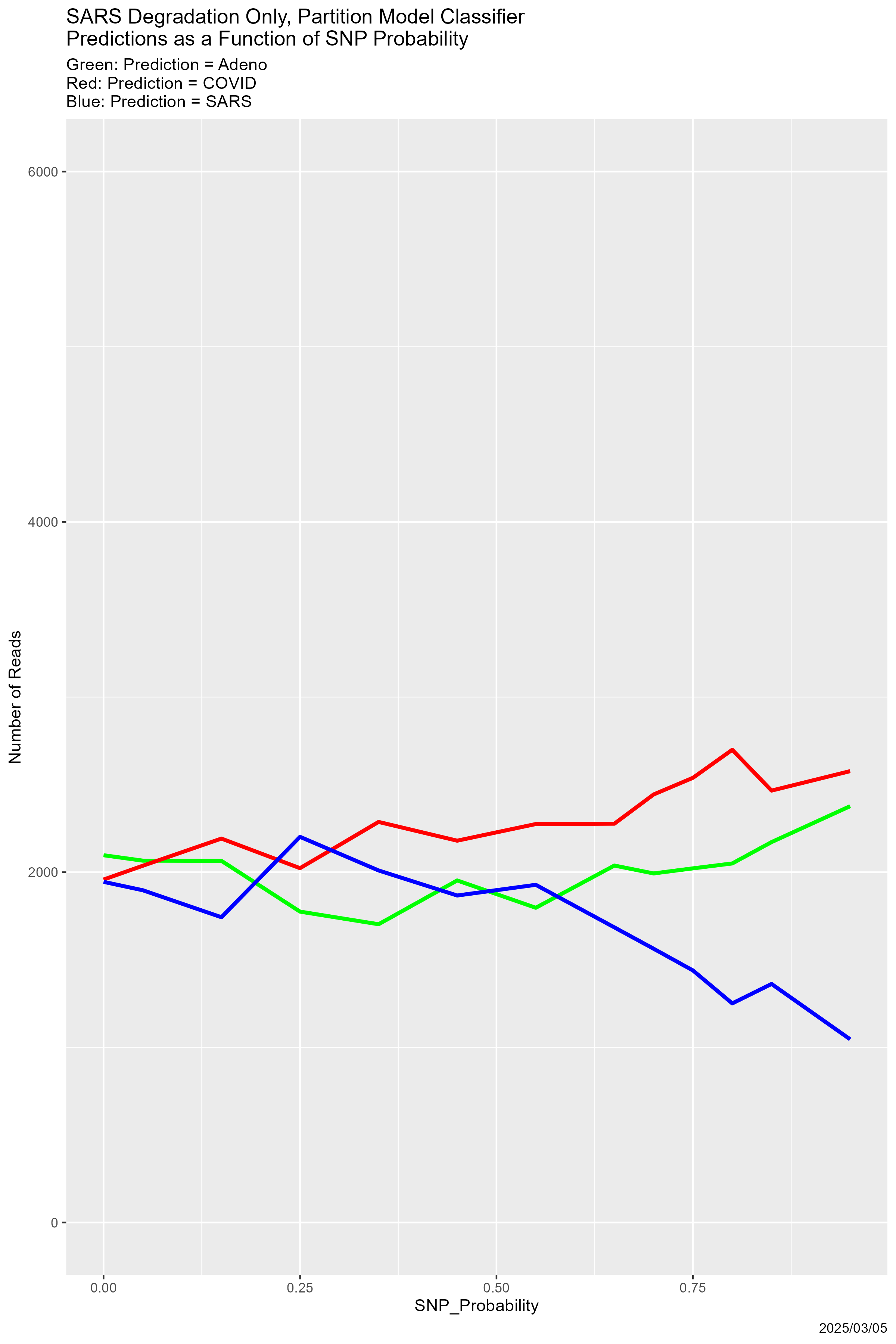}\hspace{.25in}
\includegraphics[width = 1.75in]{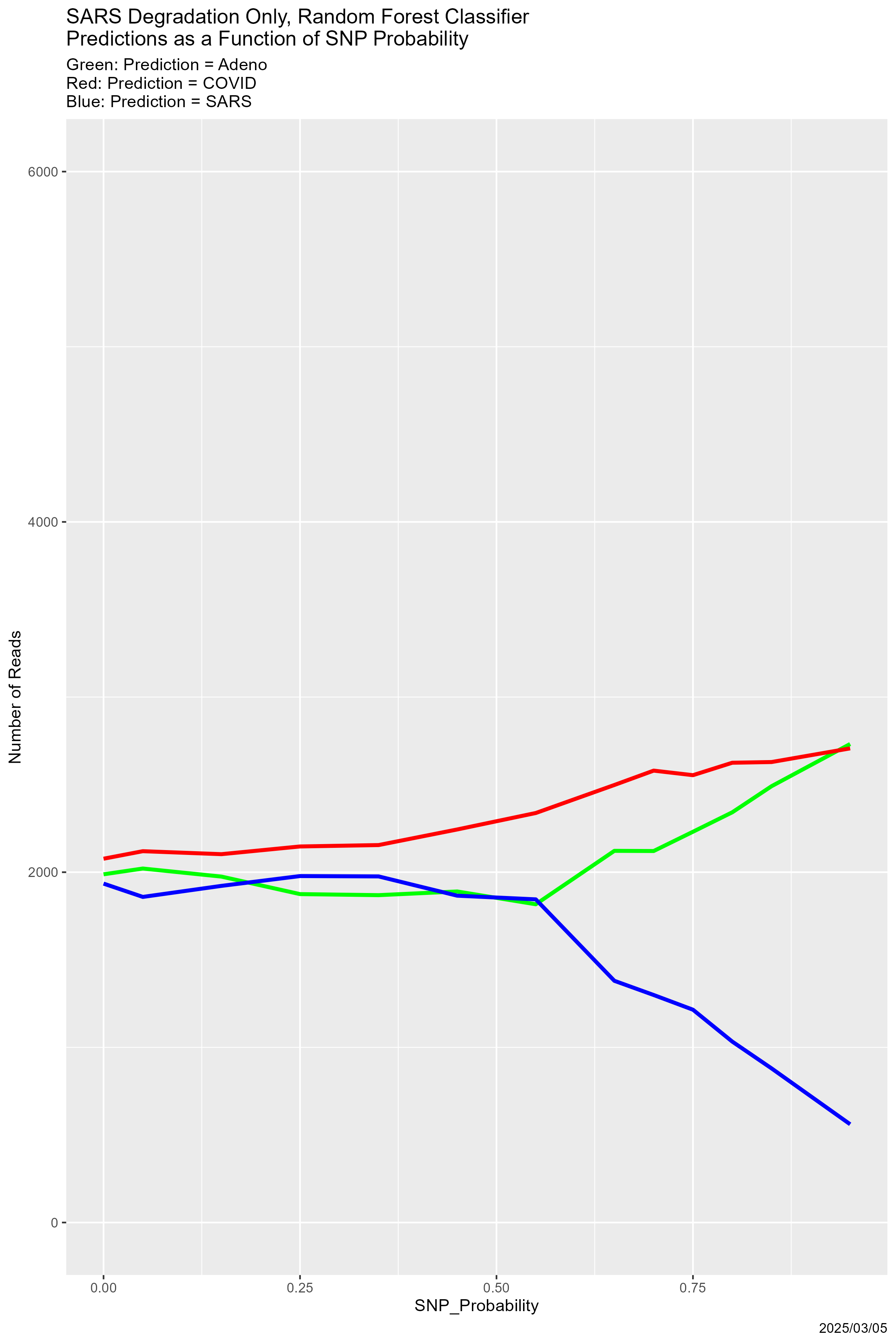}
\caption{SNP degradation applied only to SARS reads: classifier predictions as a function of SNP\_Probability. \emph{Left:} neural net. \emph{Center:} partition model. \emph{Right:} random forest. \emph{Green:} prediction = Adeno. \emph{Red:} prediction = COVID. \emph{Blue:} prediction = SARS.}
\label{fig.SARSOnly-predictions}
\end{figure}

\begin{figure}[ht]
\centering
\includegraphics[width = 1.75in]{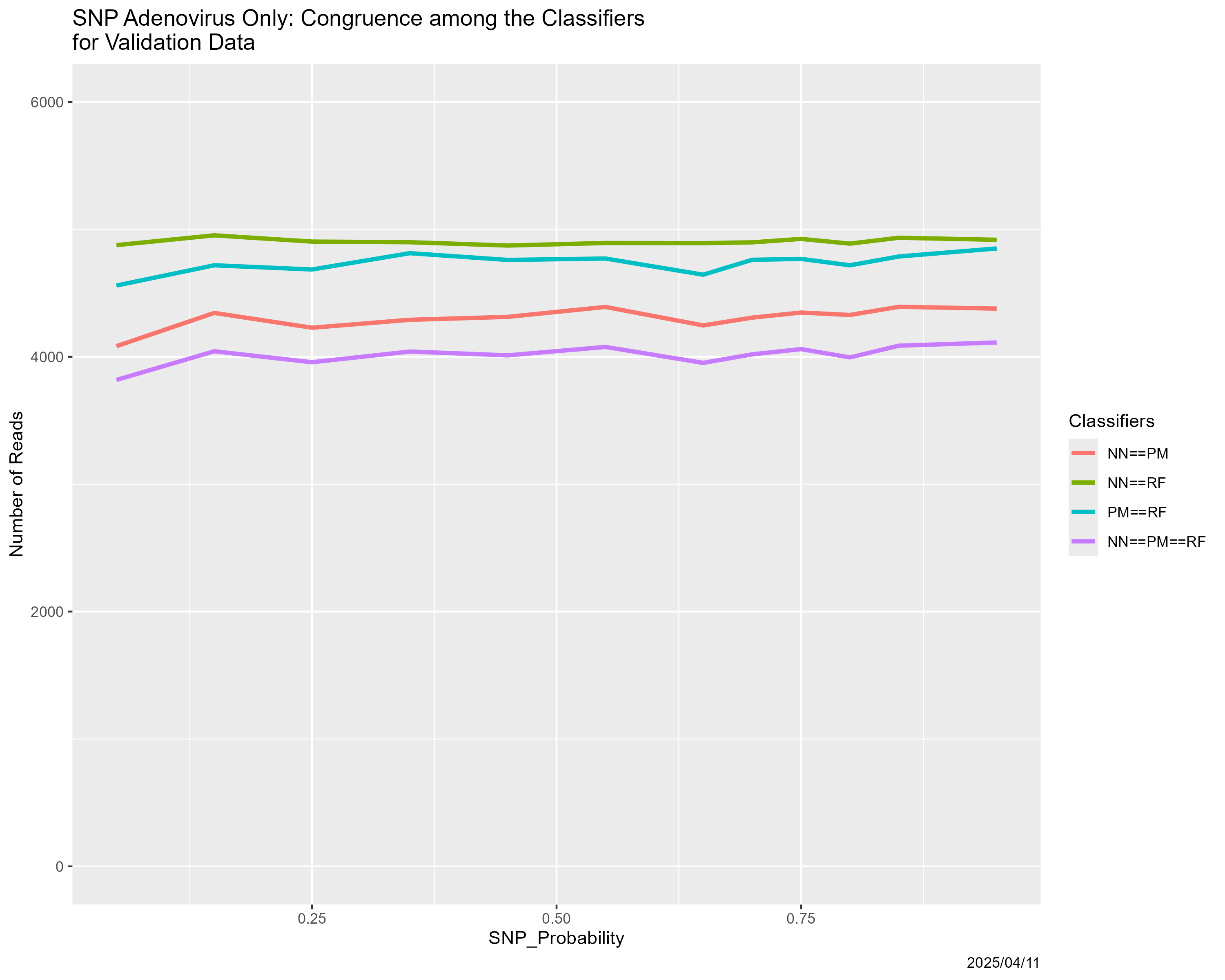}\hspace{.25in}
\includegraphics[width = 1.75in]{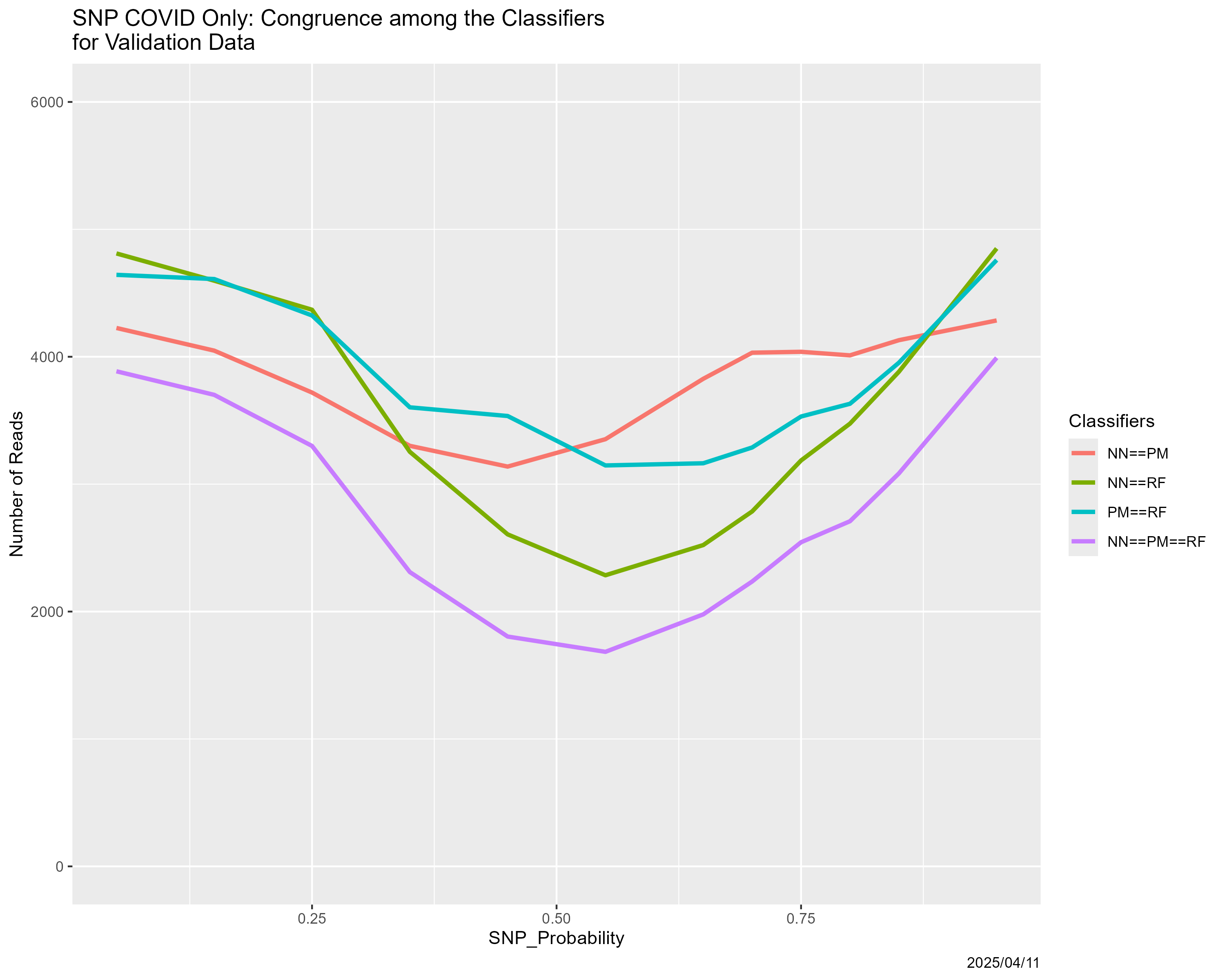}\hspace{.25in}
\includegraphics[width = 1.75in]{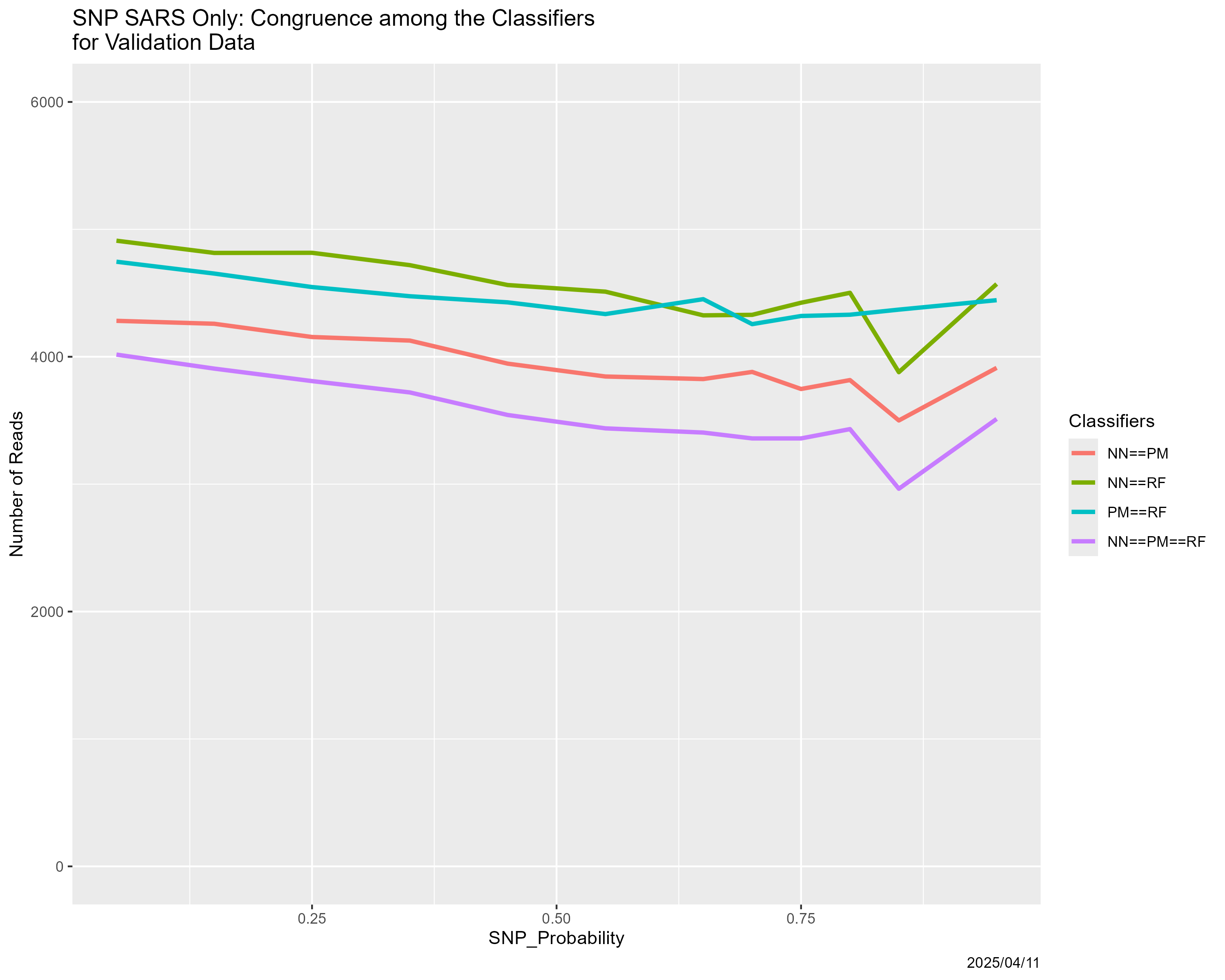}
\caption{SNP degradation applied only to one set of reads: classifier congruence as a function of SNP\_Probability. \emph{Left:} applied only to Adeno reads in training data. \emph{Center:} applied only to COVID reads in training data. \emph{Right:} applied only to SARS reads in training data.}
\label{fig.one-only-congruence}
\end{figure}

Figures \ref{fig.AdenoOnly-predictions}--\ref{fig.SARSOnly-predictions} show predictions from the ``degrade only one genome'' scenario. Congruence is in Figure \ref{fig.one-only-congruence}. The results reinforce those for ``protecting one genome.'' Uniformly across genomes and classifiers, predictions are increasingly diverted from the one affected genome to the two genomes. Evidence of breakdown is, interestingly, different and not uniform. In Figure \ref{fig.COVIDOnly-predictions}, something is happening in the vicinity of SNP\_Probability = 0.25 for all three classifiers. There is no hint of this in Figures \ref{fig.AdenoOnly-predictions} and \ref{fig.SARSOnly-predictions}. Whether something is happening in Figure \ref{fig.SARSOnly-predictions} for values of SNP\_Probability in the vicinity of 0.8 is even less clear, because in any event, it is not happening for random forests. But, then is something happening for random forests at SNP\_Probability = 0.60?

\subsection{Other Quality Problems}\label{subsec.other-forms}
Here we address the question of whether the behavior described in Section \ref{subsec.snp} is specific to \SNP-degradation, or holds as well for ways of decreasing data quality.

\subsubsection{Mislabeled Reads}\label{subsubsec.mislabel}
All four of the classifiers depend on labeled training data. What happens if there are problems with the labeling process, as there inevitably are in reality? In the experiment described next, randomly selected elements of \TD\ have their labels changed. The Mislabel\_Probability varies over 
$$
\{0, 0.05, 0.15, 0.25, 0.35, 0.45, 0.55, 0.65, 0.70, 0.75, 0.80, 0.85, 0.95\}.
$$

\begin{figure}[ht]
\centering
\includegraphics[width = 1.75in]{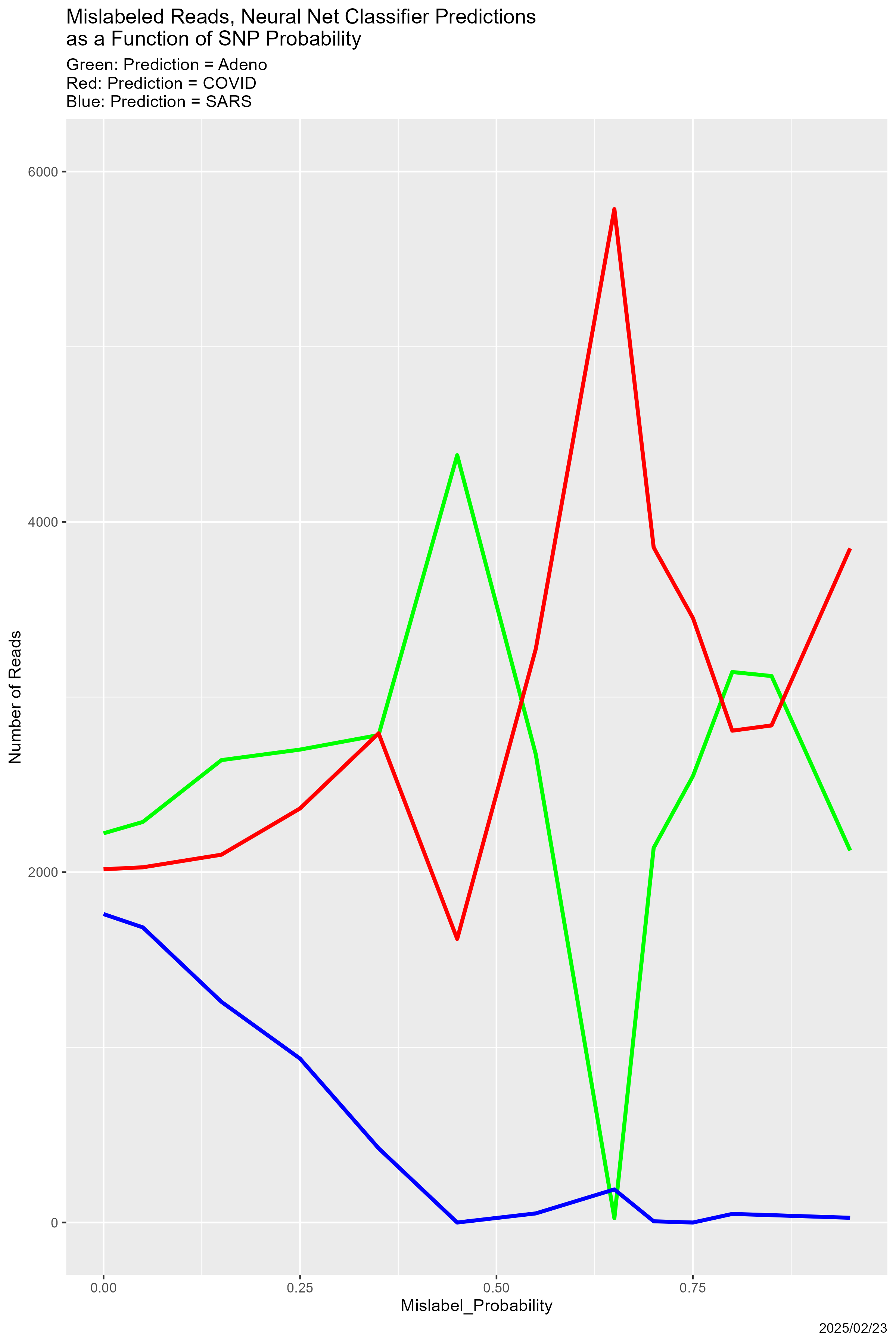}\hspace{.25in}
\includegraphics[width = 1.75in]{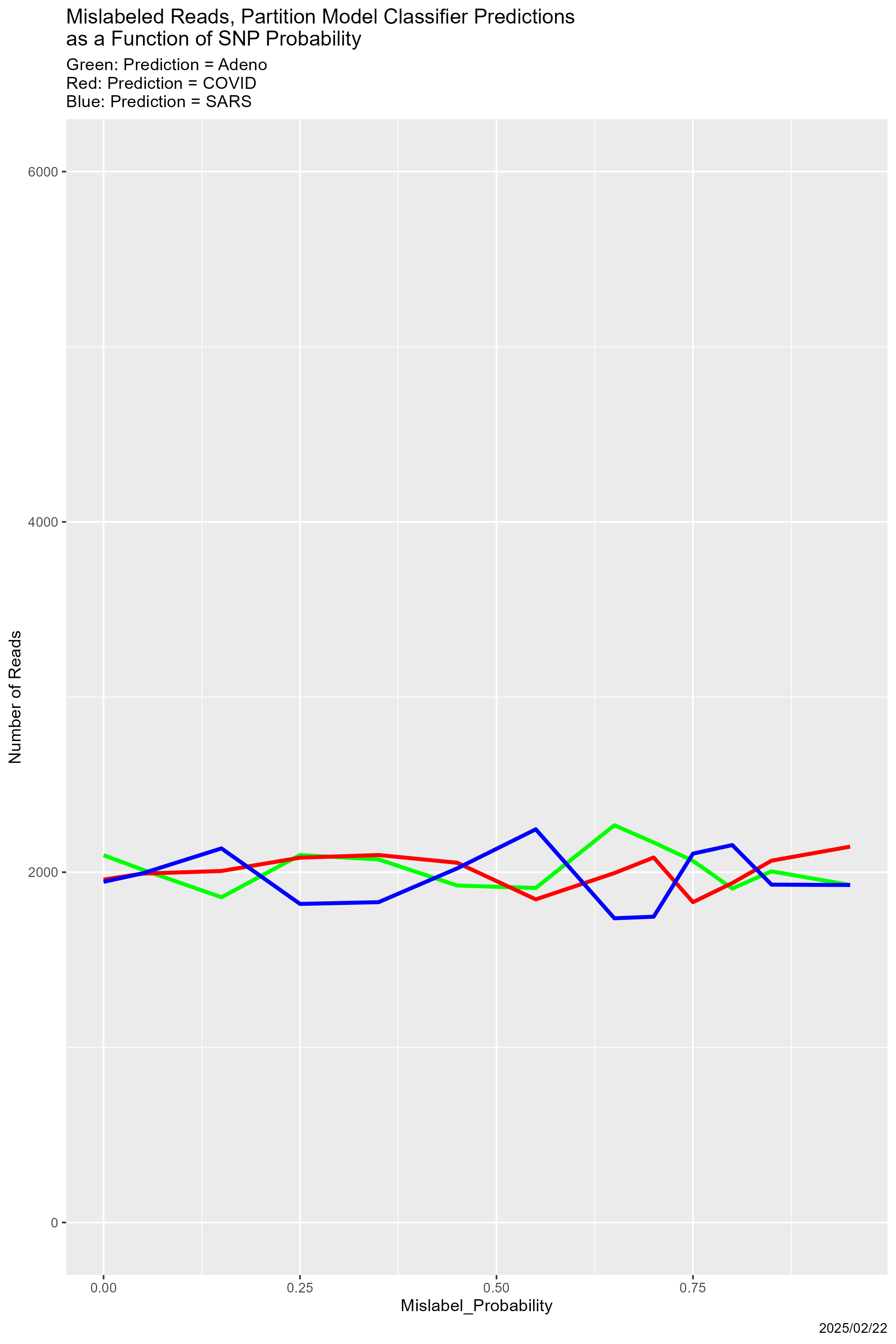}\hspace{.25in}
\includegraphics[width = 1.75in]{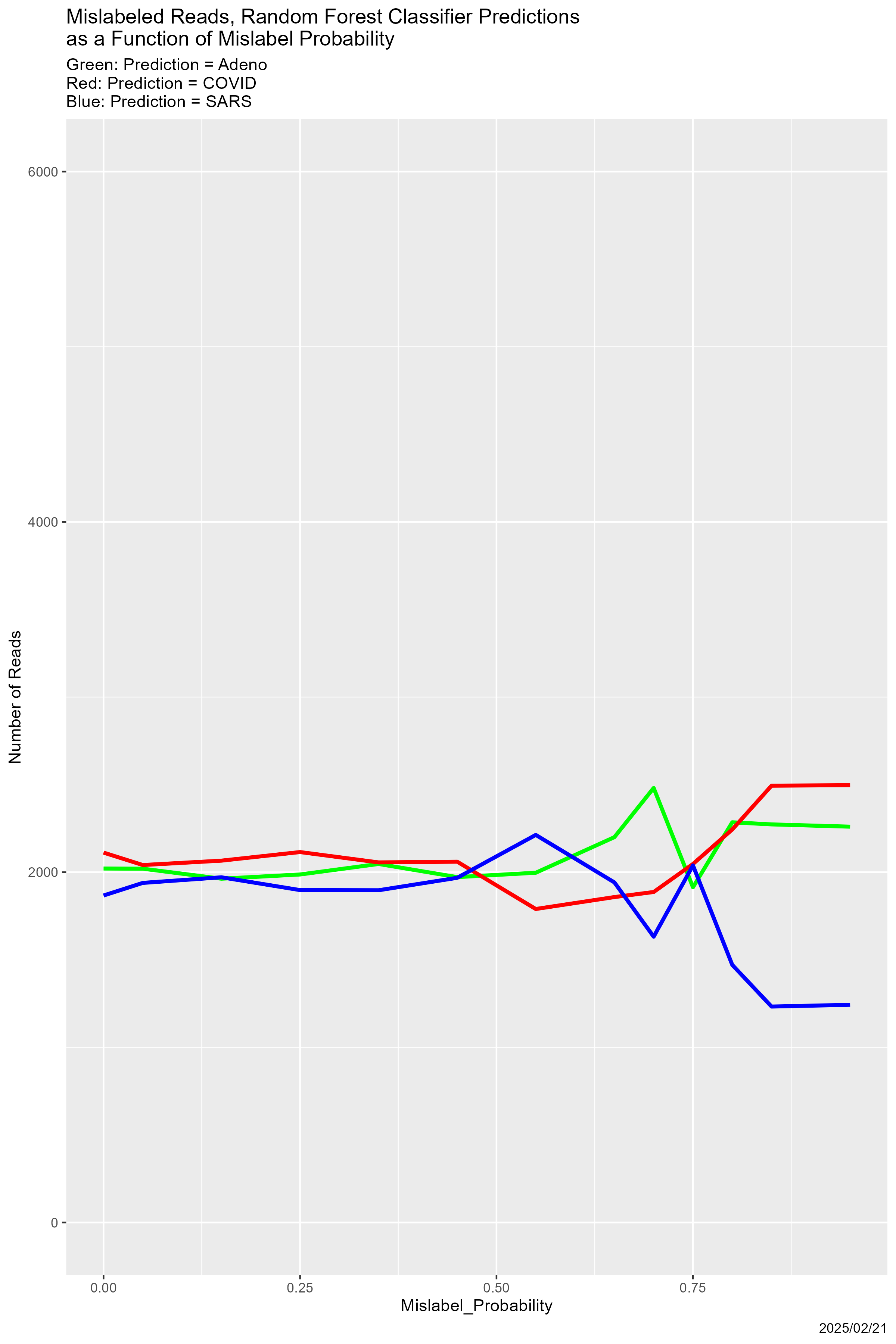}
\caption{Mislabeled training data: classifier predictions as a function of Mislabel\_PROBABILITY. \emph{Left:} neural net. \emph{Center:} partition model. \emph{Right:} random forest. \emph{Green:} prediction = Adeno. \emph{Red:} prediction = COVID. \emph{Blue:} prediction = SARS.}
\label{fig.mislabeled-predictions}
\end{figure}

\begin{figure}[ht]
\centering
\includegraphics[width=3in]{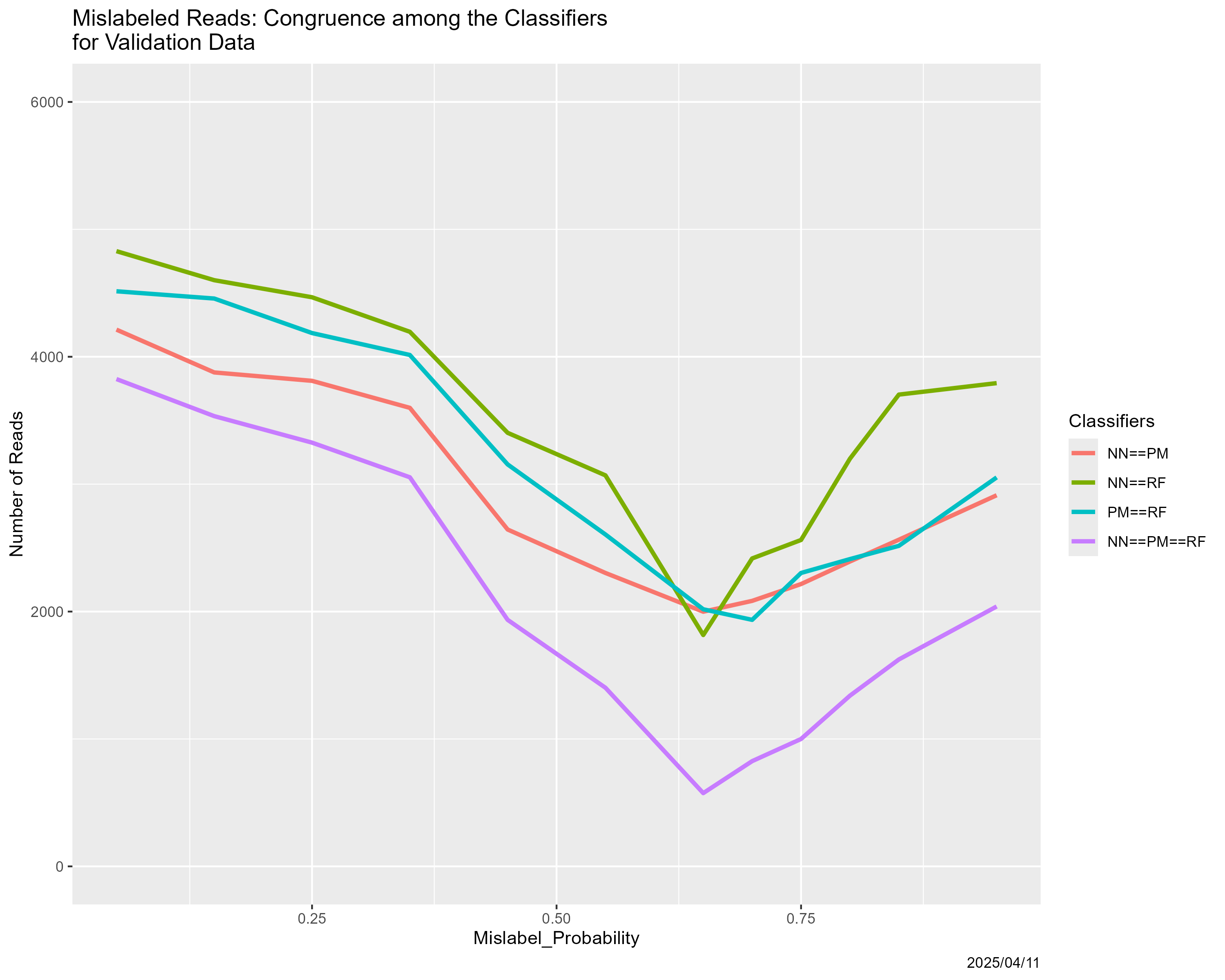}
\caption{Mislabeled training data: classifier congruence as a function of Mislabel\_PROBABILITY.}
\label{fig.mislabeled-congruence}
\end{figure}

The results, in Figures \ref{fig.mislabeled-predictions} and \ref{fig.mislabeled-congruence}, are ambiguous compared to those in previous sections. Starting with the former, predictions by the partition model are largely insensitive to Mislabel\_Probability. Those for the random forest begin to diverge when Mislabel\_Probability reaches approximately 0.75 (again!). The neural net seems inexplicably erratic. The disappearance of SARS from its predictions mirrors random forests, but is much more extreme and begins for much smaller values of Mislabel\_Probability. The two extreme oscillations between predictions of Adeno and predictions of COVID have no analogues elsewhere in our experiments. 

And yet, Figure \ref{fig.mislabeled-congruence} echoes its predecessors, although the minimum values congruence appear to be for Mislabel\_Probability less than 0.75.

\subsubsection{Reversed Reads}\label{subsubsec.reversed}
\begin{figure}[ht]
\centering
\includegraphics[width = 1.75in]{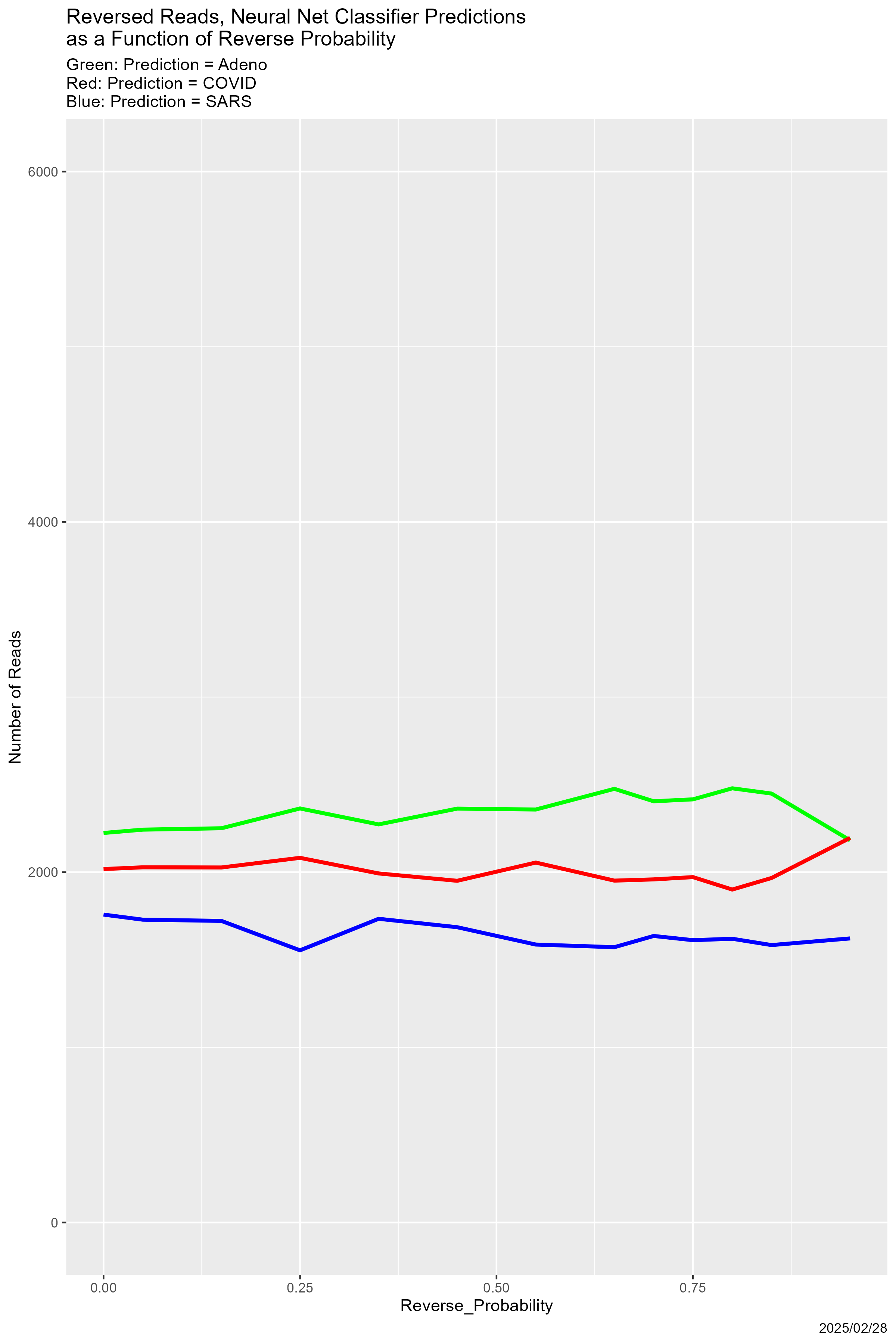}\hspace{.25in}
\includegraphics[width = 1.75in]{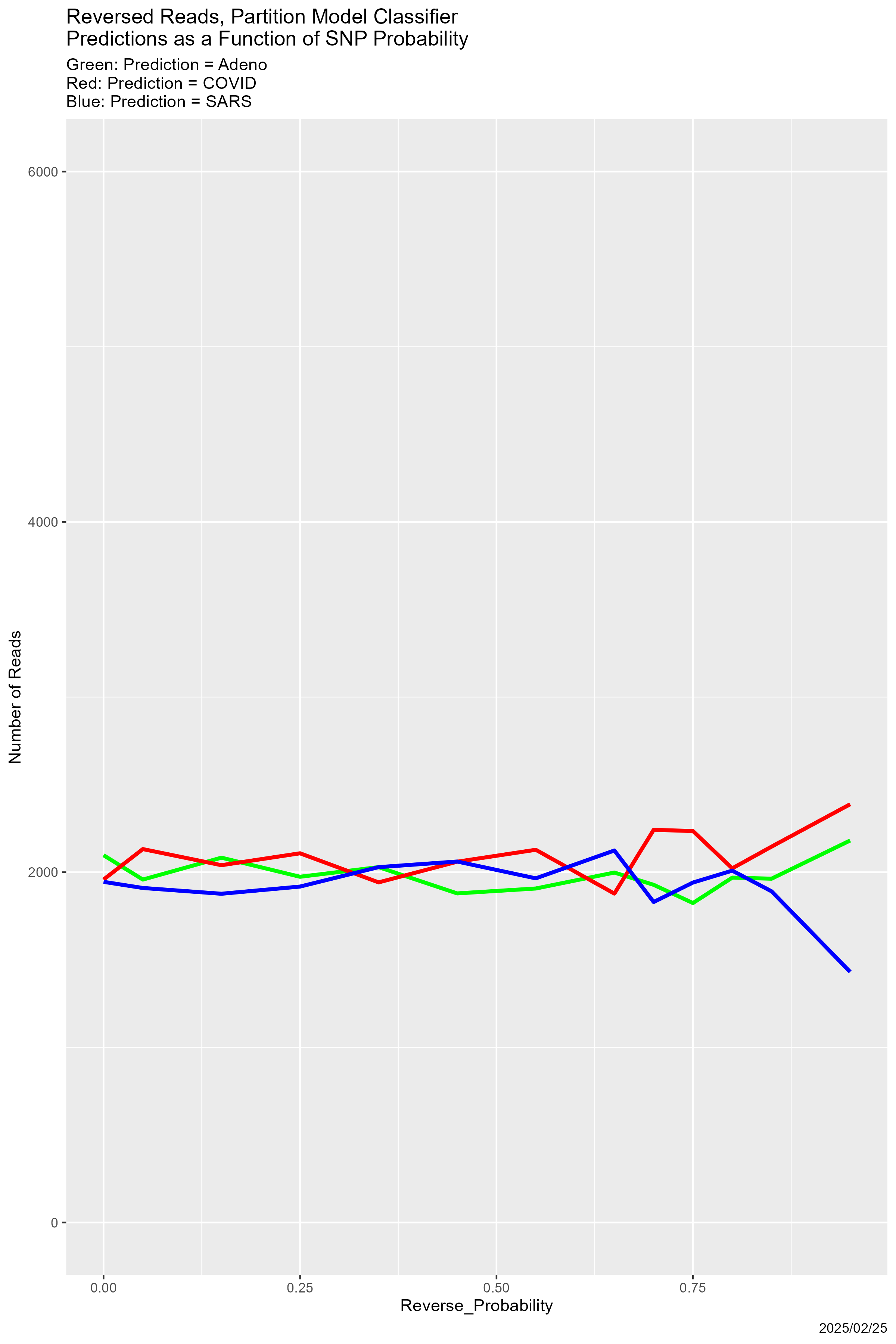}\hspace{.25in}
\includegraphics[width = 1.75in]{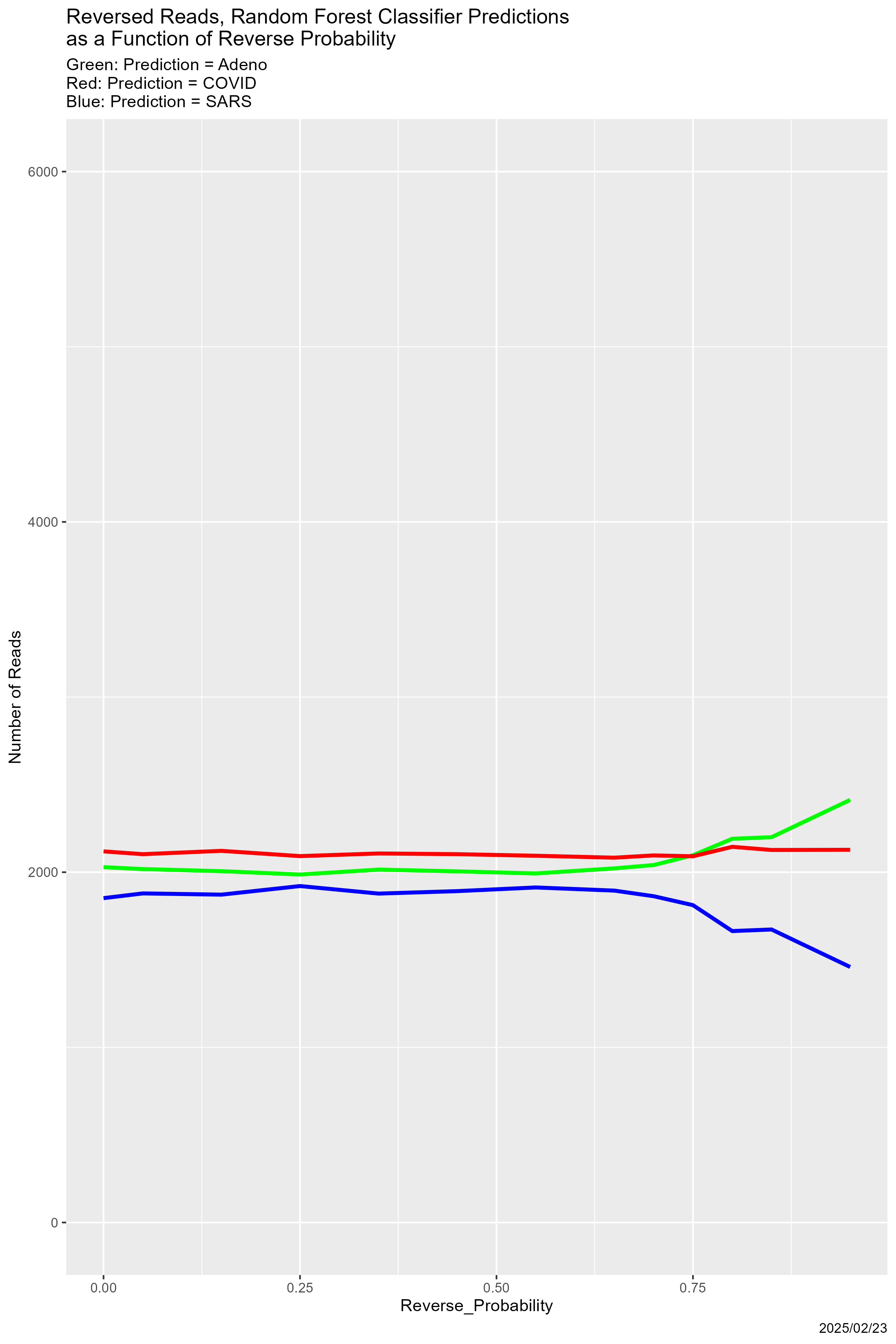}
\caption{Reversed training data: classifier predictions as a function of the probability of reversing reads in the training dataset \TD. \emph{Left:} neural net. \emph{Center:} partition model. \emph{Right:} random forest. \emph{Green:} prediction = Adeno. \emph{Red:} prediction = COVID. \emph{Blue:} prediction = SARS.}
\label{fig.reversed-predictions}
\end{figure}

\begin{figure}[ht]
\centering
\includegraphics[width=3in]{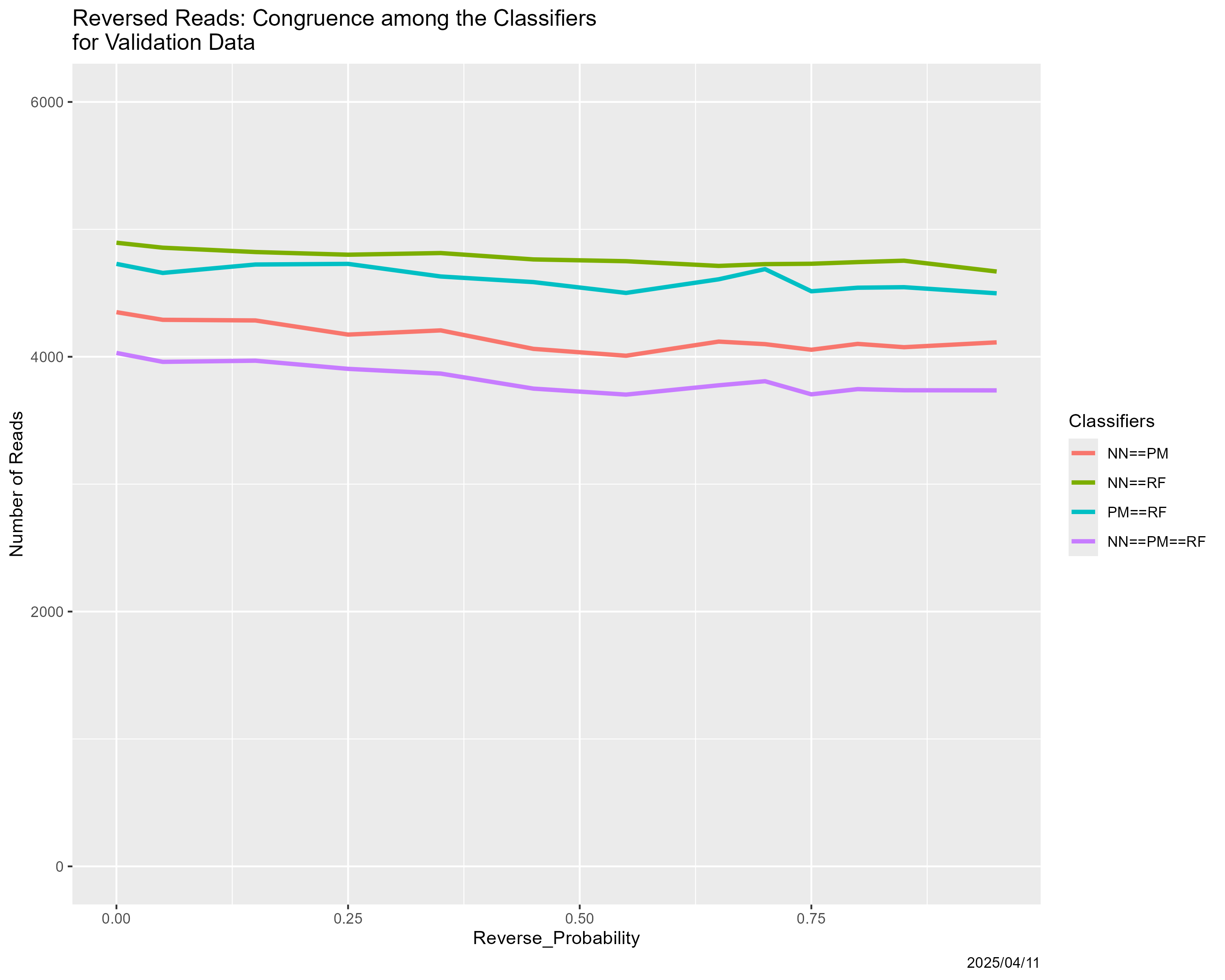}
\caption{Reversed training data: classifier congruence as a function of probability of reversing reads in the training dataset \TD.}
\label{fig.reversed-congruence}
\end{figure}

This is one case where we correctly expected nothing to happen. Recalling that DNA molecules are double helices \citep{watson2001}, there is inherent uncertainty in \NGS\ reads regarding which strand they come from. (Sequenced genomes have a well-determined 5' end, characterized by a free phosphate, and 3' end, characterized by a free hydroxyl.) Therefore, reversing some reads, as character strings, should have minimal effect. Figures \ref{fig.reversed-predictions} and \ref{fig.reversed-congruence} confirm this expectation.

\subsubsection{Reduced Training Data}\label{subsubsec.reduced}
Here we explore the effect of a more global quality characteristic of training data---their size and balance. Recall from Section \ref{subsec.datasets} that \TD\ contains approximately equal numbers of Adeno, COVID and SARS reads. so does \VD, both of which may matter. In each run of this experiment, reads with a specified source are deleted at random from \TD, so that the experiment is parameterized by a Removal\_Probability taking values in 
$$
\{0, 0.05, 0.15, 0.25, 0.35, 0.45, 0.55, 0.65, 0.70, 0.75, 0.80, 0.85, 0.95\}. 
$$
There is no reason to expect that, if there is breakdown-like behavior, it will occur at Removal\_Probability = 0.75. There is, of course, reason to suspect that as reads from one source are removed, the number of elements in \VD\ predicted to be from that source will decrease.

\begin{figure}[ht]
\centering
\includegraphics[width = 1.75in]{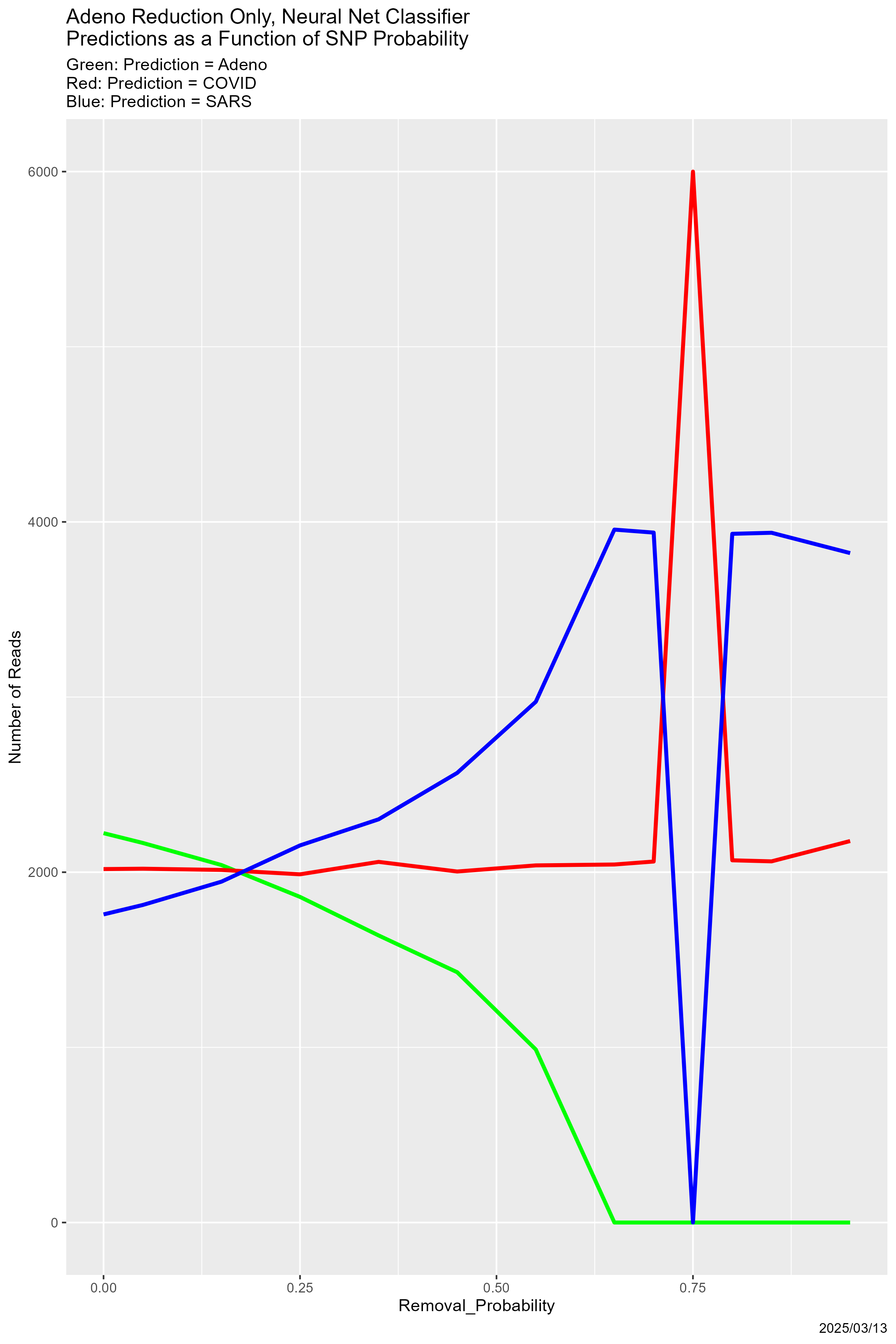}\hspace{.25in}
\includegraphics[width = 1.75in]{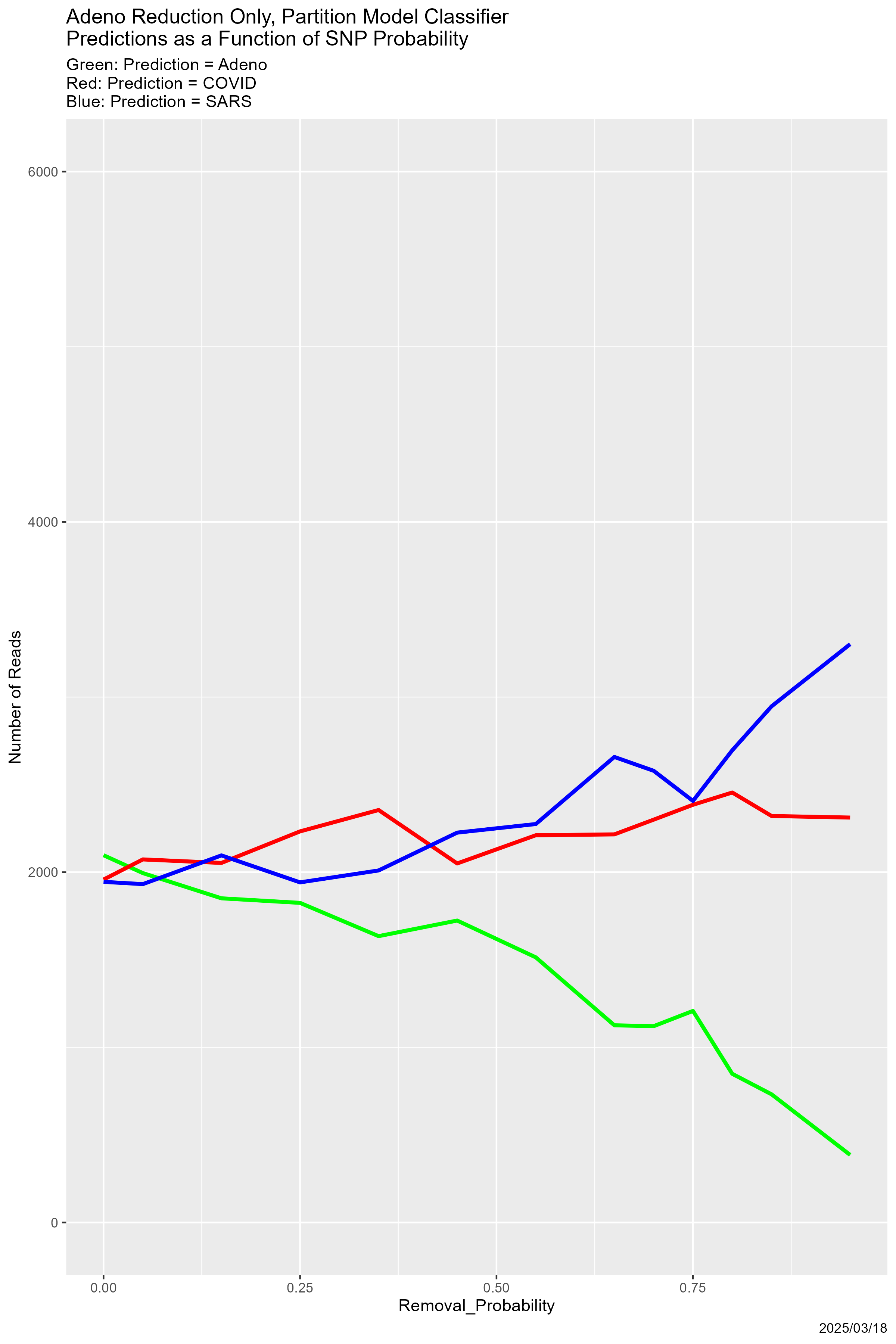}\hspace{.25in}
\includegraphics[width = 1.75in]{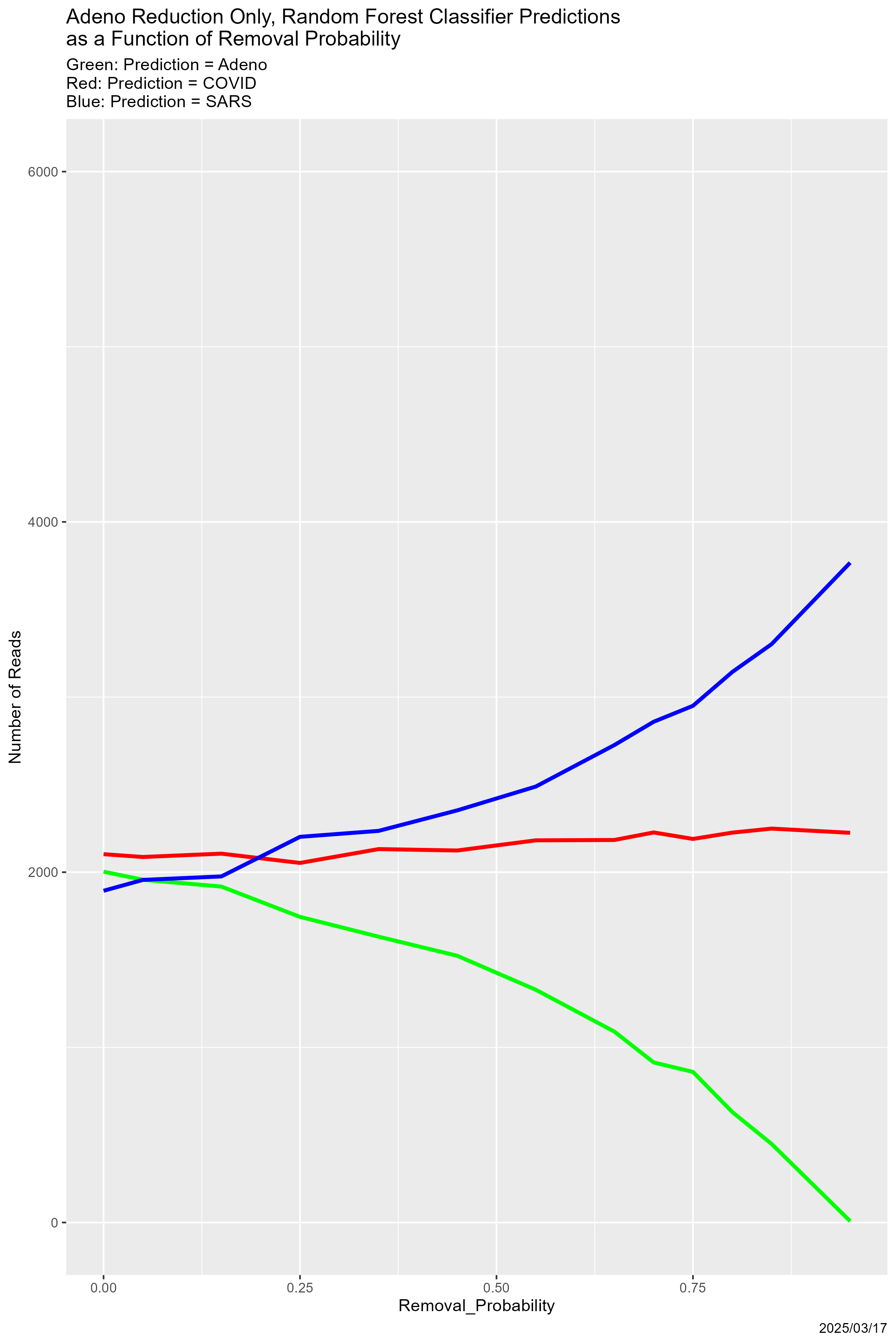}
\caption{Reduced Adeno reads: classifier predictions as a function of Removal\_Probability. \emph{Left:} neural net. \emph{Center:} partition model. \emph{Right:} random forest.  \emph{Green:} prediction = Adeno. \emph{Red:} prediction = COVID. \emph{Blue:} prediction = SARS.}
\label{fig.ReducedAdeno-predictions}
\end{figure}

\begin{figure}[ht]
\centering
\includegraphics[width = 1.75in]{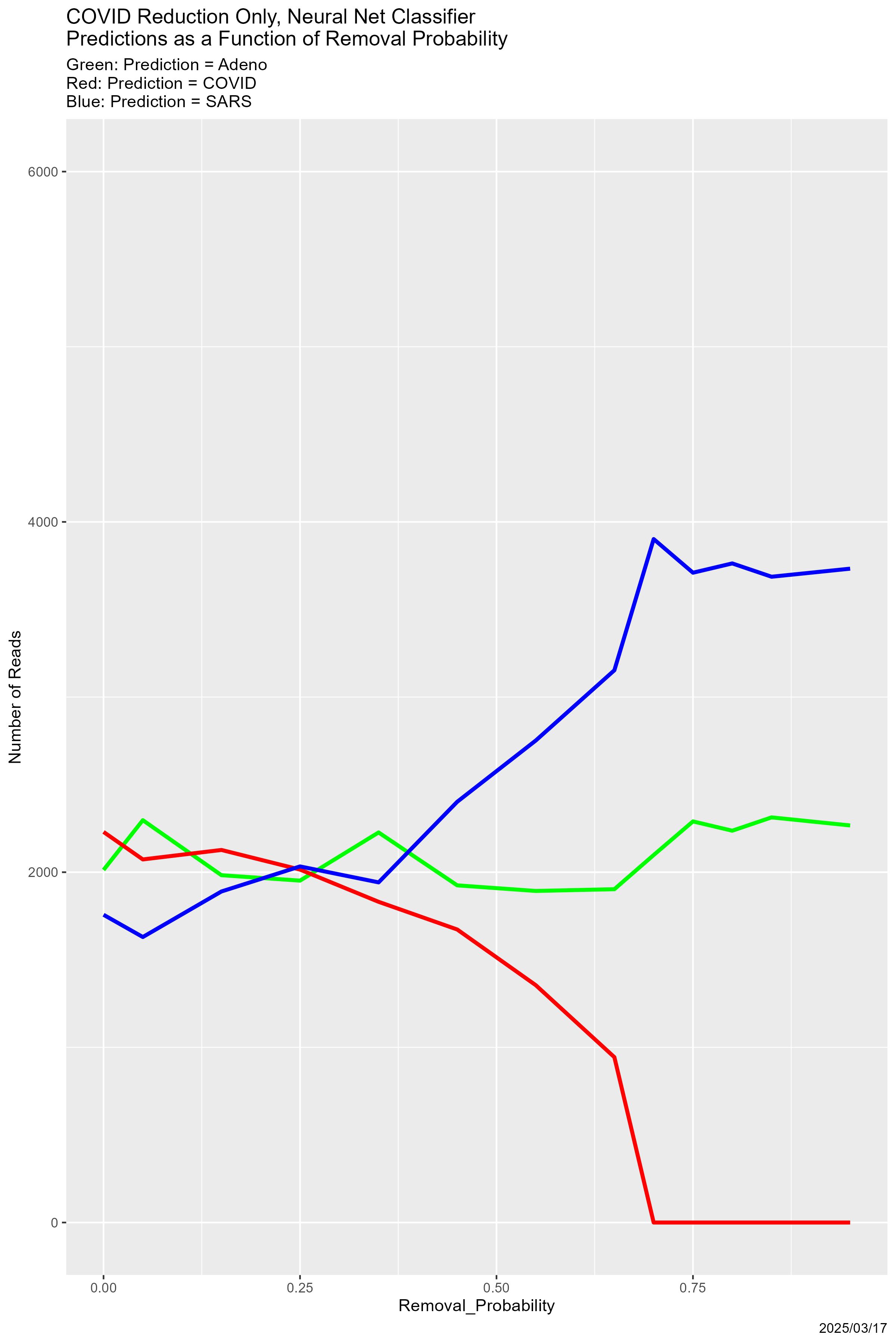}\hspace{.25in}
\includegraphics[width = 1.75in]{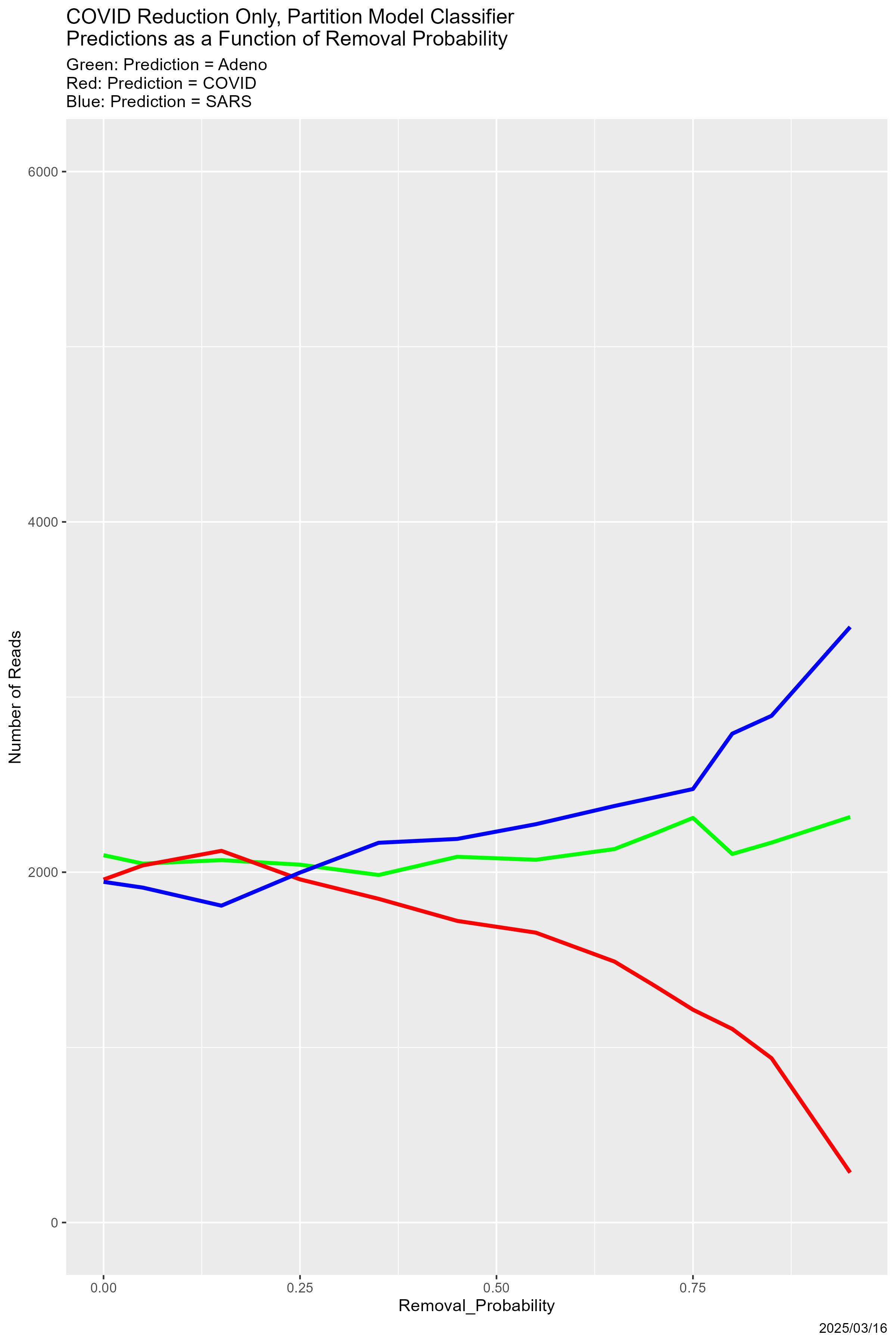}\hspace{.25in}
\includegraphics[width = 1.75in]{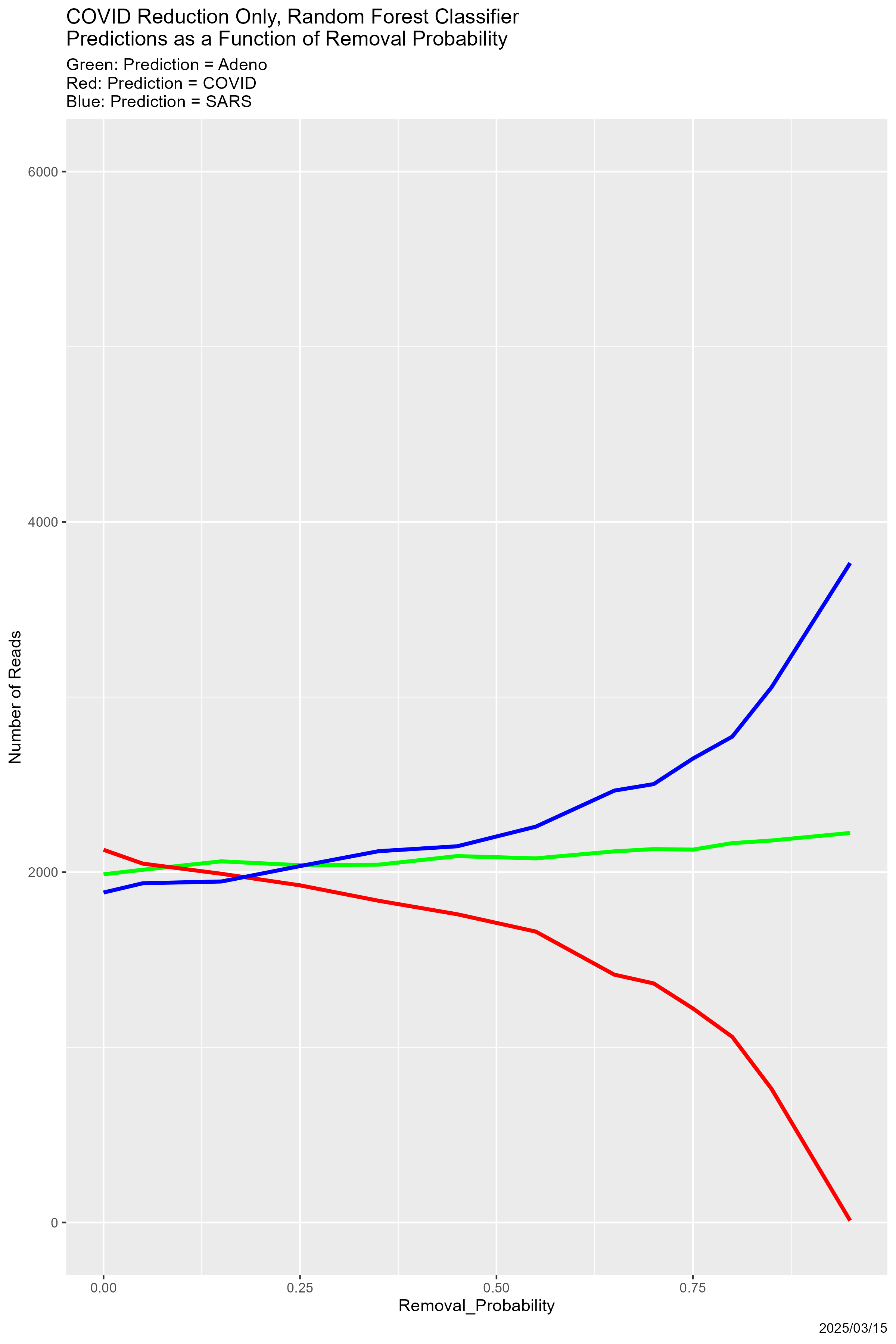}
\caption{Reduced: COVID reads: classifier predictions as a function of Removal\_Probability. \emph{Left:} neural net. \emph{Center:} partition model. \emph{Right:} random forest.  \emph{Green:} prediction = Adeno. \emph{Red:} prediction = COVID. \emph{Blue:} prediction = SARS.}
\label{fig.ReducedCOVID-predictions}
\end{figure}

\begin{figure}[ht]
\centering
\includegraphics[width = 1.75in]{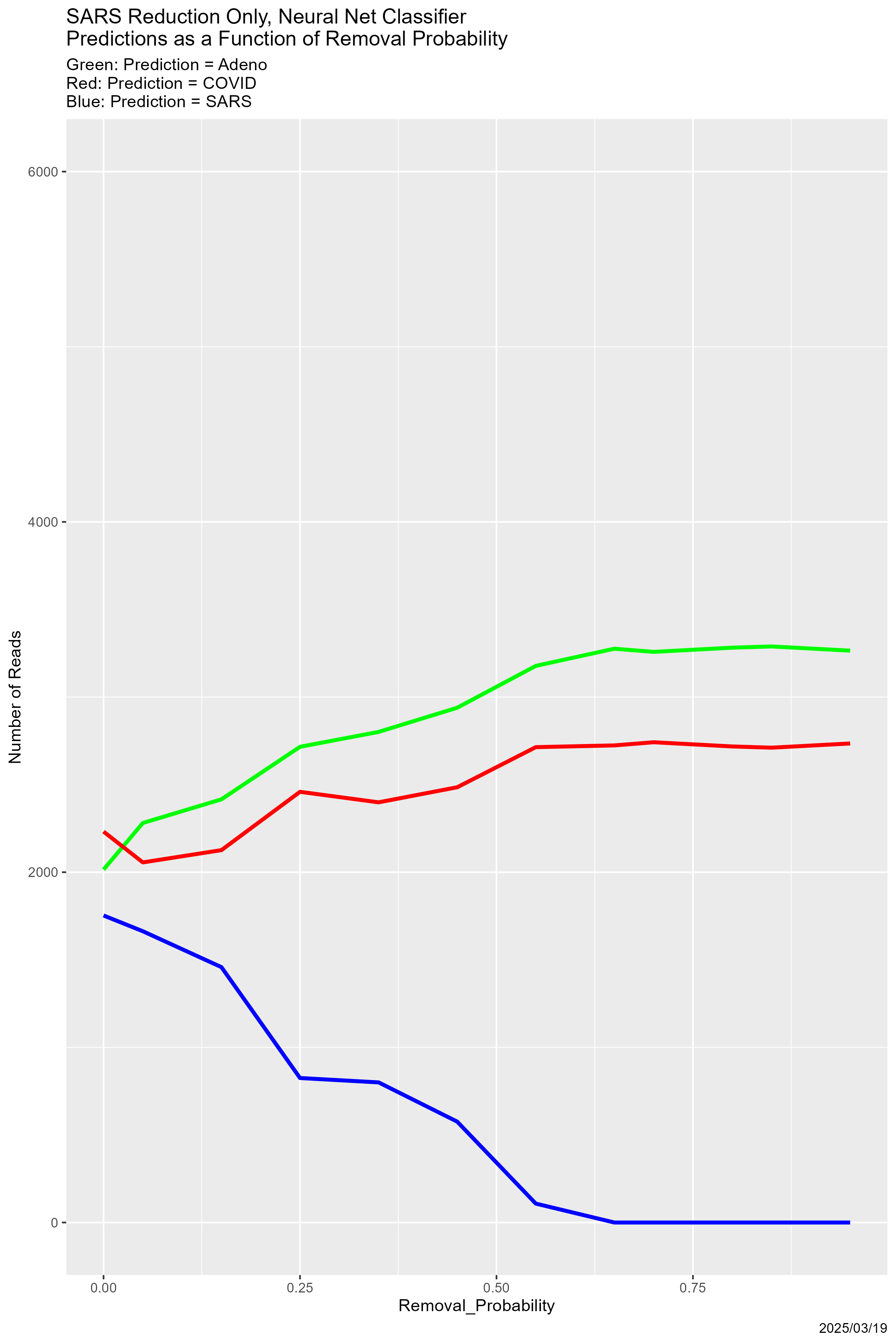}\hspace{.25in}
\includegraphics[width = 1.75in]{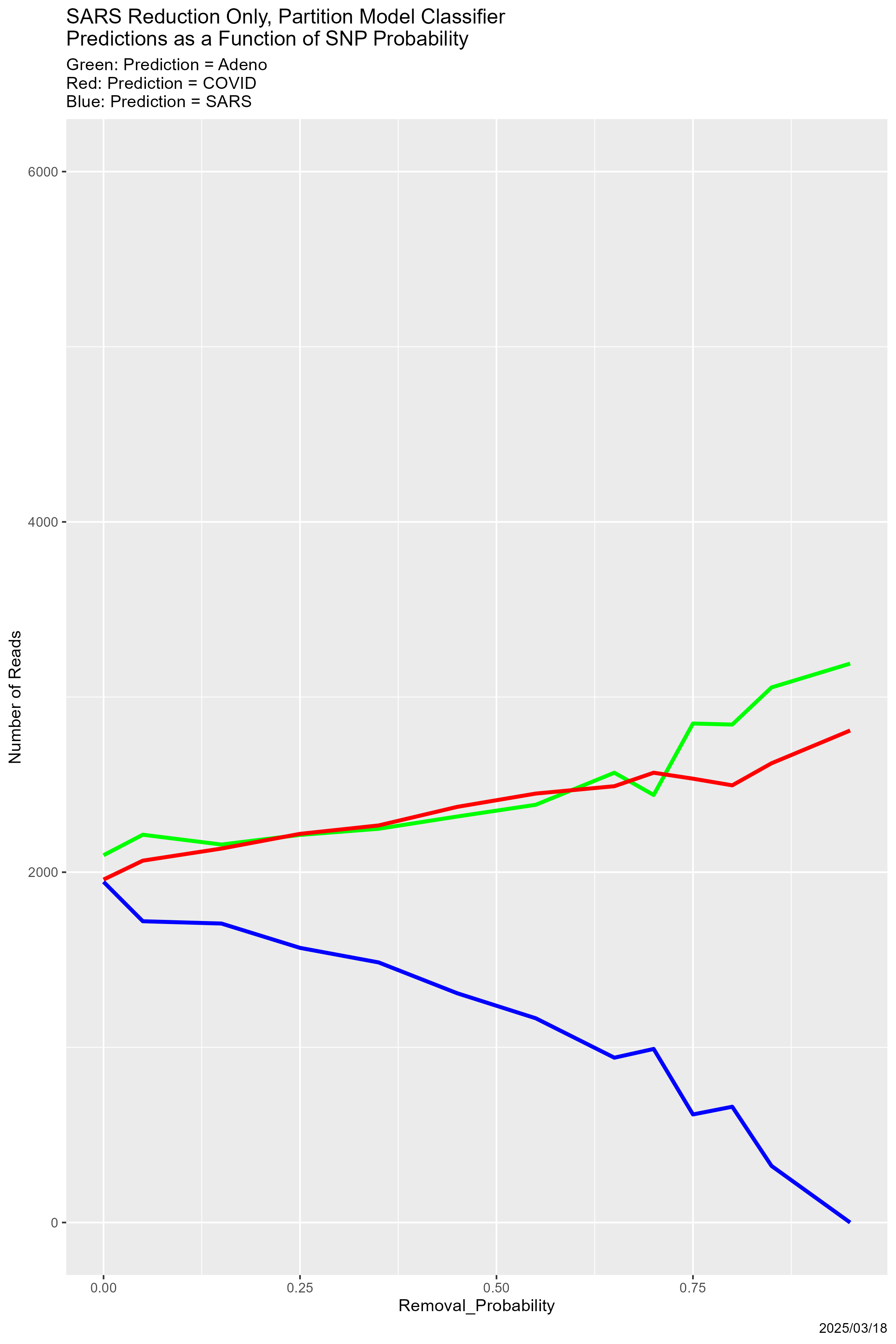}\hspace{.25in}
\includegraphics[width = 1.75in]{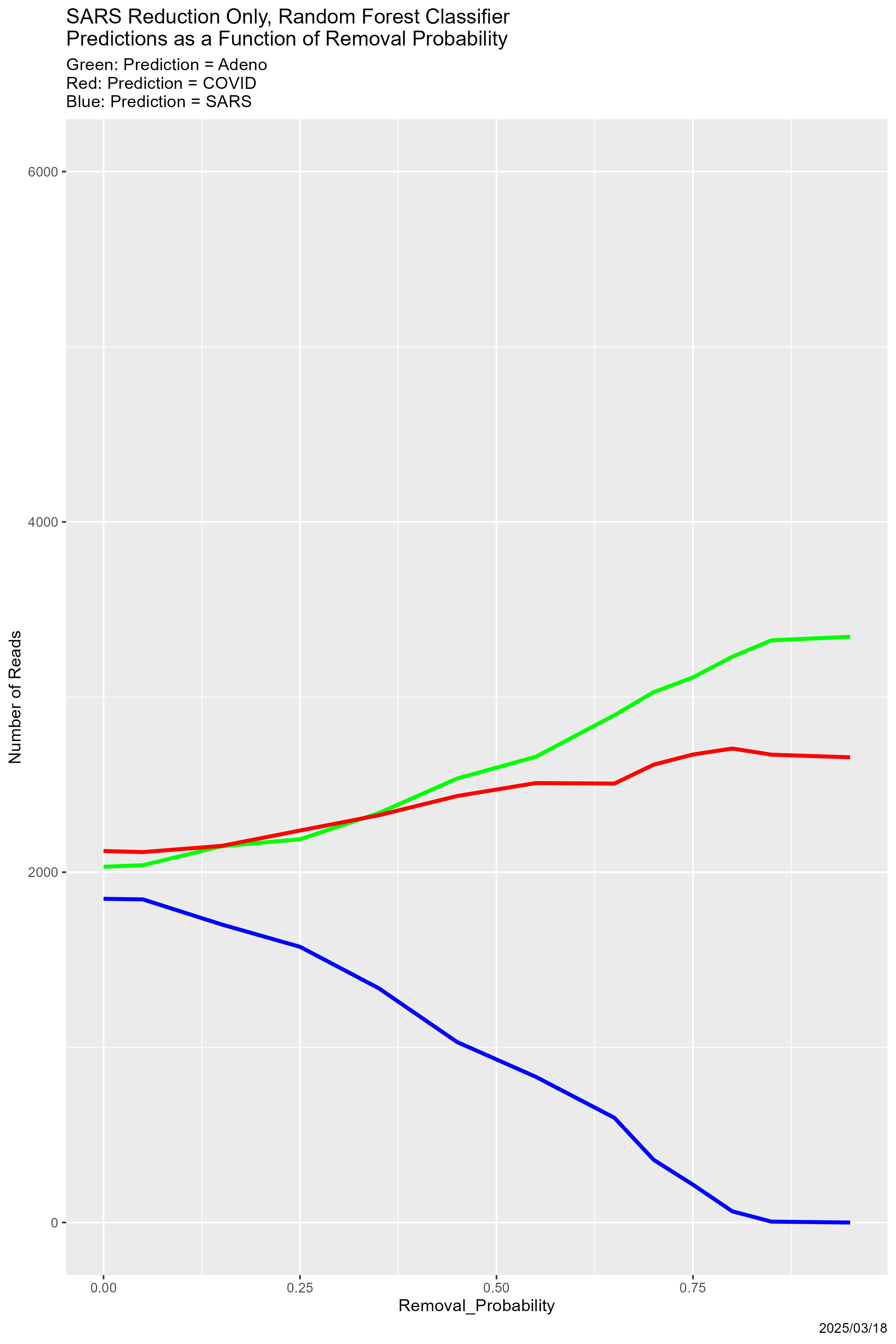}
\caption{Reduced SARS reads: classifier predictions as a function of Removal\_Probability. \emph{Left:} neural net. \emph{Center:} partition model. \emph{Right:} random forest.  \emph{Green:} prediction = Adeno. \emph{Red:} prediction = COVID. \emph{Blue:} prediction = SARS.}
\label{fig.ReducedSARS-predictions}
\end{figure}

\begin{figure}[ht]
\centering
\includegraphics[width = 1.75in]{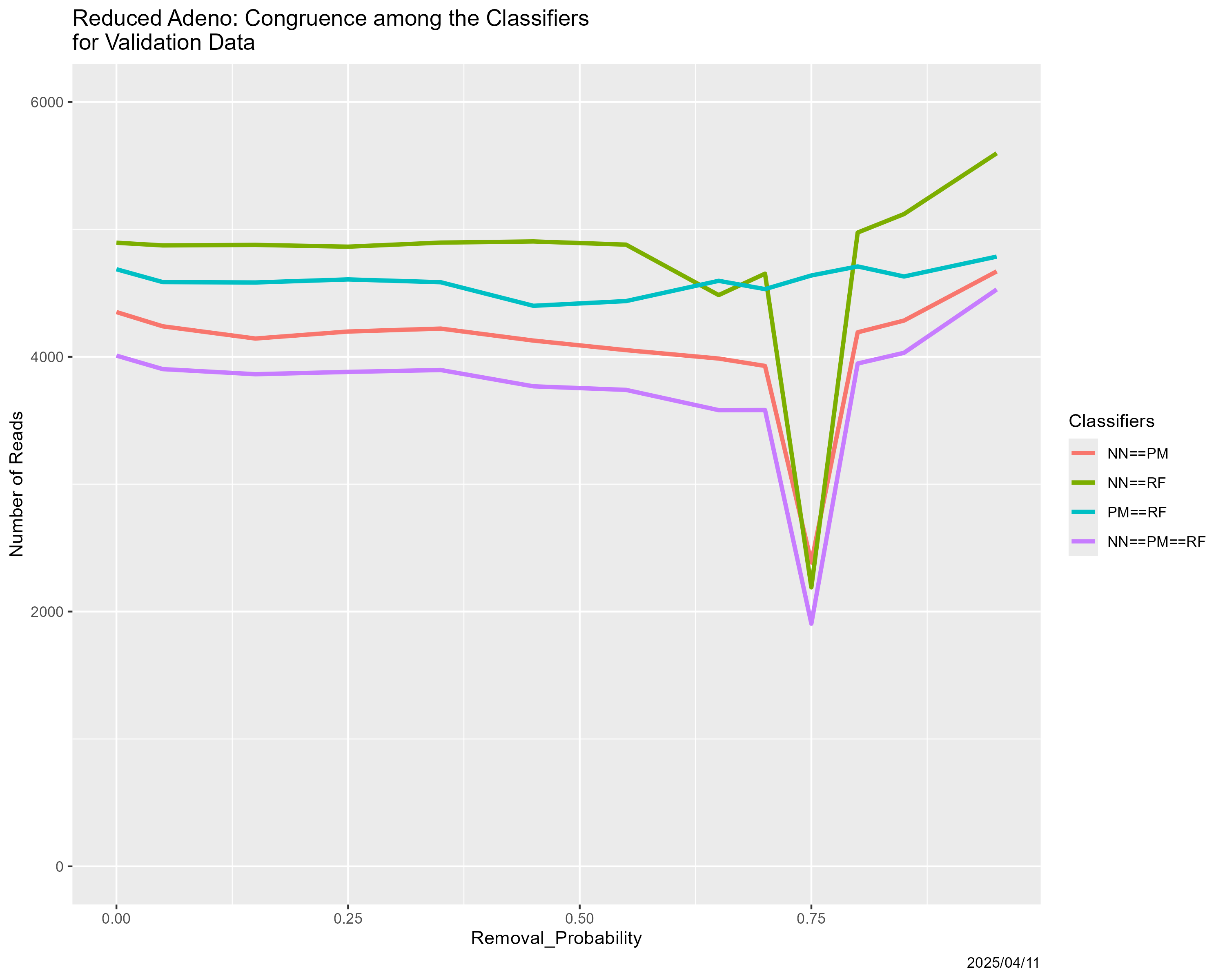}\hspace{.25in}
\includegraphics[width = 1.75in]{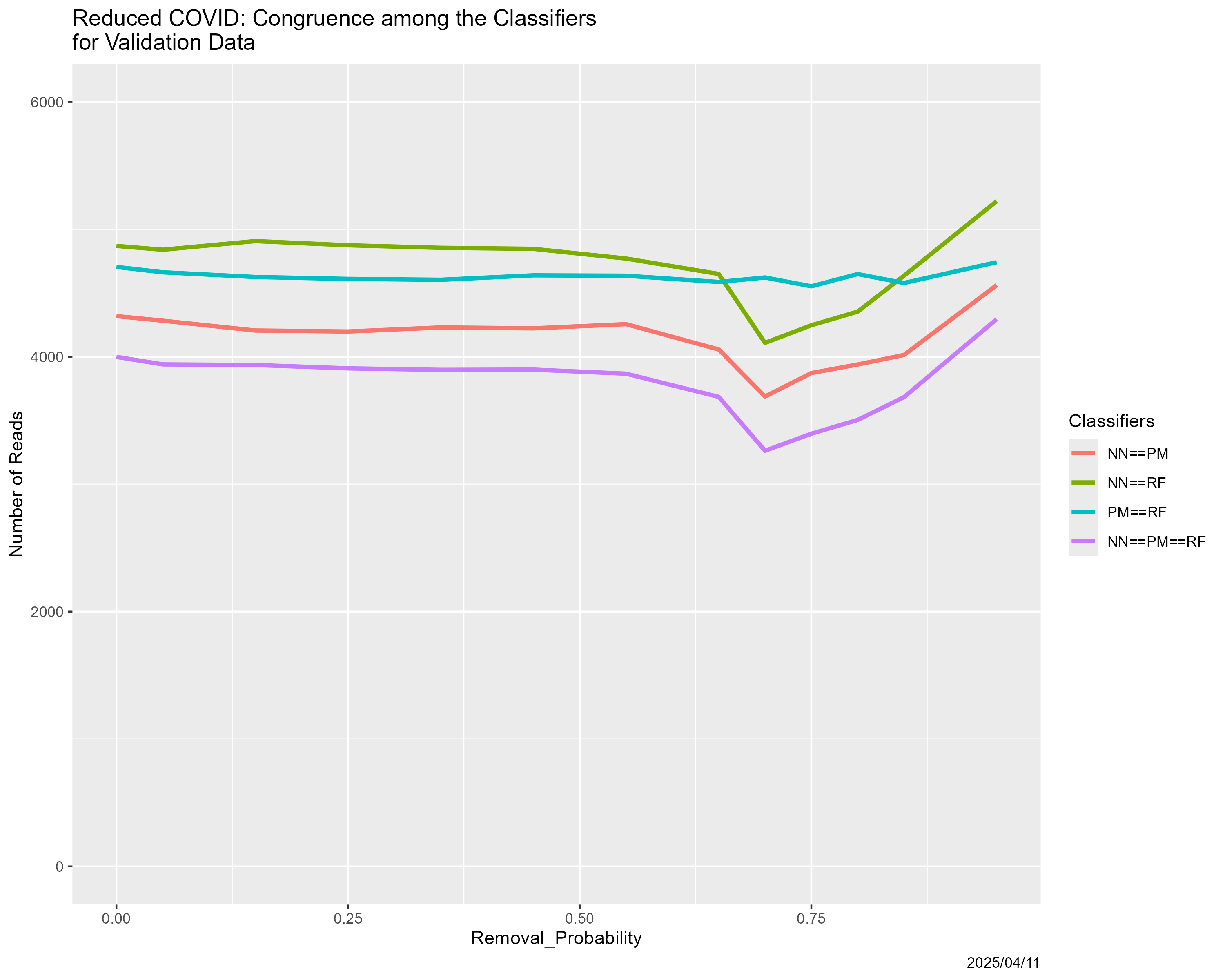}\hspace{.25in}
\includegraphics[width = 1.75in]{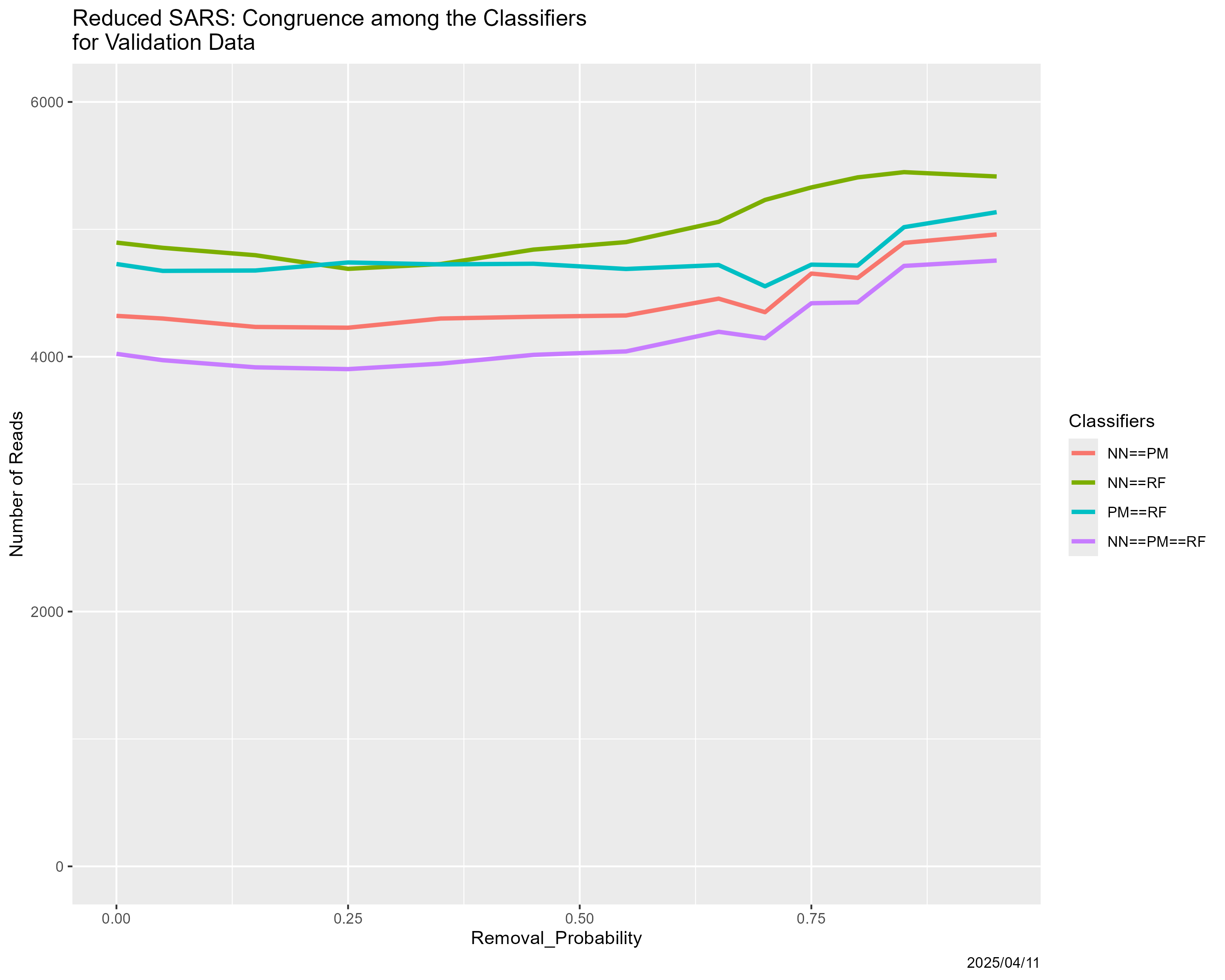}
\caption{Reduced training data reads: classifier congruence as a function of Removal\_Probability. \emph{Left:} reduced Adeno reads. \emph{Center:} reduced COVID reads. \emph{Right:} reduced SARS reads.}
\label{fig.reduced-congruence}
\end{figure}

Prediction results appear in Figures \ref{fig.ReducedAdeno-predictions}--\ref{fig.ReducedSARS-predictions}, and congruence in Figure \ref{fig.reduced-congruence}. Regarding predictions, the disappearance of prediction associated with removed reads happens smoothly and uniformly across removal source and classifiers. For the neural net, it disappears entirely for all three genomes in the vicinity of Removal\_Probability = 0.70. We have no explanation for the oscillation of the neural net with removed Adeno reads at Removal\_Probability = 0.75. which is also reflected in the congruences that involve the neural net. We find no other evidence of breakdown.

\subsubsection{Superfluous Training Data}\label{subsubsec.superfluous}
It may seem paradoxical that too much data can be a data quality problem, especially if they are of high quality, but here we demonstrate that this is so.
Moreover, our proposed explanation gibes perfectly with the overall themes of the paper. Intuitively and surely over-simplistically, a classifier that is trained to recognize data not in \VD\ is wasting its finite potential.

The experiment is as follows: we create a superfluous dataset $\mathcal{S}$ containing 12,000 \emph{E. coli} reads, 785 reads from the Avian flu virus H5N1, 475 reads from Human Papillomavirus (HPV), 453 reads from Norovirus, and 477,357 reads from \emph{P. Gingivalis}, a bacterium associated with chronic periodontitis. Its total size is 491,070. The ``laundry list'' feel reflects the varying sizes of the genomes with which we have worked. None of these is a close relative of the three virus genomes in \TD. The validation data \VD\ remain unchanged. As suggested above and confirmed below, training a classifier to recognize reads present are not in \VD\ is actually counterproductive, and not simply wasteful. The experiment is parameterized by an Add\_Probability taking values in 
$$
\{0, 0.001, 0.005, 0.01, 0.05, 0.1, 0.2, 0.3, 0.4, 0.5, 0.6, 0.7, 0.8, 0.9, 1\}, 
$$
which is applied each element of $\mathcal{S}$ is added to \TD\ before re-training the classifiers. The classifiers are applied to the unaltered validation data. 

\begin{figure}[ht]
\centering
\includegraphics[width = 1.75in]{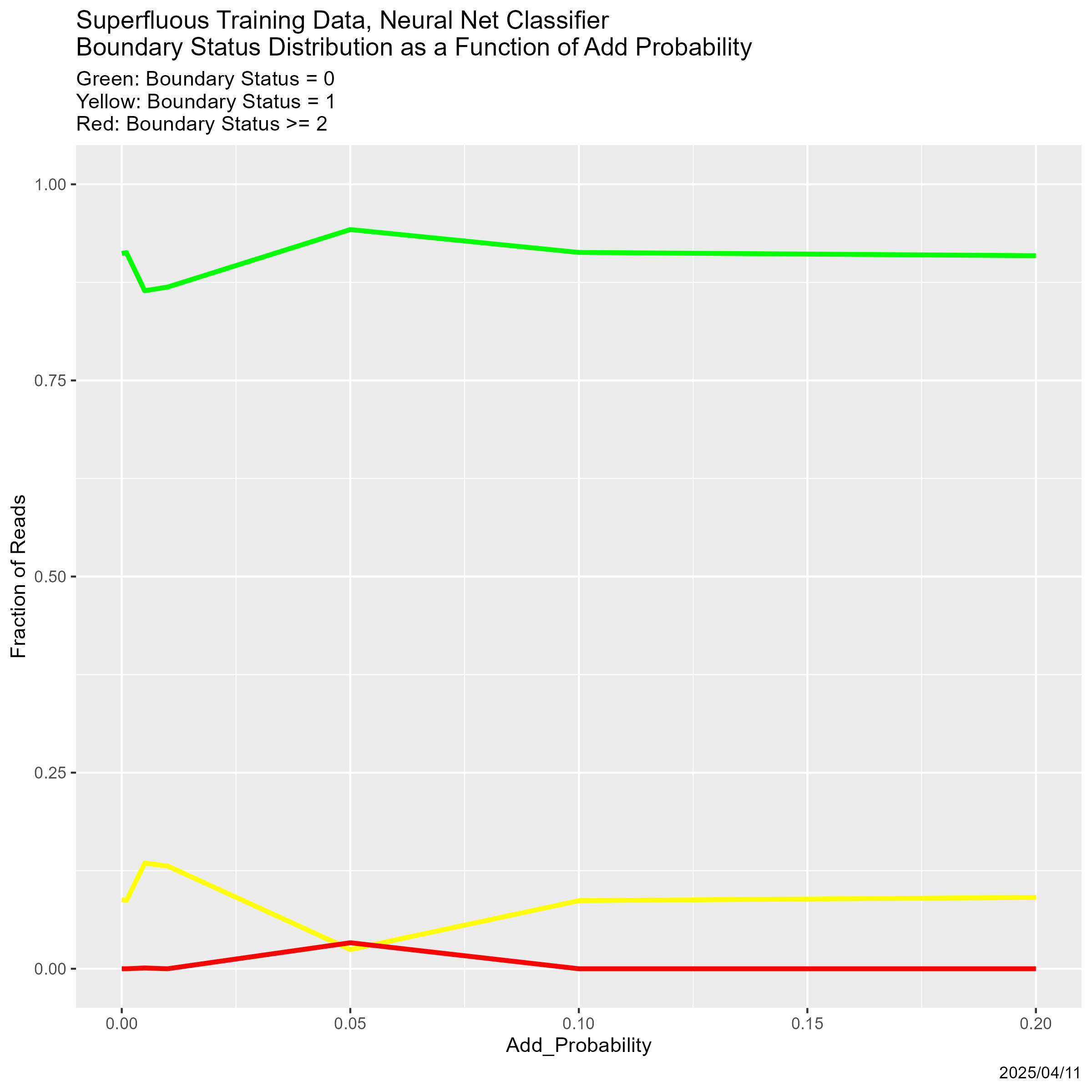}\hspace{.25in}
\includegraphics[width = 1.75in]{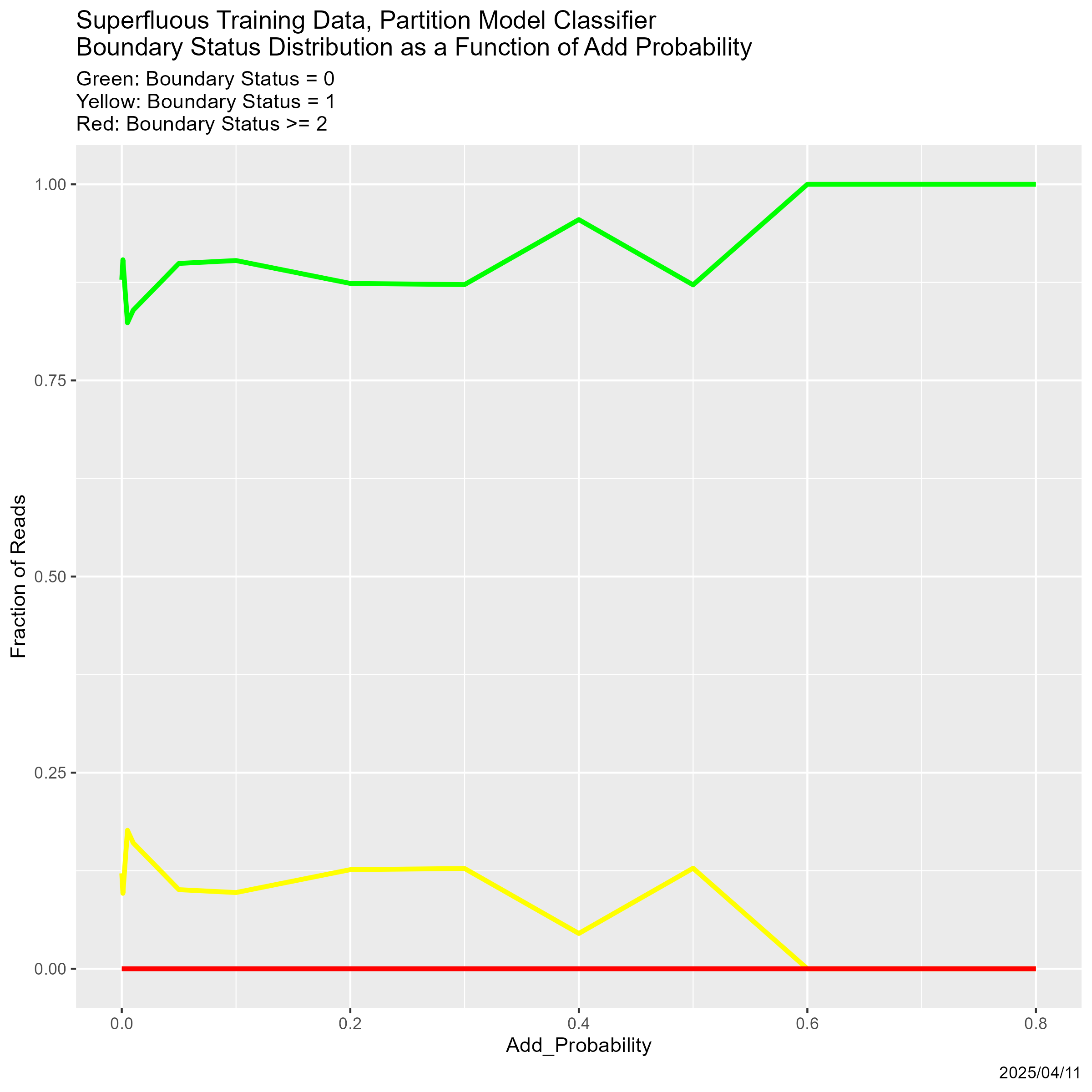}\hspace{.25in}
\includegraphics[width = 1.75in]{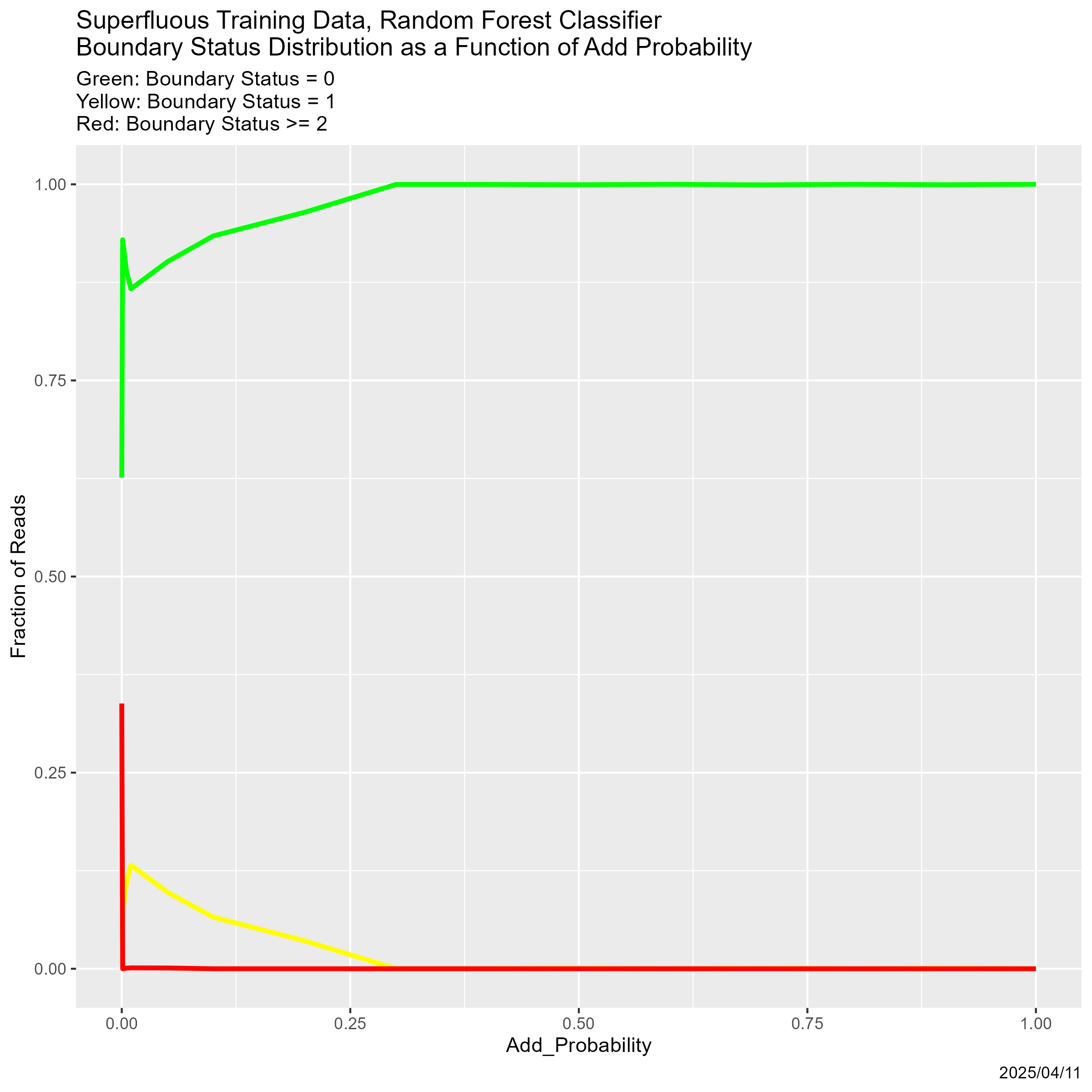}
\caption{Superfluous training data: boundary status distributions as a function of Add\_Probability. In each of these, $\mbox{BS} = 0$ is the green line, $\mbox{BS} = 1$ is the yellow line, and $\mbox{BS} = 2$ is the red line.}
\label{fig.superfluous-boundary}
\end{figure}

\begin{figure}[ht]
\centering
\includegraphics[width=3in]{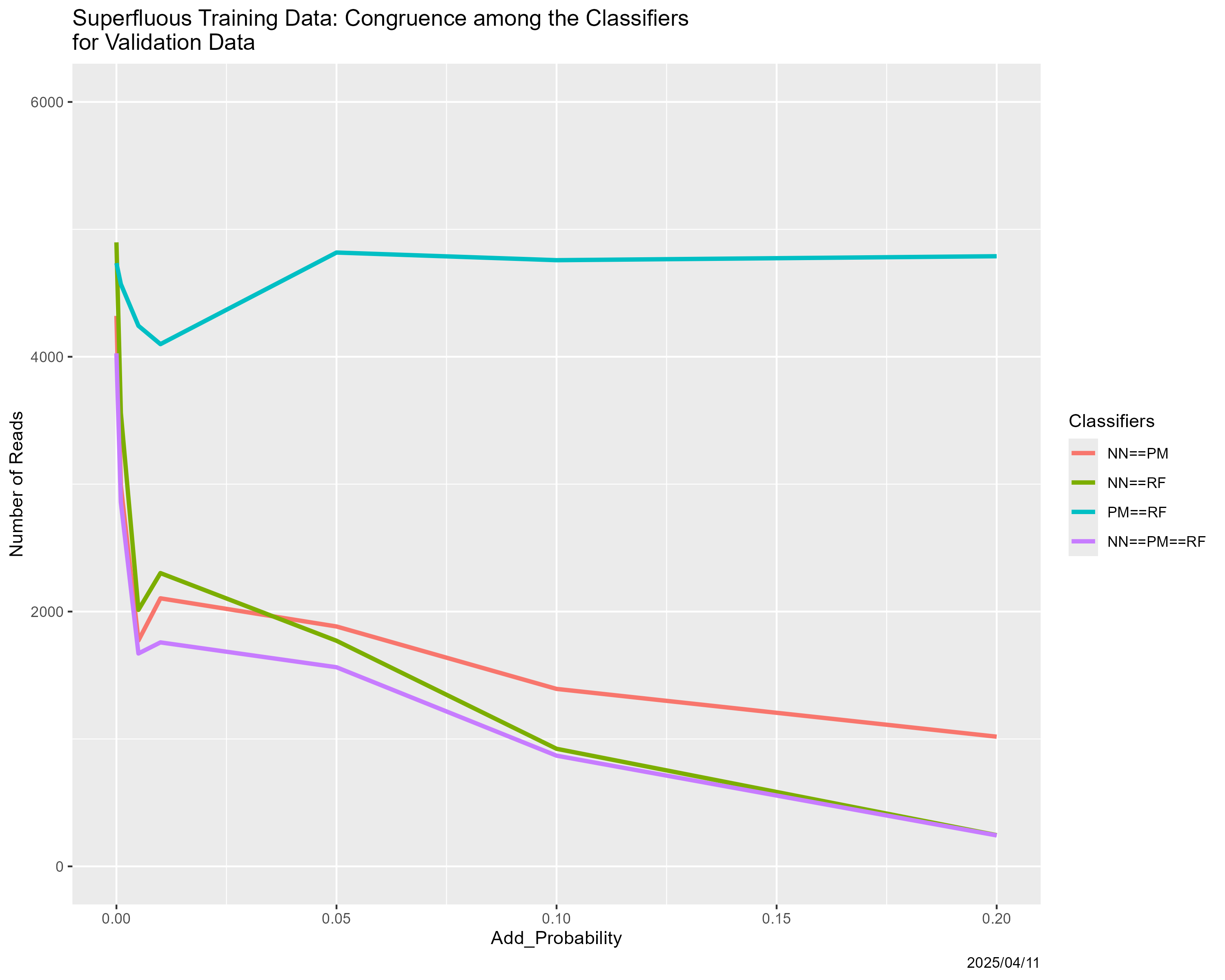}
\caption{Superfluous training data: classifier congruence as a function of Add\_Probability.}
\label{fig.superfluous-congruence}
\end{figure}

They usual results appear in Figures \ref{fig.superfluous-predictions}--\ref{fig.superfluous-congruence}. There is mixed evidence, at best, of breakdown, but seemingly evidence of another form of instability entirely. Once there is any additional of superfluous reads (for Add\_Probability = 0.001, an average of 491 elements of $\mathcal{S}$ are added to \TD, which seems relatively innocuous), Adeno and SARS vanish from the predictions for the neural net. They vanish almost as rapidly for the random forest, and more slowly for the partition model. For these latter two, COVID also ultimately disappears from the predictions, but more slowly. It does not disappear for the neural net. This may be another instance of being right for the wrong reason.

Figure \ref{fig.superfluous-predictions} shows, quite expectedly, that as more and more superfluous reads are added to \TD, the overwhelming majority of which are \emph{P. Gingivalis}, they overwhelm the predictions for \VD. However, even for relatively small values of Add\_Probability, predictions are ``siphoned off.'' For instance, for the partition model, which seems to be the most vulnerable of the three, even at Add\_Probability = 0.001, 107 (of 6000) predictions are shifted to \emph{P. Gingivalis}.

The Boundary Status distributions in Figure \ref{fig.superfluous-boundary} are equally intriguing. For all three classifiers, once there are superfluous reads, the boundary virtually disappears---and rapidly. Combined with the prediction results, we see a false sense of security from the superfluous data, at least if Neighbor Similarity were used as a confidence measure.

Regarding congruence, there is a breakdown for Add\_Probability = 0.01, but with no recovery for the neural net. The partition model--random forest congruence, the breakdown is mild, and recovery occurs.

Clearly there is much more to be learned about this case, but the message to ``be very afraid'' of using a classifier when \TD\ and \VD\ are far apart is unmistakable. It is, of course, not a new message, but our work provides a more nuanced understanding of it.

\begin{figure}[ht]
\centering
\begin{tiny}
\begin{tabular}{rrrrrrrrr}
  \hline
Add\_Probability & Pred\_Adeno & Pred\_COVID & Pred\_SARS & Pred\_EColi & Pred\_H5N1 & Pred\_HPV & Pred\_Norovirus & Pred\_PGingivalis \\ 
  \hline
0.000 & 2015 & 2232 & 1753 & 0 & 0 & 0 & 0 & 0 \\ 
  0.001 & 1945 & 2192 & 0 & 0 & 0 & 1863 & 0 & 0 \\ 
  0.005 & 279 & 2249 & 0 & 0 & 0 & 867 & 2605 & 0 \\ 
  0.010 & 0 & 2467 & 0 & 0 & 0 & 0 & 1335 & 2198 \\ 
  0.050 & 0 & 2873 & 0 & 0 & 0 & 0 & 2923 & 204 \\ 
  0.100 & 0 & 2610 & 0 & 0 & 0 & 0 & 3390 & 0 \\ 
  0.200 & 0 & 2027 & 0 & 0 & 0 & 0 & 3973 & 0 \\ 
   \hline
\end{tabular}

\vspace{.25in}\begin{tabular}{rrrrrrrrr}
  \hline
Add\_Probability & Pred\_Adeno & Pred\_COVID & Pred\_SARS & Pred\_EColi & Pred\_H5N1 & Pred\_HPV & Pred\_Norovirus & Pred\_PGingivalis \\ 
  \hline
0.000 & 2097 & 1958 & 1945 & 0 & 0 & 0 & 0 & 0 \\ 
  0.001 & 1864 & 2006 & 2023 & 0 & 0 & 0 & 0 & 107 \\ 
  0.005 & 1507 & 1986 & 1787 & 0 & 0 & 0 & 0 & 720 \\ 
  0.010 & 1081 & 2003 & 1708 & 0 & 0 & 0 & 0 & 1208 \\ 
  0.050 & 0 & 2098 & 707 & 0 & 0 & 0 & 0 & 3195 \\ 
  0.100 & 0 & 1653 & 450 & 0 & 0 & 0 & 0 & 3897 \\ 
  0.200 & 0 & 1459 & 0 & 0 & 0 & 0 & 0 & 4541 \\ 
  0.300 & 0 & 931 & 0 & 0 & 0 & 0 & 0 & 5069 \\ 
  0.400 & 0 & 788 & 0 & 0 & 0 & 0 & 0 & 5212 \\ 
  0.500 & 0 & 932 & 0 & 0 & 0 & 0 & 0 & 5068 \\ 
  0.600 & 0 & 0 & 0 & 0 & 0 & 0 & 0 & 6000 \\ 
  0.700 & 0 & 0 & 0 & 0 & 0 & 0 & 0 & 6000 \\ 
  0.800 & 0 & 0 & 0 & 0 & 0 & 0 & 0 & 6000 \\ 
   \hline
\end{tabular}

\vspace{.25in}\begin{tabular}{rrrrrrrrr}
  \hline
Add\_Probability & Pred\_Adeno & Pred\_COVID & Pred\_SARS & Pred\_EColi & Pred\_H5N1 & Pred\_HPV & Pred\_Norovirus & Pred\_PGingivalis \\ 
  \hline
0.000 & 2033 & 2119 & 1848 & 0 & 0 & 0 & 0 & 0 \\ 
  0.001 & 2024 & 2081 & 1889 & 0 & 0 & 0 & 0 & 6 \\ 
  0.005 & 1552 & 2110 & 1844 & 0 & 0 & 0 & 0 & 494 \\ 
  0.010 & 793 & 2173 & 1640 & 0 & 0 & 0 & 0 & 1394 \\ 
  0.050 & 0 & 1745 & 250 & 0 & 0 & 0 & 0 & 4005 \\ 
  0.100 & 3 & 972 & 0 & 0 & 0 & 0 & 0 & 5025 \\ 
  0.200 & 2 & 256 & 0 & 0 & 0 & 0 & 0 & 5742 \\ 
  0.300 & 1 & 0 & 0 & 0 & 0 & 0 & 0 & 5999 \\ 
  0.400 & 0 & 0 & 0 & 0 & 0 & 0 & 0 & 6000 \\ 
  0.500 & 3 & 0 & 0 & 0 & 0 & 0 & 0 & 5997 \\ 
  0.600 & 0 & 0 & 0 & 0 & 0 & 0 & 0 & 6000 \\ 
  0.700 & 0 & 0 & 0 & 0 & 0 & 0 & 0 & 6000 \\ 
  0.800 & 0 & 0 & 0 & 0 & 0 & 0 & 0 & 6000 \\ 
  0.900 & 0 & 0 & 0 & 0 & 0 & 0 & 0 & 6000 \\ 
  1.000 & 0 & 0 & 0 & 0 & 0 & 0 & 0 & 6000 \\ 
   \hline
\end{tabular}
\end{tiny}
\caption{Superfluous training data: classifier predictions as a function of Add\_Probability. \emph{Top:} neural net. \emph{Center:} partition model. \emph{Bottom:} random forest.} 
\label{fig.superfluous-predictions}
\end{figure}


\subsubsection{Mixed Data Quality Problems}\label{subsubsec.mixed}
Finally, as we have intimated previously, there is no reason to consider only one form of data quality degradation per experiment. Indeed the spirit of total survey error demands so. Here, we mix the SNP degradation of Section \ref{subsec.snp} with the mislabeling of Section \ref{subsubsec.mislabel}. There are, then, three parameters:
\begin{itemize}
\item
Sel\_Probability, the probability that each read in \TD\ is selected for degradation;
\item
SNP\_Fraction, the conditional probability that a read selected for degradation is subject to SNP degradation---otherwise it is subjected to mislabeling;
\item
SNP\_Probability, with the same meaning as previously: given that a read is selected for SNP degradation, the probability with which each nucleotide in it is changed.
\end{itemize}
Potential values are 
$$
\{0.05, 0.15, 0.25, 0.35, 0.45, 0.55, 0.65, 0.70, 0.75, 0.80, 0.85, 0.95\}
$$
for Sel\_Probability and SNP\_Probability, and 
$$
\{0.05, 0.15, 0.25, 0.35, 0.45, 0.55, 0.65, 0.75, 0.85, 0.95\}
$$ 
for SNP\_Fraction.

And we introduce one further consideration, which whether we can model the contributions of the parameters to congruence. \emph{A priori}, based on results above, congruence should decrease as either Sel\_Probability or SNP\_Probability increases. Whether it increases with SNP\_Fraction sheds light on which of the two forms of degradation is more injurious. A full factorial design of 1440 cases was not feasible computationally, so we instead used a 25-point Fedorov design explicitly meant to support fitting linear models \citep{fedorov1972, wheeler2004}. It and the associated results appear in Table \ref{tab.mixed}.

\begin{table}[ht]
\centering
\caption{Mixed degradation: experimental design and congruence among neural net, partition model and random forest.} 
\label{tab.mixed}
\vspace{.1in}
\begin{small}
\begin{tabular}{rrrrrrr}
  \hline
Sel\_Prob & SNP\_Frac & SNP\_Prob & NN\_PM & NN\_RF & PM\_RF & NN\_PM\_RF \\ 
  \hline
0.050 & 0.050 & 0.050 & 4165 & 4809 & 4549 & 3822 \\ 
  0.150 & 0.050 & 0.050 & 4191 & 4695 & 4541 & 3770 \\ 
  0.850 & 0.050 & 0.050 & 2334 & 2770 & 2410 & 1135 \\ 
  0.950 & 0.050 & 0.050 & 2722 & 3171 & 2777 & 1555 \\ 
  0.950 & 0.150 & 0.050 & 2582 & 3734 & 2569 & 1704 \\ 
  0.050 & 0.850 & 0.050 & 4229 & 4872 & 4623 & 3917 \\ 
  0.950 & 0.850 & 0.050 & 3910 & 4652 & 4441 & 3561 \\ 
  0.050 & 0.950 & 0.050 & 4363 & 4882 & 4728 & 4032 \\ 
  0.850 & 0.950 & 0.050 & 4028 & 4833 & 4485 & 3735 \\ 
  0.950 & 0.950 & 0.050 & 4257 & 4792 & 4602 & 3886 \\ 
  0.050 & 0.050 & 0.150 & 4202 & 4849 & 4578 & 3866 \\ 
  0.050 & 0.950 & 0.150 & 4243 & 4826 & 4653 & 3912 \\ 
  0.950 & 0.050 & 0.850 & 2956 & 3970 & 2821 & 2031 \\ 
  0.050 & 0.950 & 0.850 & 4228 & 4778 & 4679 & 3893 \\ 
  0.950 & 0.950 & 0.850 & 2247 & 3017 & 2400 & 1188 \\ 
  0.050 & 0.050 & 0.950 & 4268 & 4787 & 4704 & 3938 \\ 
  0.150 & 0.050 & 0.950 & 3982 & 4730 & 4394 & 3644 \\ 
  0.850 & 0.050 & 0.950 & 2626 & 3409 & 2714 & 1602 \\ 
  0.950 & 0.050 & 0.950 & 3076 & 3841 & 3167 & 2179 \\ 
  0.050 & 0.150 & 0.950 & 4240 & 4866 & 4625 & 3923 \\ 
  0.050 & 0.850 & 0.950 & 4166 & 4835 & 4642 & 3874 \\ 
  0.950 & 0.850 & 0.950 & 3299 & 4716 & 3072 & 2711 \\ 
  0.050 & 0.950 & 0.950 & 4175 & 4809 & 4591 & 3860 \\ 
  0.850 & 0.950 & 0.950 & 2535 & 2835 & 3082 & 1632 \\ 
  0.950 & 0.950 & 0.950 & 2569 & 3632 & 2825 & 1746 \\ 
   \hline
\end{tabular}
\end{small}
\end{table}

We then fitted linear models with each of the four congruences as the response and the three parameters of the experiment as predictors. This is not a modeling decision informed by science, but is a reasonable place to begin. Addition of quadratic terms, as might be proposed based on the modeling discussion in Section \ref{subsubsec.snp-selective}, does not improve model fit. The four models are consistent with one another, and with the exception of that for NN\_RF, all fit the data comparably. The intercepts, which are the congruences in the absence of degradation, are consistent with those in previous sections. For all four models, the coefficient of Sel\_Probability is strongly negative and highly significant: the more overall degradation, the lower the congruence. While in no case is the coefficient of SNP\_Probability significant, it is always negative: more \SNPs\ mean less congruence. And finally, the coefficient of SNP\_Fraction, while again not significant, is always positive and of the same order to magnitude. The interpretation is that shifting from Mislabeling to SNP degradation \emph{increases} congruence, meaning that the former is more disruptive.

\begin{table}[ht]
\centering
\caption{Estimated coefficients and adjusted $R^2$-values for linear models of congruence as a function of Sel\_Pobability, SNP\_Fraction and SNP\_Probability. Labels for statistical significance are the standard ones.}
\label{fig.mixed-model-results}
\vspace{.1in}
\begin{tabular}{lrrrrr}
\hline
Response & \multicolumn{4}{c}{Predictors} & Adjusted $R^2$
\\
\cline{2-5}
 & Intercept & Sel\_Prob & SNP\_Frac & SNP\_Prob
\\
\hline 
NN\_PM & 4305.3 (***) & -1383.1 (***) & 336.6 & -365.6 & 0.6403
\\
NN\_RF & 4818.8 (***) & -1157.4 (***) & 344.1 & -232.8 & 0.4468
\\
PM\_RF & 4670.7 (***) & -1669.5 (***) & 515.6 & -418.9 & 0.6794
\\
NN\_PM\_RF & 3964.7 (***) & -1934.0 (***) & 576.3 (.) & -462.3 & 0.6374
\\   
\hline
\end{tabular}
\end{table}


\clearpage\newpage
\section{Discussion}\label{sec.discussion}
Despite its length, this paper leaves many questions unanswered.

\subsection{Scientific Generalizability}\label{subsec.generalizability}
We have examined one scientific context, with one input graph $\mathcal{I}$ and one output space $\mathcal{O}$, four classifiers, and a handful of context-specific data quality characteristics. Of course, we have no evidence-based claim that what have observed occurs more generally. We conjecture that it does, but asserting that breakdown is an inherent characteristics of classifiers, even those whose input space is a graph, let alone for all forms of data quality degradation, is simply silly. Nevertheless, our tools apply and our explanations make sense in virtually any context. And the world is full of discontinuities, including classifier boundaries, which are local discontinuities with respect to the analysis data, and are omnipresent.

Generalizability issues are of at least three varieties: with respect to scientific context/nature of datasets, classifiers, and data quality measures. In this paper we have addressed the latter two to some extent, but the first not at all. Doing so is most urgent question, and we plan to investigate it.

\subsection{Theory vs.\ Empirical Evidence}\label{subsec.theory-empirical}
We turn now to the question with which Section \ref{subsec.snp} ended ``Why SNP\_Probability = 0.75?'' Can there be theory that explains this? Given that our experiments are simulations in which we examined a limited set of possible values of SNP\_Probability, it seems sensible to believe that 0.75 is the mean (or another characteristic) of some random variable that varies depending on exactly which bases in which reads are altered. While the same breakdown phenomenon may exist in other settings, the numerical value is setting- and data quality measure-specific.

There may be a deeper issue here, which is the role of empirical analyses in understanding computational tools that lack theory capable of predicting their behavior in reality, especially for datasets sampled from no describable population. To the extent that it is an issue for the classifiers here, it is a much more profound issue for large language models (LLMs) and other AI tools.  

\subsection{Actionable Implications}\label{subsec.action}
And now the really big question: so what? With slightly more precision, our high-level path to impact is that:
\begin{itemize}
\item
Classifier performance depends on training data, which we have demonstrated in one context.
\item
In today's world, many datasets are collected with limited attention to quality, and little-to-no characterization or quantification of it.
\item
The principled way to deal with these two phenomena is via uncertainties associated with classifier decisions, as measured by Neighbor Similarity, for instance. Doing so without changing the training data quality is not possible.
\end{itemize}

We also note that one of us (AFK) has argued that data quality is for aficionados, and that it is really decision quality that matters \citep{karr-josdiscussion-2013}. To this end, combining this paper with findings in \cite{bayesboundary2026} and \cite{amost2024}, what may be the central data quality issue emerges---beyond the quality of \TD\ (addressed here) and that of \VD, there is the quality of relationship between the training data and analysis data. ``Everyone knows'' that if they are too divergent, performance suffers. This paper and \cite{bayesboundary2026} are steps to quantifying both the divergence and the performance consequences. Is there a paradigm, probably iterative and therefore necessarily Bayesian, that addresses the problem? For instance, start with a large training dataset \TD, run the classifier(s) and see whether there are unused elements of \TD\ that can be removed. Assess quality of elements \TD\ to see if there are low quality elements whose removal would improve performance. Then, re-run the process. And at some point, assess whether there are elements of \VD\ that may be misclassified because they are of low quality. The role of Bayes in such a paradigm is to track uncertainties properly.

\section{Conclusion}\label{sec.conclusion}
In this paper, we have shown that in one context, decreasing the quality of data used to train classifiers impairs their performance, in some cases causing it to break down entirely. We have examined multiple mechanisms for effecting the quality decrease, which behave similarly. The consequences for decisions based on classifier output are beginning to be explored.

\section*{Acknowledgements}\label{sec.acknowledgements}
Precursor research by AFK was supported in part by NIH grant 5R01AI100947--06, ``Algorithms and Software for the Assembly of Metagenomic Data,'' to the University of Maryland College Park (Mihai Pop, PI). He thanks Professor Pop for numerous insightful discussions.

Both authors gratefully acknowledge ongoing support and collaboration from Professor Adam Porter of the University of Maryland College Park. This paper flows directly from \cite{dqdegradation-2021}, \cite{amost2024}, and \cite{bayesboundary2026}, of which he is a co-author.

\section*{Declarations}\label{sec.declarations}
\begin{itemize}
\item
Both authors contributed equally to the paper.
\item
The authors declare no conflicts of interest or competing interests.
\item
Data and R code for the analyses in this paper are available at \texttt{https://doi.org/TBD}.
\end{itemize}

\clearpage\newpage
\appendix
\section{Neighbor Similarity for SNP Degradation}
See Section \ref{subsec.snp} for discussion.

\begin{figure}[ht]
\begin{center}
\includegraphics[width=3in]{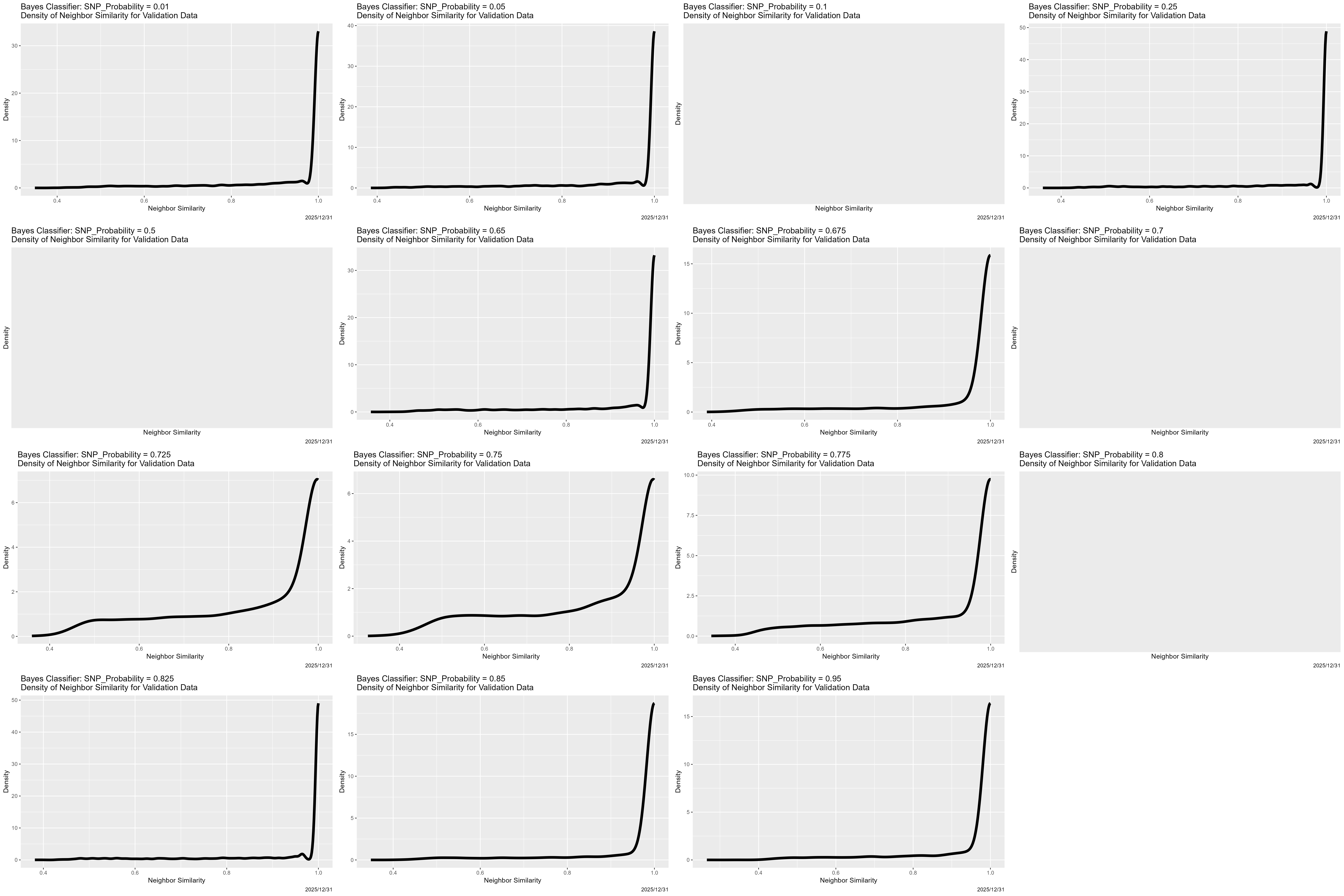}
\end{center}
\caption{SNP degradation, Bayes classifier: Neighbor Similarity as a function of SNP\_Probability.}
\label{fig.snp-ns-bayes}
\end{figure}

\begin{figure}[ht]
\begin{center}
\includegraphics[width=3in]{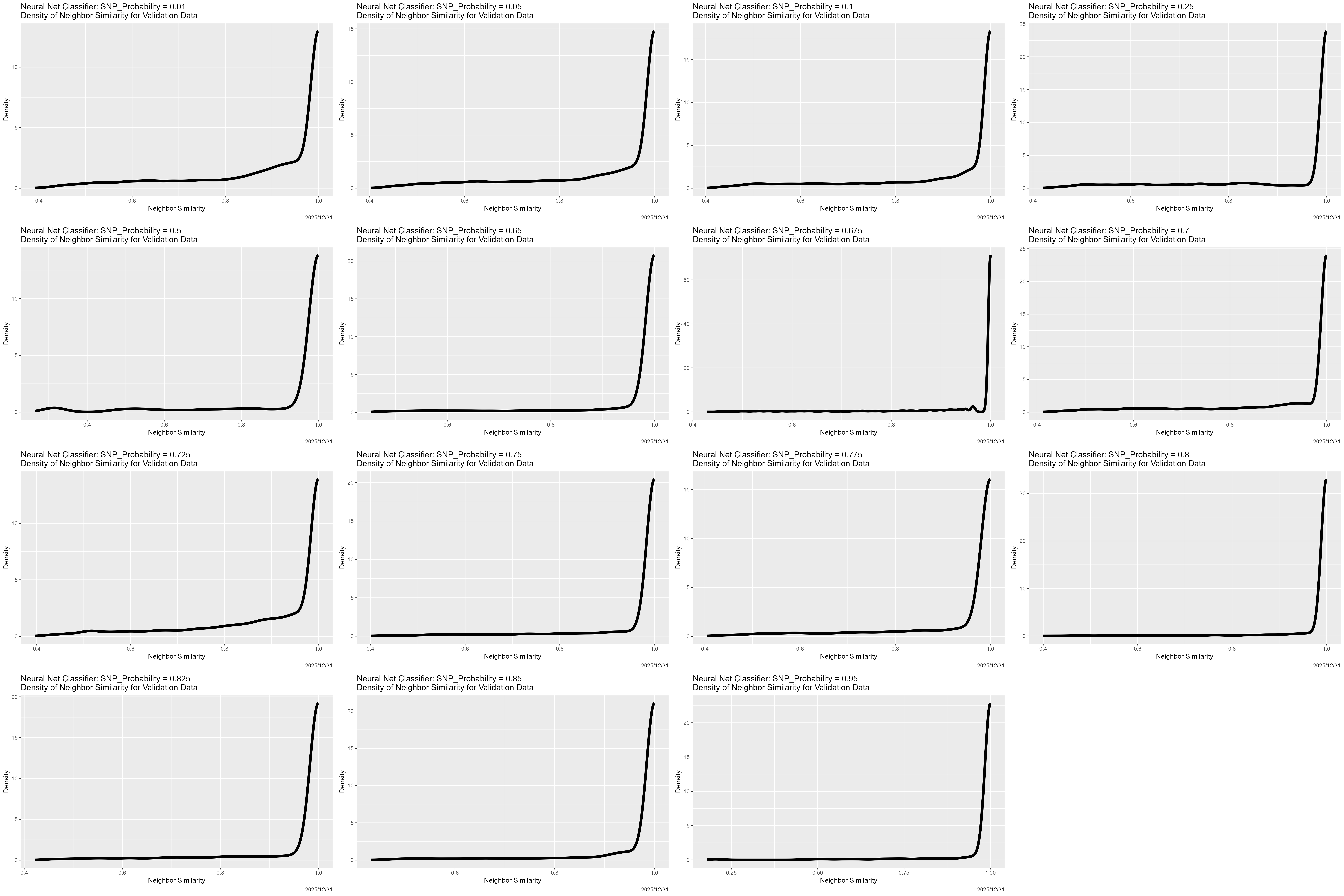}
\end{center}
\caption{SNP degradation, neural net classifier: Neighbor Similarity as a function of SNP\_Probability.}
\label{fig.snp-ns-nn}
\end{figure}

\begin{figure}[ht]
\begin{center}
\includegraphics[width=3in]{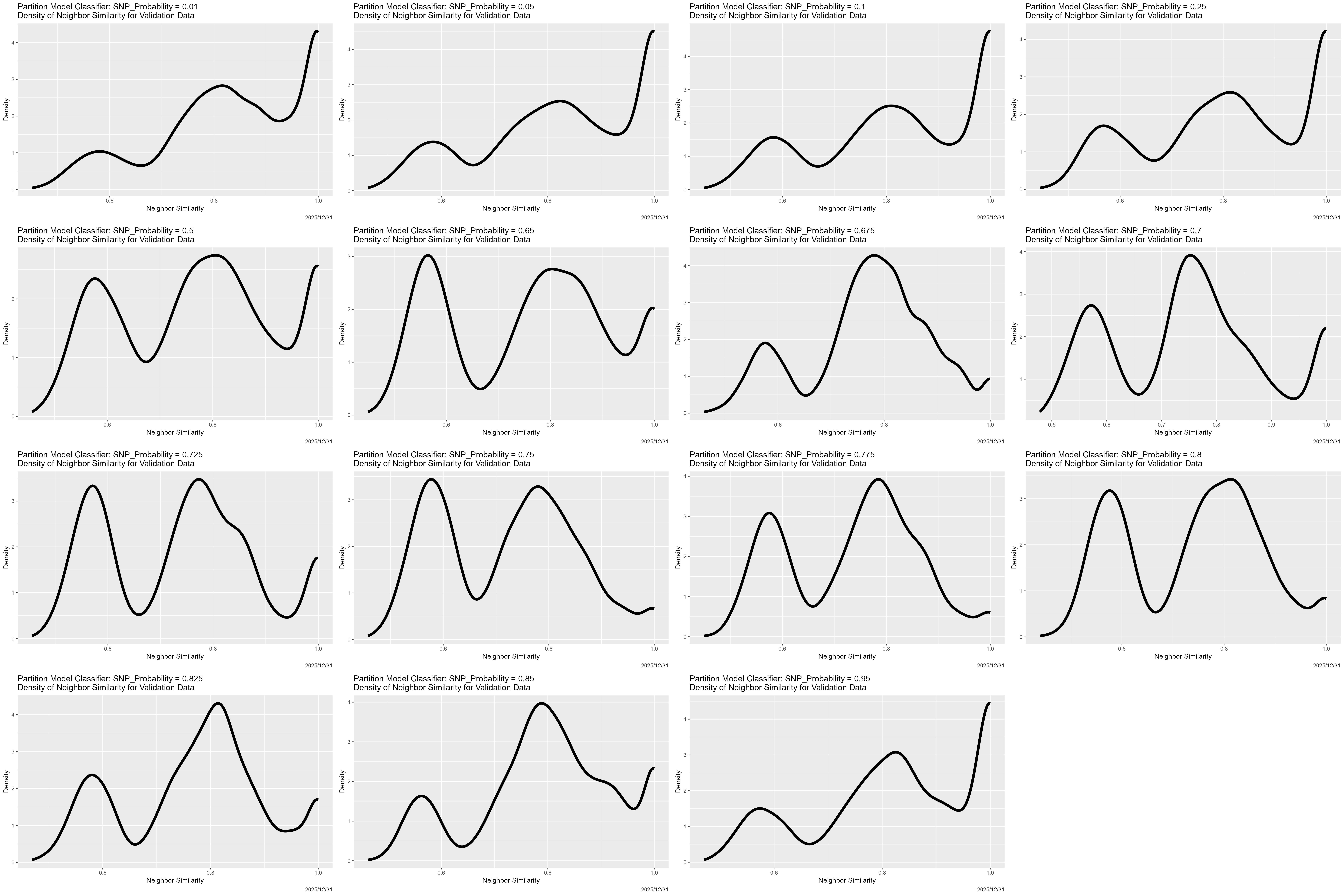}
\end{center}
\caption{SNP degradation, partition model classifier: Neighbor Similarity as a function of SNP\_Probability.}
\label{fig.snp-ns-pm,}
\end{figure}

\begin{figure}[ht]
\begin{center}
\includegraphics[width=3in]{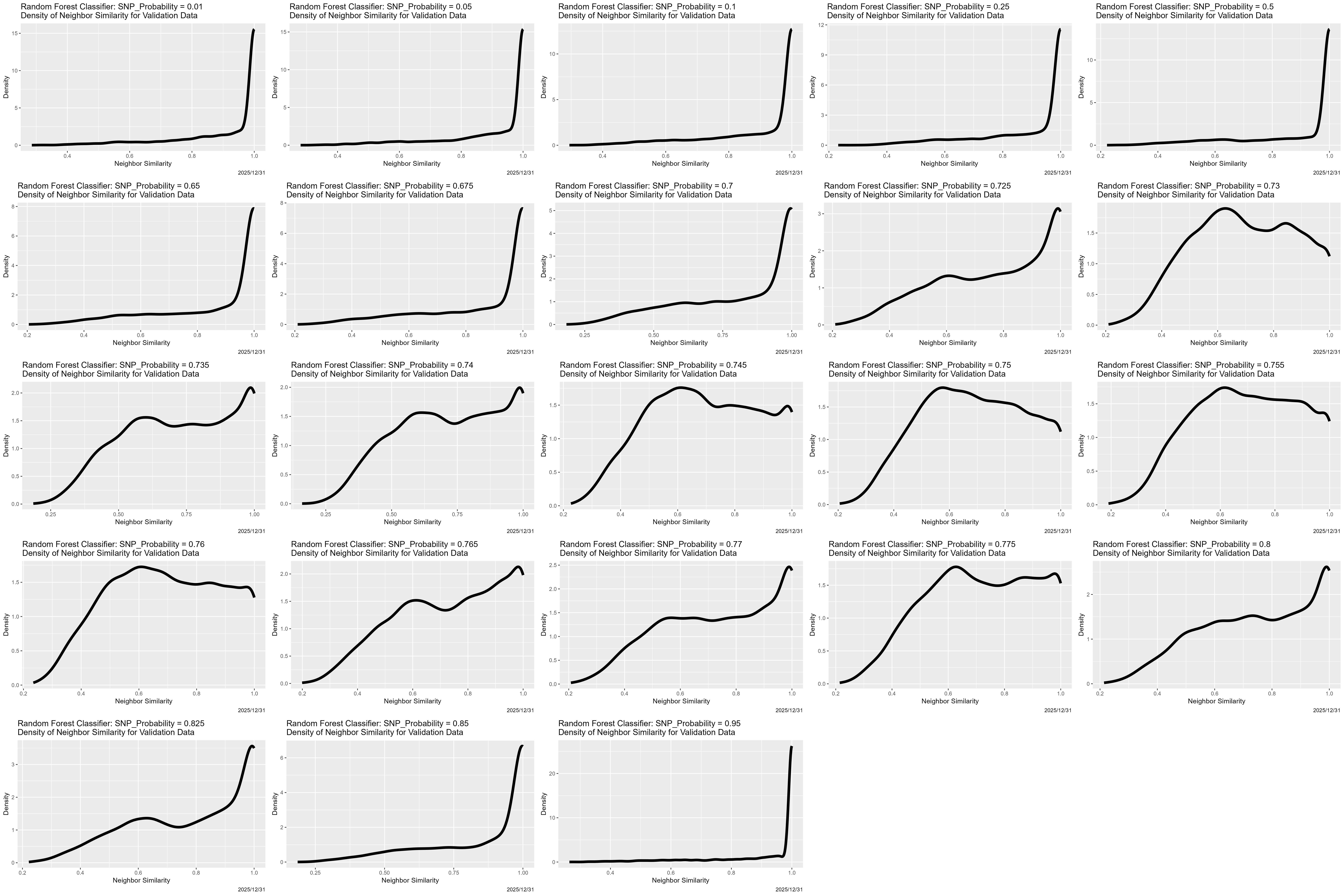}
\end{center}
\caption{SNP degradation, random forest classifier: Neighbor Similarity as a function of SNP\_Probability.}
\label{fig.snp-ns-rf}
\end{figure}


\def\thisfile{TimesTemplate.tex}
\def\thisfiledate{2021/10/14}
\typeout{***** `\thisfile' <\thisfiledate> *****}
\end{document}